\documentclass{article} % For LaTeX2e
\usepackage[preprint]{colm2025_conference}
\usepackage{microtype}
\usepackage{hyperref}
\usepackage{url}
\usepackage{tcolorbox}
\usepackage{booktabs}
\usepackage{lineno}
\usepackage{tablefootnote}
\usepackage{multirow}
\usepackage{dsfont}
\usepackage{xcolor}
\usepackage{graphicx}
\usepackage{enumitem}
\usepackage{listings}
\usepackage{tcolorbox}
\usepackage{amssymb}
\usepackage{pifont}
\usepackage{caption}
\usepackage{subcaption}
\usepackage{wrapfig}  % Include in preamble
\tcbuselibrary{listingsutf8}

\definecolor{darkblue}{rgb}{0, 0, 0.5}
\hypersetup{colorlinks=true, citecolor=darkblue, linkcolor=darkblue, urlcolor=darkblue}

\definecolor{zoey green}{rgb}{0.684,0.836,0.227}
\newcommand{\ignore}[1]{}

\definecolor{bggray}{rgb}{0.95, 0.95, 0.95}
\usepackage[%
    framemethod=tikz,
    skipbelow=\topskip,
    skipabove=\topskip
]{mdframed}
\mdfsetup{%
    leftmargin=0pt,
    rightmargin=0pt,
    backgroundcolor=bggray,
    middlelinecolor=black,
    roundcorner=3
}
\newtcolorbox[list inside=prompt,auto counter,number within=section]{prompt}[1][]{
    colbacktitle=black!60,
    fonttitle=\small,
    coltitle=white,
    fontupper=\footnotesize,
    boxsep=4pt,
    left=0pt,
    top=0pt,
    bottom=0pt,
    boxrule=1pt,
    #1,
}

\title{\includegraphics[height=1.0em]{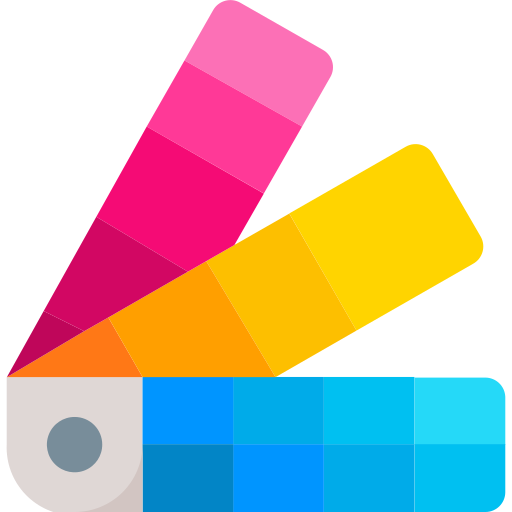} \textsc{GraphicBench}: A Planning Benchmark for Graphic Design with Language Agents}
\colmpreprinttrue
\colmsubmissionfalse
\colmfinalfalse

\author{\textbf{Dayeon Ki}\textsuperscript{\ding{70}}\textsuperscript{\ding{81}}, 
\textbf{Tianyi Zhou}\textsuperscript{\ding{70}}, 
\textbf{Marine Carpuat}\textsuperscript{\ding{70}}, \\
\textbf{Gang Wu}\textsuperscript{\ding{95}}\textsuperscript{\ding{59}}, 
\textbf{Puneet Mathur}\textsuperscript{\ding{95}}\textsuperscript{\ding{59}}, 
\textbf{Viswanathan Swaminathan}\textsuperscript{\ding{95}}\textsuperscript{\ding{59}} \\
\textsuperscript{\ding{70}}University of Maryland, College Park \\
\textsuperscript{\ding{95}}Adobe Research \\
\texttt{dayeonki@umd.edu}
}

\begin{document}

\ifcolmsubmission
\linenumbers
\fi

\maketitle

\begingroup
\renewcommand{\thefootnote}{\ding{81}} \footnotetext{Work done during internship at Adobe Research.}
\renewcommand{\thefootnote}{\ding{59}} \footnotetext{Internship co-advisors.}
\endgroup

% A lot of multi-agent planning benchmark studied -> but mostly on tasks with predefined end goal states
% We introduce DesignBench + DesignTown
% Evaluation with 6 LLMs -> present main findings
% Future research
\begin{abstract}
Large Language Model (LLM)-powered agents have unlocked new possibilities for automating human tasks. While prior work has focused on well-defined tasks with specified goals, the capabilities of agents in creative design tasks with \textit{open-ended} goals remain underexplored. 
We introduce \textsc{GraphicBench}, a new planning benchmark for graphic design that covers 1,079 user queries and input images across four design types. We further present \textsc{GraphicTown}, an LLM agent framework with three design experts and 46 actions (tools) to choose from for executing each step of the planned workflows in web environments.
Experiments with six LLMs demonstrate their ability to generate workflows that integrate both explicit design constraints from user queries and implicit commonsense constraints. However, these workflows often do not lead to successful execution outcomes, primarily due to challenges in: \textbf{(1)} reasoning about spatial relationships, \textbf{(2)} coordinating global dependencies across experts, and \textbf{(3)} retrieving the most appropriate action per step. We envision \textsc{GraphicBench} as a challenging yet valuable testbed for advancing LLM-agent planning and execution in creative design tasks.\footnote{Code and data will be released on the Adobe Research Github after internal approval: \url{https://github.com/adobe-research}.}
\end{abstract}

% \footnote{We will release the dataset and code upon publication.}

% \begin{quote}
% \textit{Design is a high-level and complex thinking activity of human beings, using existing knowledge and tech to solve problems and create new things.}\\
% \hfill --- General Design Theory (1998)
% \end{quote}

\section{Introduction}

% Multi-agent prior works
Recent advances of Large Language Models (LLM) agents have expanded the potential for automating various human tasks. Prior research has explored different aspects of LLM agent capabilities \citep{sumers2024cognitivearchitectureslanguageagents}, including tool use \citep{schick2023toolformer, qin2023toolllm, shen2023hugginggptsolvingaitasks, qin2024tool, liang2024taskmatrix}, reasoning strategies \citep{wei2022chain, yao2022react, shinn2023reflexionlanguageagentsverbal}, and evaluation methods \citep{zhuge2024agentasajudgeevaluateagentsagents}. However, most studies focus on tasks with predefined end-goal states, such as filling out spreadsheets and generating charts \citep{wu2024oscopilotgeneralistcomputeragents}, or adding items to a shopping cart \citep{koh-etal-2024-visualwebarena}.

% No work exploring creative design generation
On the other hand, research on the planning capabilities of LLM agents for creative design tasks remains limited, primarily due to underspecified open-ended goals from users \citep{guo2024lubanbuildingopenendedcreative, ge2025autopresentdesigningstructuredvisuals}.
They require delicate planning that translates a high-level user request into a structured workflow composed of executable sub-tasks that collectively produce the final design. This is inherently complex, posing multiple challenges:  
% in terms of handling user queries, planning workflows, and evaluating design outcomes: 
\textbf{(1)} A complex design often requires collaborations among multiple experts; \textbf{(2)} A design workflow is usually long-horizon, involving a sequence of decisions for expert selection, action calls, and tool uses, which constitute  
% a large number of interdependent decisions across agents, with 
an expansive action space to explore \citep{xie2024travelplanner};  \textbf{(3)} A design plan must accommodate both explicit constraints from user queries (e.g., ``\textit{the title text color must be white}'') and implicit constraints inferred through commonsense reasoning (e.g., ``\textit{the background should contrast with the color of text elements}'') since user queries are often incomplete with unspecified details~\citep{qian-etal-2024-tell}; 
% might be incomplete and lack explicit details necessary for planning \citep{qian-etal-2024-tell}. 
\textbf{(4)} Assessing design outcomes is inherently subjective, as the notion of \textit{better} design varies among individuals. These challenges raise a key question: Can LLM agents generate cohesive workflow plans for creative design tasks with only high-level or open-ended user queries provided?

\begin{figure*}
    \centering
    \includegraphics[width=\linewidth]{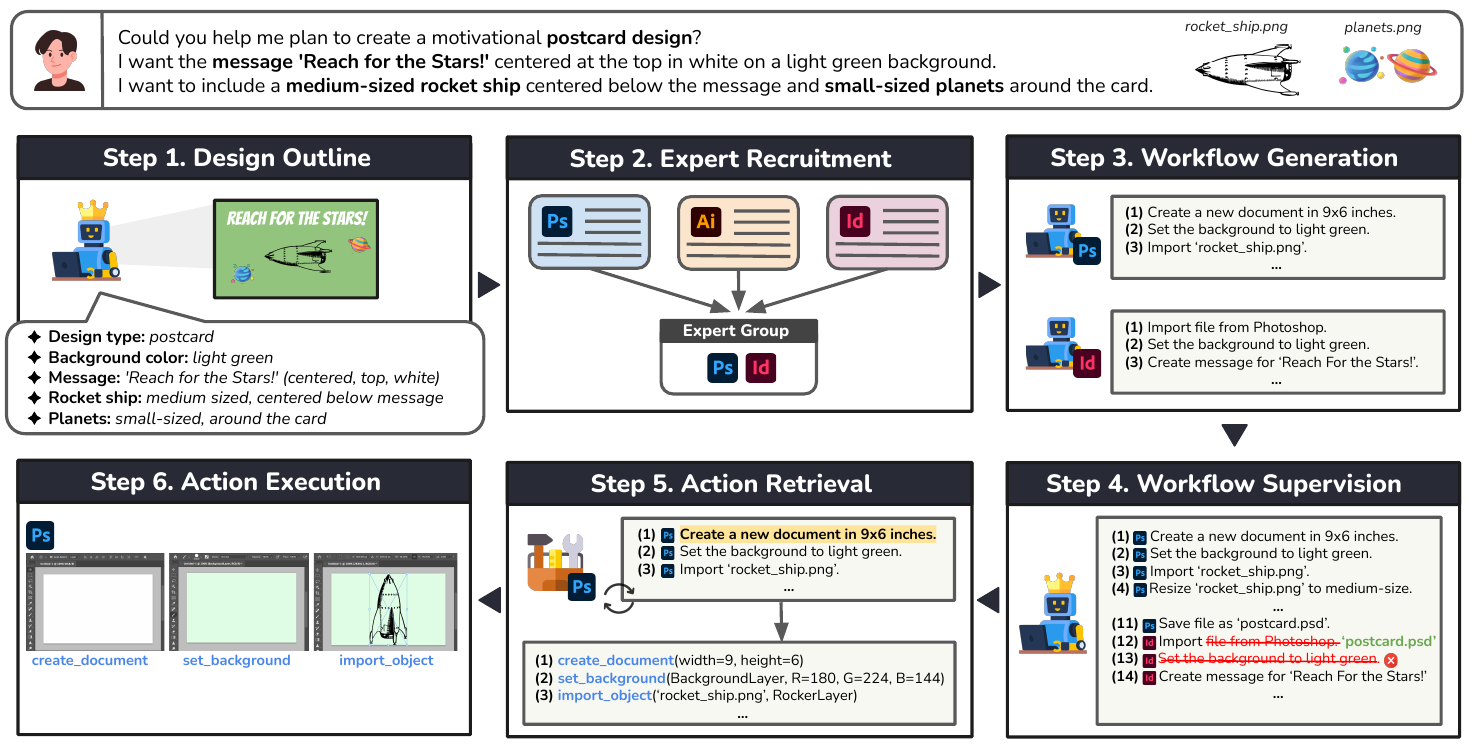}
    \caption{Overview of the \textsc{GraphicTown} framework. The user query and input image captions are from \textsc{GraphicBench}. The same LLM is used across all steps.}
    \label{fig:main_figure}
\end{figure*}

% \textbf{Step 1:} Given the user query and captions of input images from \textsc{CreativeBench}, the supervisor agent generates a design outline. \textbf{Step 2:} Expert agents are recruited according to their expertise. \textbf{Step 3:} Each expert agent formulates a workflow plan. \textbf{Step 4:} The supervisor agent integrates the individual plans into a cohesive design plan. \textbf{Step 5:} For each step, the supervisor retrieves the appropriate action and its parameters. \textbf{Step 6:} The actions are executed within the design tool environments.

% Creative content generation is a difficult task for humans
In this paper, we focus on graphic design, a task that is challenging even for humans as it requires specialized knowledge of design tools, cost, and effort \citep{Bedford_2006}. We introduce \textbf{\textsc{GraphicBench}} as a testbed (\S \ref{sec:creativebench}), consisting of 1,079 user queries paired with input images, covering four design types \---\ book covers, business cards, postcards, and posters \---\ capturing a broad range of design concepts. We further propose an LLM agent framework, \textbf{\textsc{GraphicTown}} (Figure \ref{fig:main_figure}, \S \ref{sec:creativetown}), to evaluate the planning abilities of LLM agents for creative design on \textsc{GraphicBench}. \textsc{GraphicTown} consists of the following steps: (1) generate a design outline based on the user query and image captions; (2) recruit expert agents; (3) generate a workflow for each expert; (4) integrate experts' workflows into a cohesive plan; (5) retrieve appropriate actions for each step in the plan, and (6) execute the plan to produce a final outcome of the design. For action retrieval, we define a set of 46 actions executable within three environments of web-based design tools.

% Key points from results and analysis
We comprehensively evaluate six LLMs, ranging from smaller open-weights models to larger closed-source models, on their ability to deliver a plan of design workflow based on the user query and images. Our key findings are as follows:

\begin{itemize}[leftmargin=*, itemsep=2pt, parsep=-1pt]
    \item All tested LLM agents can plan design workflows that incorporate both explicit design constraints from user queries and implicit commonsense constraints.
    \item LLM-generated design workflows often include action sequences that closely align with those in human-developed workflows.
    \item However, these planned workflows often fail to yield successful execution outcomes. Further error analysis reveals three common failure modes: \textbf{(1)} difficulty in spatial reasoning between design components, \textbf{(2)} lack of coordination across experts to manage global dependencies, and \textbf{(3)} retrieval of invalid actions.
\end{itemize}
\section{\includegraphics[height=1.0em]{figures/logo/pantone.png} \textsc{GraphicBench}}
\label{sec:creativebench}

% We outline the construction pipeline of \textbf{CreativeBench}, which involves the following steps: \textbf{1)} Reference plan annotation, \textbf{2)} Query construction, \textbf{3)} Automatic quality check, \textbf{4)} Image pairing, and \textbf{5)} Human quality check. 

In this section, we outline our dataset curation pipeline for \textsc{GraphicBench}, which contains 1,079 pairs of diverse user queries and input images across four types of graphic design: book covers, business cards, postcards, and posters. The dataset is divided into training and test sets, with the training set containing 5 instances per design type with human-annotated reference plans (20 pairs in total) and the test set comprising 1,059 instances. Detailed distributions and examples are shown in Table \ref{tab:creativebench_examples}.

% \footnote{We use the training set as in-context examples for LLM inference on the test set.}

\begin{table*}
\centering
\resizebox{\linewidth}{!}{%
    \begin{tabular}{l p{0.8\textwidth} p{0.3\textwidth} c} % Adjust column widths
    \toprule
    \textbf{Design Type} & \textbf{User Query} & \textbf{Images \& Captions} & \textbf{\# Train/Test} \\
    \toprule
    \textbf{Book Cover} & 
    \begin{minipage}{0.8\textwidth}
        Please help me create a travel guide cover design titled `Exploring Tuscany' with a blue sky background, featuring rolling vineyards in the background below a large Tuscan villa centered on the lower half. The title `Exploring Tuscany' should be at the top center in dark green, with `Travel Guide' below the title and `Discover the Heart of Italy' below the author, both in dark green.
    \end{minipage} & 
    \begin{minipage}{0.3\textwidth}
        \centering
        \makebox[\textwidth]{ % Ensures both images fit inside the column
            \includegraphics[width=0.45\textwidth]{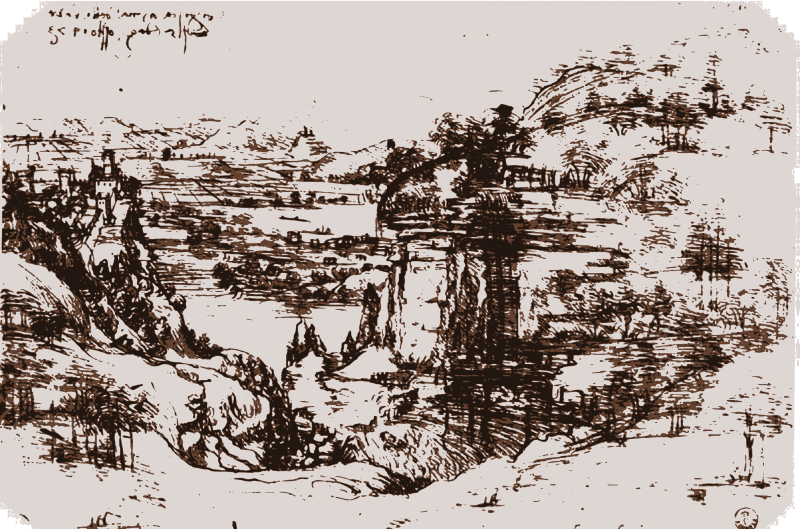} 
            \hspace{0.5em} % Adds small space between images
            \includegraphics[width=0.45\textwidth]{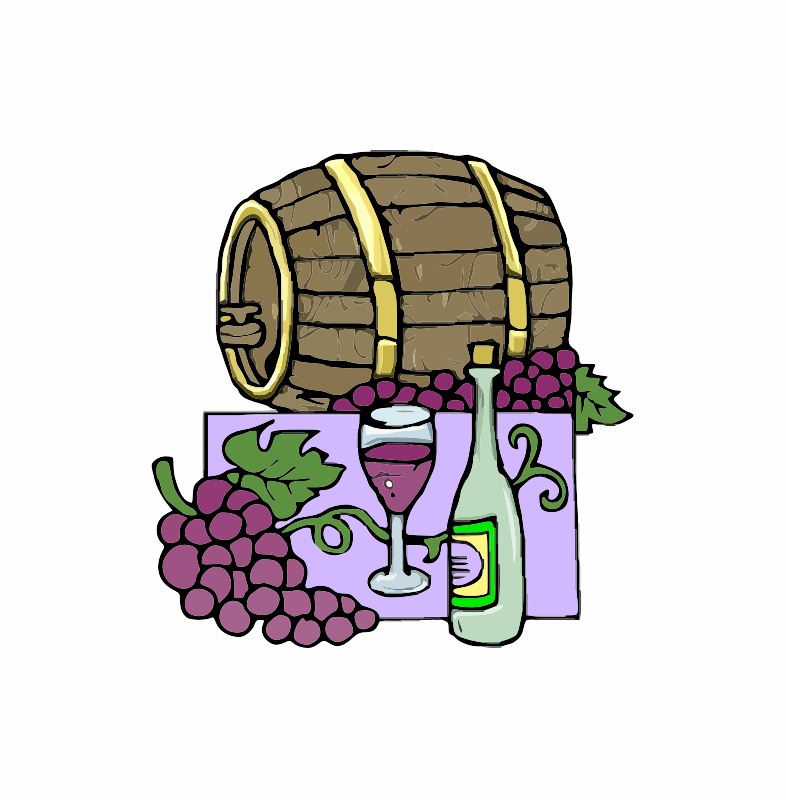} 
        } \\
        \makebox[\textwidth]{ % Ensures captions stay aligned
            \scriptsize{\parbox{0.48\textwidth}{\centering (1) Study of a \\ Tuscan Landscape}} 
            \hspace{1.5em} 
            \scriptsize{(2) A Winery}
        }
    \end{minipage} & 5/260 \\
    \midrule

    \textbf{Business Card} & 
    \begin{minipage}{0.8\textwidth}
        Create a business card design in teal background for a software company named `OceanSoft'. Please include the company name in huge white font at the top center. I want to include the tagline `Sailing to Success' in medium white font placed below the company name and a large wave icon at the bottom center.
    \end{minipage} & 
    \begin{minipage}{0.3\textwidth}
        \centering
        \makebox[\textwidth]{ % Ensures both images fit inside the column
            \includegraphics[width=0.32\textwidth]{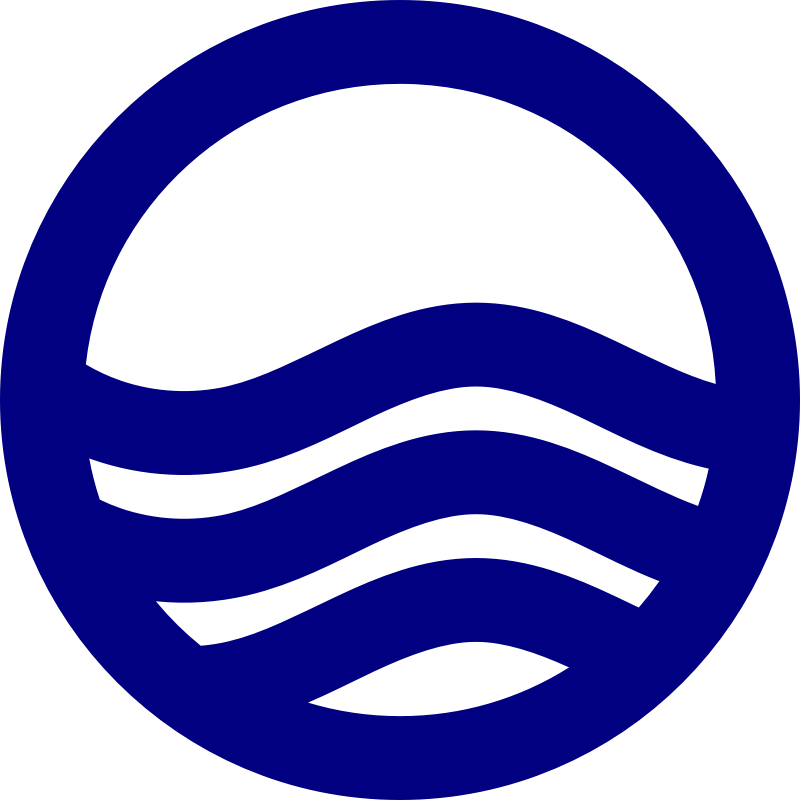} 
            % \hspace{0.5em} % Adds small space between images
            % \includegraphics[width=0.4\textwidth]{figures/images/bookcover2.png} 
        } \\
        \makebox[\textwidth]{ % Ensures captions stay aligned
            \scriptsize{(1) A blue wave icon} 
            % \hspace{1.5em} 
            % \scriptsize{(2) A Winery}
        }
    \end{minipage} & 5/203 \\
    \midrule

    \textbf{Postcard} & 
    \begin{minipage}{0.8\textwidth}
        Kindly assist in creating an Easter postcard design with the message `Happy Easter!' centered at the top in pastel pink on a lavender background, featuring medium-sized decorated Easter eggs at the bottom left and bottom right, and a large bunny illustration centered below the message.
    \end{minipage} & 
    \begin{minipage}{0.3\textwidth}
        \centering
        \makebox[\textwidth]{ % Ensures both images fit inside the column
            \includegraphics[width=0.4\textwidth]{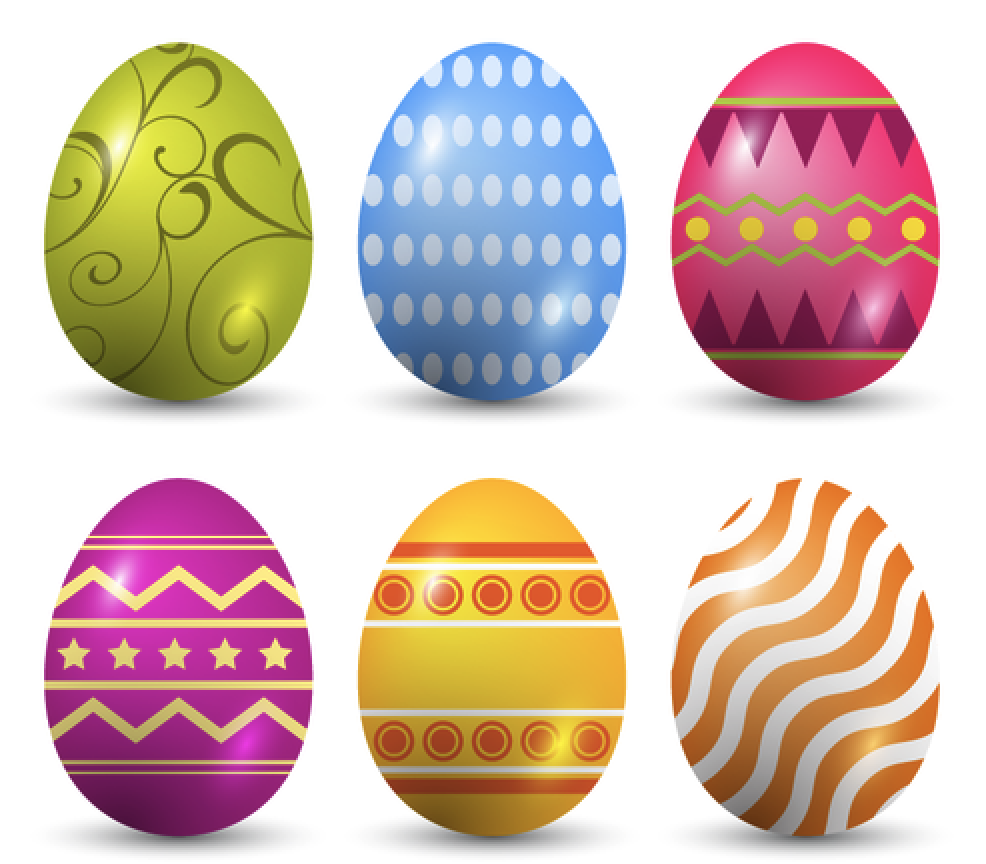} 
            \hspace{2em} % Adds small space between images
            \includegraphics[width=0.14\textwidth]{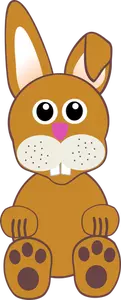} 
        } \\
        \makebox[\textwidth]{ % Ensures captions stay aligned
            \scriptsize{(1) Easter eggs} 
            \hspace{1.5em} 
            \scriptsize{\parbox{0.48\textwidth}{\centering (2) Easter bunny \\ illustration}}
        }
    \end{minipage} & 5/260 \\
    \midrule

    \textbf{Poster} & 
    \begin{minipage}{0.8\textwidth}
        Could you provide me with a plan to create a poster design for an art workshop on a beige background featuring a large abstract painting in the center? Place the title `MASTERING MODERN ART' in huge bold black font at the top center, and the registration details `Register now at www.artworkshop.com or call 123-456-7890' in small dark brown font at the bottom left.
    \end{minipage} & 
    \begin{minipage}{0.3\textwidth}
        \centering
        \makebox[\textwidth]{ % Ensures both images fit inside the column
            \includegraphics[width=0.36\textwidth]{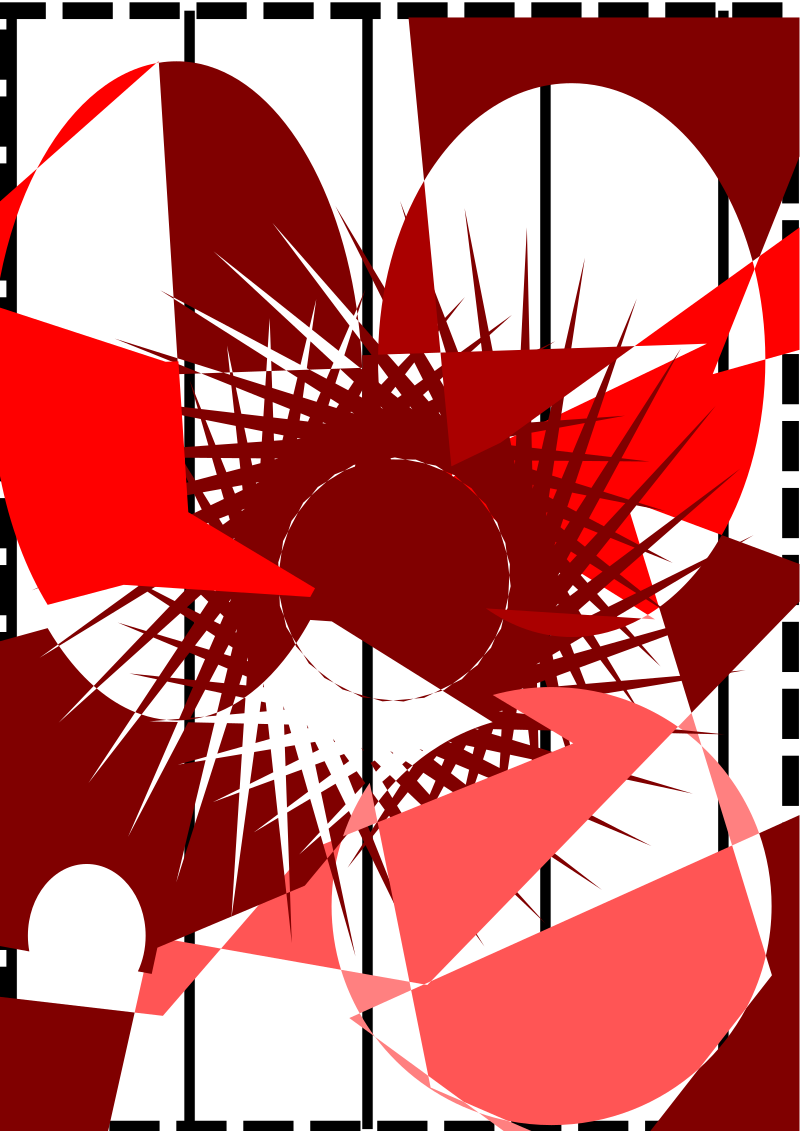} 
            % \hspace{2em} % Adds small space between images
            % \includegraphics[width=0.14\textwidth]{figures/images/postcard2.png} 
        } \\
        \makebox[\textwidth]{ % Ensures captions stay aligned
            \scriptsize{(1) Abstract art in red and black} 
            % \hspace{1.5em} 
            % \scriptsize{\parbox{0.48\textwidth}{\centering (2) Easter bunny \\ illustration}}
        }
    \end{minipage} & 5/336 \\

    \bottomrule
    \end{tabular}
}
\caption{Examples of the four design types in \textsc{GraphicBench}. \textbf{Images \& Captions:} Input images with associated captions. Appendix \ref{appendix:concept_distribution} provides the distribution of design concepts.}
\label{tab:creativebench_examples}
\end{table*}

\paragraph{Reference Plan Annotation.}
We first generate human-annotated user queries and plans for five instances per design type, totaling 20 pairs serving as the training set. To ensure the dataset reflects real-world design needs, we collect screenshots of various design projects shared on the Behance platform\footnote{\url{https://www.behance.net/}}, created by designers using Adobe Creative Cloud (CC) design tools.\footnote{\url{https://www.adobe.com/creativecloud}} Using each screenshot as a reference, we invite three graduate students with experience in Adobe CC tools to collaboratively craft realistic user queries, write workflows, and execute each step in the workflow to produce a final design resembling the reference screenshot. Through this process, we identify key design components associated with each design type, as detailed in Appendix Table \ref{tab:design_components}. Examples of human-annotated user queries, corresponding workflows, and the design outputs are further detailed in Appendix \ref{appendix:reference_plan_annotation}.

\paragraph{Query Construction.}
We incorporate the identified design components as placeholders to form the skeleton of user queries for each design type, which serve as prompt templates \citep{qian-etal-2024-tell, xie2024travelplanner, yoran2024assistantbenchwebagentssolve}. For each design type, we prompt \textsc{GPT-4} \citep{achiam2023gpt} to randomly populate the design components in the skeleton queries, as shown in Appendix \ref{appendix:cb_prompts}. Subsequently, we manually give variations in query headers (e.g., ``\textit{Please help me create a design}'', ``\textit{Could you provide me a design}'') to capture diverse phrasing styles in user queries, as illustrated in Table \ref{tab:creativebench_examples}.

\paragraph{Diversity Check.}
We observe that directly using the queries generated by GPT-4 presents a challenge, as many queries tend to share similar design concepts (i.e., multiple postcard queries are related to ``Happy Birthday'', differing only in trivial aspects such as color choice). To ensure the diversity of the generated user queries and their associated design components, we perform the following steps:
\begin{enumerate}[leftmargin=*, itemsep=2pt, parsep=-1pt]
    \item Discard redundant queries with a bi-gram match in any of the design components.
    \item Discard highly similar queries with a semantic similarity above 0.8, measured using \textsc{SentenceBERT} \citep{reimers-gurevych-2019-sentence}.
\end{enumerate}

\paragraph{Image Pairing.}
Each validated user query includes a short description of the image(s) needed in their design, as shown in Table \ref{fig:main_figure}. To map each image description to an image file, we first compile a search pool by collecting images from OpenCLIPArt\footnote{\url{https://openclipart.org/}} and Public Domain Vectors.\footnote{\url{https://publicdomainvectors.org/}} Both platforms offer a large collection of vector illustrations suitable for graphic design. We collect 179K and 95K image URL-caption pairs from the websites, forming a 274K image pool for retrieval. From this pool, we retrieve the top-3 images with the highest semantic similarity between the image description in the query and the collected captions, using \textsc{SentenceBERT} for similarity scoring.\footnote{We will release the images under the Creative Commonsense Zero (CC0) license.}

\paragraph{Human-LLM Evaluation.}
We begin with an automatic evaluation to assess the quality of user queries and the top-3 retrieved images. For each query, we prompt \textsc{GPT-}o1\footnote{\url{https://openai.com/o1/}} to: \textbf{1)} identify key design components and rate how well each contributes to the overall coherence of the final design on a 5-point Likert scale (1: Not aligned at all, 5: Completely aligned), and \textbf{2)} rank the three image candidates from 1 (best fit) to 3 (least fit) based on their relevance to the query. To validate the automatic evaluation, we conduct a manual study on a stratified random sample of 200 user queries, with 50 from each design type. Each annotator reviews 25 queries and answers the same questions. Given high inter-annotator agreement between \textsc{GPT-}o1 and human annotations (Cohen's Kappa\footnote{\url{https://en.wikipedia.org/wiki/Cohens_kappa}} is 0.586 for the first question and Kendall's $\tau$\footnote{\url{https://en.wikipedia.org/wiki/Kendall_rank_correlation_coefficient}} is 0.671 for the second question), we rely on \textsc{GPT-}o1 annotations for filtering. We discard queries that receive a rating of 1 or 2 for any design components and retain only the image ranked as best fit. Further details on the annotation setup are provided in Appendix \ref{appendix:human_quality_check}.
\section{\textsc{GraphicTown}}
\label{sec:creativetown}

% Overview
We present an overview of the \textsc{GraphicTown} framework in Figure \ref{fig:main_figure}, which consists of six key steps: \textbf{1)} generating a design outline based on the user query and image captions from \textsc{GraphicBench}, \textbf{2)} recruiting experts, \textbf{3)} generating workflow plans, \textbf{4)} integrating individual workflows into a cohesive plan, \textbf{5)} retrieving appropriate action for each step, and \textbf{6)} executing the plan. All prompts are detailed in Appendix \ref{appendix:ct_prompts}.

% Step 1
\paragraph{Design Outline.}
% \label{sec:2.1}
The first step in effective design involves users providing clear and precise outcome specifications to ensure that generated outputs align with their needs \citep{design_principles}. These specifications serve as a foundational framework for guiding subsequent design stages \citep{li2024specllmexploringgenerationreview, ma2025specgenautomatedgenerationformal}. However, user queries might often be vague or lack detail in practice \citep{qian-etal-2024-tell}. To this end, a particular LLM agent $M_s$ is prompted as the ``supervisor'' to first craft a design outline based on the user query and fill in any missing design components identified during the reference plan annotation process (Appendix Table \ref{tab:design_components}). $M_s$ infers unspecified information autonomously, ensuring all necessary details are established before proceeding to the planning phase.

% This increases the risk of ``\textit{fake success}'' during agent task execution, where the system appears to complete the task but ultimately deviates from the user's true intent. Ideally, an interactive process would allow the system to clarify ambiguities through user feedback. However, in our study, we limit our analysis to a static setting where the LLM agent infers and fills in missing details by generating a design outline that consolidates necessary details of the design components for generating a final design output.

% Step 2
\paragraph{Expert Recruitment.}
% \label{sec:2.2}
For each user query, $M_s$ forms an expert group $\mathcal{M}$ based on the design outline and predefined expert descriptions, and assigns a high-level goal to each expert agent in $\mathcal{M}$. Specifically, we introduce three design expert agents, each with distinct expertise and responsibilities aligned with Adobe CC design tools, which are widely used by professional designers \citep{tacitknowledge, yuan2024humanaisynergyuidesign}.\footnote{We consider the three most commonly used Adobe CC design tools among designers on the Behance platform (\url{https://www.behance.net/}).} Each expert agent is outlined below, with detailed descriptions in Appendix \ref{prompt:step2}:

\begin{itemize}[leftmargin=*, itemsep=2pt, parsep=-1pt]
    \item \includegraphics[height=1.0em]{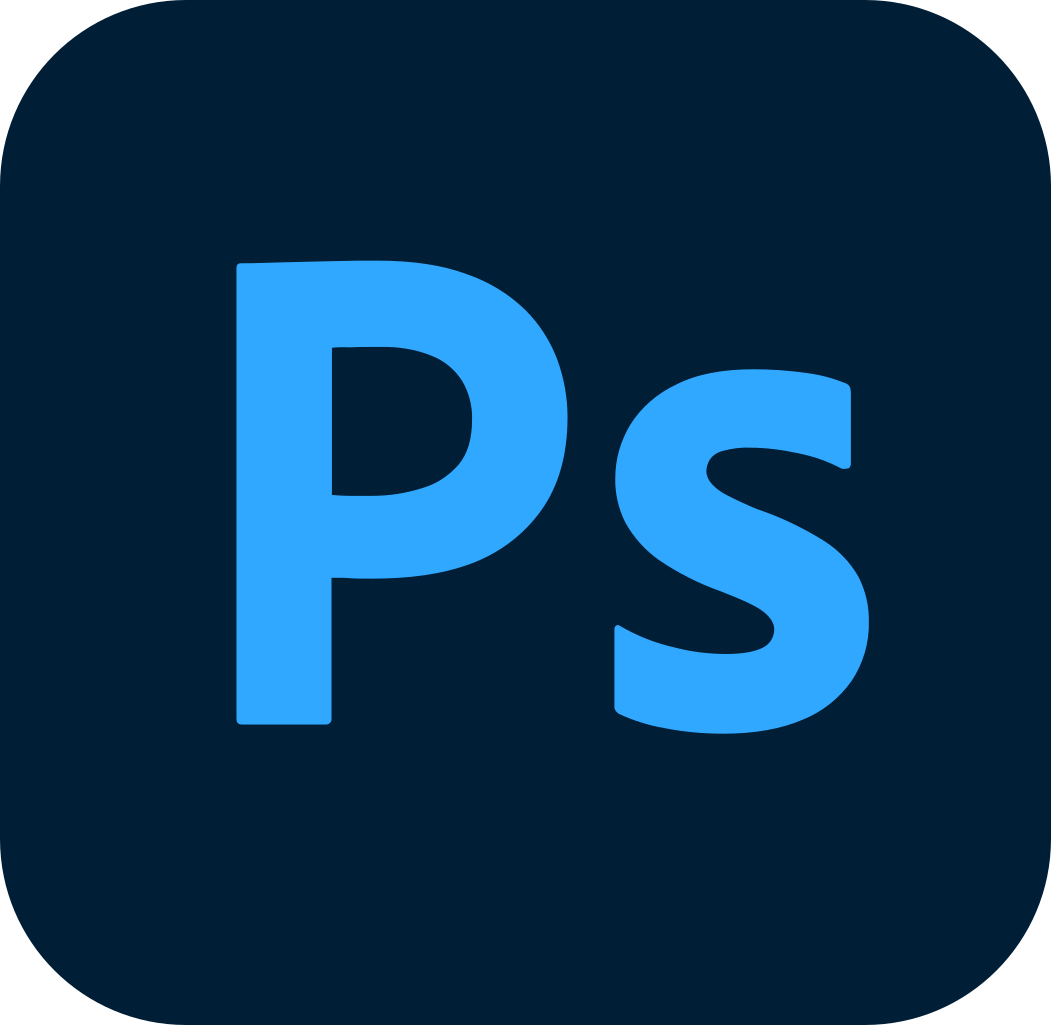} \textbf{Photo Editor:} An agent with an expertise in Adobe Photoshop,\footnote{\url{https://www.adobe.com/products/photoshop}} responsible for image editing, color correction, and applying filters.
    \item \includegraphics[height=1.0em]{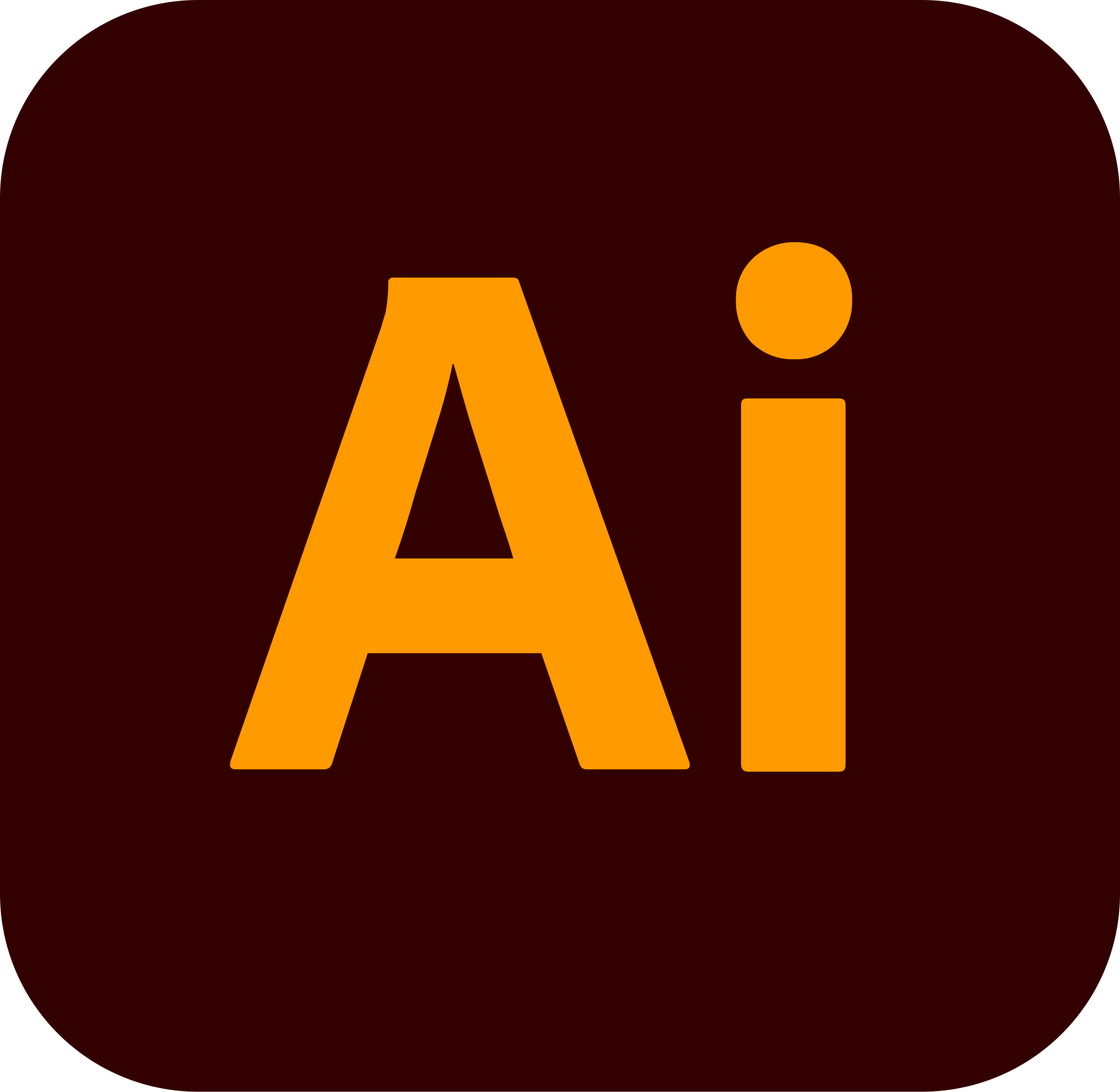} \textbf{Vector Graphic Editor:} An agent with an expertise in Adobe Illustrator,\footnote{\url{https://www.adobe.com/products/illustrator}} focused on creating and editing vector illustrations.
    \item \includegraphics[height=1.0em]{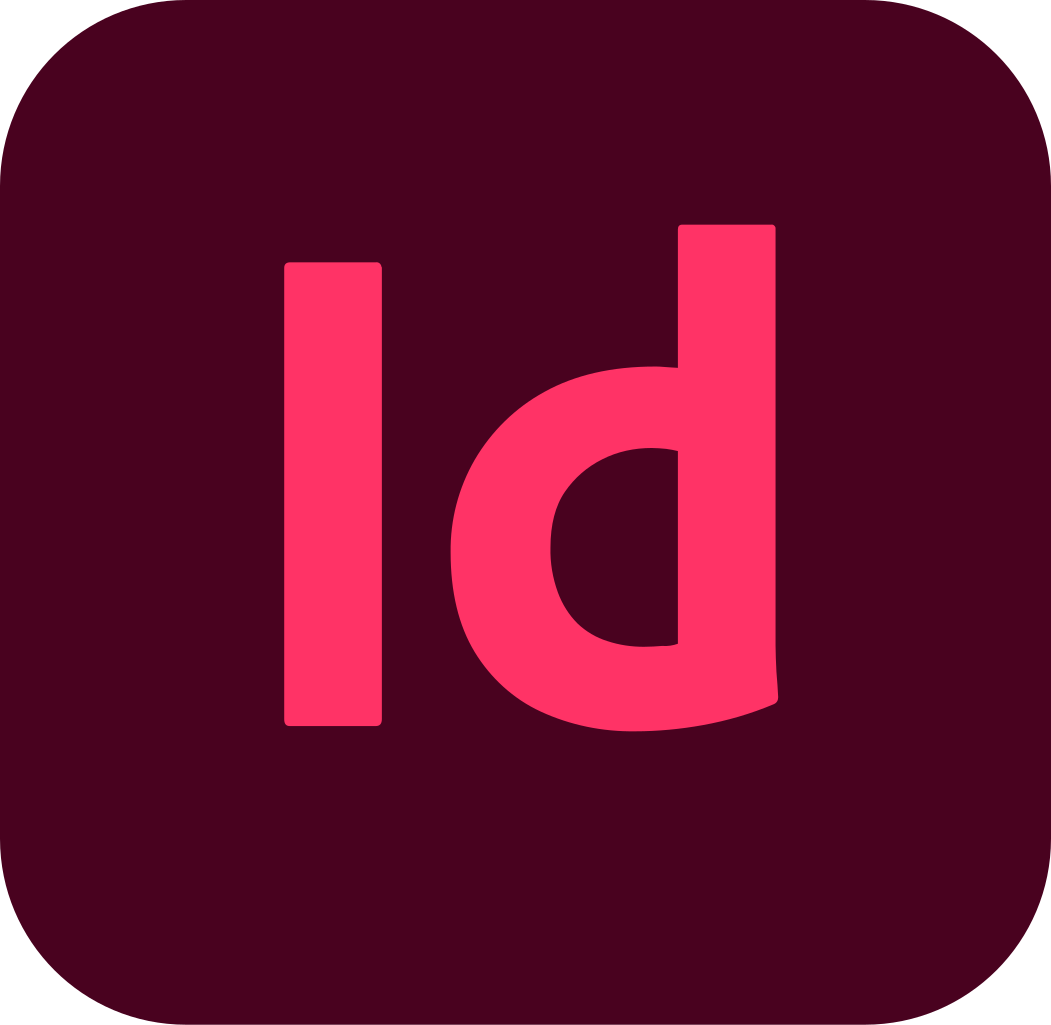} \textbf{Layout Designer:} An agent with an expertise in Adobe InDesign,\footnote{\url{https://www.adobe.com/products/indesign}} responsible for customizing layout templates, exporting files, and integrating text with visual elements.
\end{itemize}

% Step 3
\paragraph{Workflow Generation.}
% \label{sec:2.3}
Planning a design workflow is inherently a \textbf{long-horizon} task, requiring a large number of sequential steps to complete a single design. To this end, instead of generating the entire workflow plan in one step, we distribute the process across agents in the recruited expert group, denoted as $M_i \in \mathcal{M}$. Each $M_i$ plans its own workflow $W_i$ based on the design outline and assigned high-level goal. To emulate human problem-solving process \citep{zhu2023ghostminecraftgenerallycapable}, $M_i$ is instructed to decompose its high-level goal into a sequence of actionable sub-goals \citep{yang2024selfgoallanguageagentsknow, wu2024oscopilotgeneralistcomputeragents, zheng2025mcuevaluationframeworkopenended}, which further facilitates accurate retrieval of actions \citep{huang-etal-2024-planning}.

% (e.g., ``\textit{Create a deep purple night sky background with a large dreamy moon centered.}'' for the Photo Editor agent)
% For instance, the Photo Editor agent might break down its goal into a sequence of steps as follows:
% \begin{enumerate}[leftmargin=*, itemsep=2pt, parsep=-1pt]
%     \item Create a new document with the specified book cover dimensions.
%     \item Set the background color to deep purple.
%     \item Import the dreamy moon image.
%     \item Resize the dreamy moon image to make it large.
%     \item Center the dreamy moon image.
% \end{enumerate} 

% Step 4
\paragraph{Workflow Supervision.}
% \label{sec:2.4}
Since the workflow generation step is conducted independently for each $M_i$, dependencies between agents are not explicitly considered. Simply aggregating individual workflow plans can lead to issues such as: 1) multiple agents might perform the same task or 2) when one agent relies on files generated by another, file names may be inconsistently used. To address these challenges, we adopt a \textbf{hierarchical} agentic structure, where a lead agent ($M_s$) directs one or more specialized agents ($M_i \in \mathcal{M}$) to perform tasks as needed by independently communicating with them \citep{ahilan2019feudal, guo2024largelanguagemodelbased, fourney2024magenticonegeneralistmultiagentsolving, zhang-etal-2025-planning}. Therefore, $M_s$ integrates the individual workflows $W_i$ from each $M_i$ into a single cohesive workflow $W_s$, ensuring that interdependencies within and between agents are properly resolved.

% Step 5, 6
\paragraph{Action Retrieval+Execution.}
% \label{sec:2.5}
For each step in $W_s$, $M_s$ retrieves an appropriate action and infers parameter values to generate $W_r$ for execution within the Adobe CC scripting environment.\footnote{Adobe CC design tools do not support direct API calls.} As shown in Table \ref{tab:action_list}, we define 46 available actions across the three expert agents, categorized into four basic operations, 13 drawing functions, 11 text-related functions, and 18 object manipulation functions.\footnote{We identify common actions based on Adobe's tutorial videos (\url{https://www.adobe.com/learn}).} Each agent has access to a subset of these actions, with each action corresponding to a single mouse or keyboard operation (e.g., \textit{Create a new document}) \citep{he-etal-2024-webvoyager} and is linked to an executable JavaScript code that takes a list of parameter values.\footnote{Since current models struggle to generate executable code directly from long, complex plans \citep{ge2024recursivevisualprogramming}, we provide manually written JavaScript codes that only require parameter values as inputs. Parameter keys for each action are defined based on Adobe's scripting guides. We show an example of JavaScript code snippet in Appendix Figure \ref{fig:javascript}.} When $W_r$ involves multiple expert agents, actions are executed sequentially within their respective environments, leading to the final design outcome $D$.

\begin{table*}
\centering
\resizebox{\linewidth}{!}{%
    \begin{tabular}{llllll}
    \toprule
    \textbf{Category} & \textbf{Action} & \textbf{Parameters} & \textbf{Description} & \textbf{Experts}\\
    \toprule
    \multirow{3}{*}{\textbf{Basic}} & \texttt{CreateDocument} & \texttt{docType} & Create new document with pre-defined dimensions. & \includegraphics[height=1.0em]{figures/logo/photoshop.png} \includegraphics[height=1.0em]{figures/logo/illustrator.png} \includegraphics[height=1.0em]{figures/logo/indesign.png} \\

    & \texttt{SetBackgroundColor} & \texttt{red, green, blue} & Set the background color to desired RGB color. & \includegraphics[height=1.0em]{figures/logo/photoshop.png} \includegraphics[height=1.0em]{figures/logo/illustrator.png} \includegraphics[height=1.0em]{figures/logo/indesign.png} \\

    & \texttt{SaveDocument} & \texttt{fileName, format} & Save the current document into desired format. & \includegraphics[height=1.0em]{figures/logo/photoshop.png} \includegraphics[height=1.0em]{figures/logo/illustrator.png} \includegraphics[height=1.0em]{figures/logo/indesign.png} \\
    \midrule

    \multirow{3}{*}{\textbf{Drawing}} & \texttt{DrawCircle} & \texttt{layerName, radius, red, green, blue} & Draw a circle of desired radius and RGB color. & \includegraphics[height=1.0em]{figures/logo/illustrator.png} \\

    & \texttt{OpacityDrawing} & \texttt{layerName, opacity} & Adjust opacity of a drawing. & \includegraphics[height=1.0em]{figures/logo/illustrator.png} \\

    & \texttt{ResizeDrawing} & \texttt{layerName, width, height} & Resize a drawing to desired width and height. & \includegraphics[height=1.0em]{figures/logo/illustrator.png} \\
    \midrule

    \multirow{3}{*}{\textbf{Text}} & \texttt{CreateText} & \texttt{layerName, textString} & Create a new text (default to Arial font). & \includegraphics[height=1.0em]{figures/logo/photoshop.png} \includegraphics[height=1.0em]{figures/logo/illustrator.png} \includegraphics[height=1.0em]{figures/logo/indesign.png} \\

    & \texttt{ApplyFont} & \texttt{layerName, fontName} & Apply font to text. & \includegraphics[height=1.0em]{figures/logo/photoshop.png} \includegraphics[height=1.0em]{figures/logo/illustrator.png} \includegraphics[height=1.0em]{figures/logo/indesign.png} \\

    & \texttt{RotateText} & \texttt{layerName, angle} & Rotate text to desired angle. & \includegraphics[height=1.0em]{figures/logo/photoshop.png} \includegraphics[height=1.0em]{figures/logo/illustrator.png} \includegraphics[height=1.0em]{figures/logo/indesign.png} \\
    \midrule

    \multirow{3}{*}{\textbf{Object}} & \texttt{ImportObject} & \texttt{fileName, layerName} & Import an image or object from file path. & \includegraphics[height=1.0em]{figures/logo/photoshop.png} \includegraphics[height=1.0em]{figures/logo/illustrator.png} \includegraphics[height=1.0em]{figures/logo/indesign.png} \\

    & \texttt{GenerateQRObject} & \texttt{layerName, linkURL} & Generate a QR code with desired URL embedded. & \includegraphics[height=1.0em]{figures/logo/indesign.png} \\

    & \texttt{PhotoFilter} & \texttt{layerName, filterType, density} & Apply a photo filter to an object with desired density. & \includegraphics[height=1.0em]{figures/logo/photoshop.png} \\

    \bottomrule
    \end{tabular}
}
\caption{Actions in \textsc{GraphicTown}. Each action requires specific parameters for execution. \textbf{Experts:} The expert(s) which supports the execution of a specific action. The complete list of 46 available actions is provided in Table \ref{tab:action_full_list} of Appendix.}
\label{tab:action_list}
\end{table*}

\section{Experiment Setup}

\subsection{Models}
\label{sec:models}
We evaluate the design planning abilities of various LLMs on \textsc{GraphicBench} using the \textsc{GraphicTown} framework. Due to the extensive textual information involved in the planning process, we limit our evaluation to LLMs capable of processing inputs exceeding 8K in length. We benchmark five open-weights models across different model sizes and families: \textsc{LLaMA-3.1 8b} \citep{grattafiori2024llama3herdmodels}, \textsc{Gemma-2 9b} and \textsc{27b} \citep{gemmateam2024gemma2improvingopen}, and \textsc{Qwen-2.5 7b} and \textsc{14b} \citep{qwen2025qwen25technicalreport}, and one closed-source model: GPT-3.5.\footnote{\url{https://openai.com/index/chatgpt/}} For open-weights models, we set the sampling temperature to 0.0.\footnote{HuggingFace model names for open-weights models are listed in Appendix Table \ref{tab:huggingface_api}.}

% We exclude \textsc{GPT-4} from our model baseline since this is used as part of the construction process of CreativeBench.
% All experiments are conducted in a one-shot setting, using a human-annotated reference plan as in-context example.

\subsection{Evaluation Metrics}
\label{sec:evaluation}
To ensure a comprehensive evaluation of workflow plans $W_r$ and execution outcomes $D$, we assess them across multiple dimensions. Detailed prompts are provided in Appendix \ref{appendix:eval_prompts}.

\paragraph{Workflow Evaluation.}
To evaluate $W_r$, we define four evaluation criteria:
\begin{itemize}[leftmargin=*, itemsep=2pt, parsep=-1pt]
    \item \textbf{Delivery Rate:} This metric measures whether agents can successfully deliver a workflow within a limited number of steps. The step limit is determined by the difficulty level, based on the number of expert agents involved: \textbf{1)} Easy: 1 expert, max 10 steps; \textbf{2)} Medium: 2 experts, max 20 steps; \textbf{3)} Hard: 3 experts, max 30 steps.\footnote{The maximum number of steps is determined by the average steps in human-annotated workflows.} Workflow that fall into dead loops or exceed the step limit are considered as failures \citep{xie2024travelplanner}.
    \item \textbf{Design Pass Rate:} This metric assesses whether agents can correctly incorporate both explicit design components specified in the user query and implicit commonsense constraints. We prompt \textsc{GPT-4} to provide a score from 1 to 5 for each of the three aspects: color, text, and images.
    \item \textbf{Step Efficiency:} We measure the ratio of non-duplicate steps to the total number of steps.
    \item \textbf{Expert Use Efficiency:} Since a single workflow might involve multiple expert agents, we define efficiency as minimizing the frequency of switching between expert agents. A higher efficiency score indicates fewer transitions and better expert utilization. Formally, for a workflow $W$ with $N$ steps and $E$ unique experts:
\end{itemize}
\begin{equation}
    \mathbf{ExpertUseEff.}(W) = \frac{E-1}{\sum^{N}_{i=1} \mathds{1}(\mathrm{expert}_i \neq \mathrm{expert}_{i-1})}
\end{equation}

\paragraph{Execution Evaluation.}
To evaluate $D$, we define five evaluation criteria:
\begin{itemize}[leftmargin=*, itemsep=2pt, parsep=-1pt]
    \item \textbf{Execution Success Rate:} This metric measures the success rate of execution attempts, calculated as the percentage of successful executions out of the total executions performed.
    \item \textbf{Fidelity:} A successful design should correctly incorporate the input images specified in the user query. We quantify fidelity using template matching from \texttt{opencv} library.\footnote{\url{https://docs.opencv.org/4.x/d4/dc6/tutorial_py_template_matching.html}}
    \item \textbf{Content Similarity:} We measure the semantic similarity between the user query and the execution outcome using \textsc{CLIPScore} \citep{hessel-etal-2021-clipscore}.
    \item \textbf{VQA Pass Rate:} We measure whether the execution outcome aligns with the design components in the user query using Visual Question Answering (VQA) \citep{agrawal2016vqavisualquestionanswering}. We generate questions for each query by prompting \textsc{GPT-4}.\footnote{On average, 9.07, 10.0, 7.89, 8.70 questions are generated per user query for book covers, business cards, postcards, and posters, respectively. Examples of questions are detailed in Appendix Table \ref{tab:vqa_questions}.} We use a recent multimodal model \textsc{LLaVA-1.5 7b} \citep{llava} to generate answers as Yes or No \citep{zhao2024lova}. The final pass rate is the average accuracy across all questions.
    \item \textbf{Creativity:} Following \citet{torrance1966torrance, runco2012standard, assessing_creativity}, we assess the creativity of our design outcomes along two axes: \textbf{1)} Originality, which measures the uniqueness of the design, and \textbf{2)} Elaboration, which measures the extent to which the design expands on the information in the user query by adding meaningful details. We prompt \textsc{GPT-}o1 to provide a score from 1 to 5 for each axis.
\end{itemize}

% Following \citet{si2025design2codebenchmarkingmultimodalcode, ge2025autopresentdesigningstructuredvisuals}, 
\section{Results}

In this section, we discuss the performance of various LLMs in terms of planning design workflows and the executed design outcomes. We highlight several main findings below:

\begin{wrapfigure}{r}{0.45\textwidth}  % 'r' for right alignment
    \centering
    \includegraphics[width=0.85\linewidth]{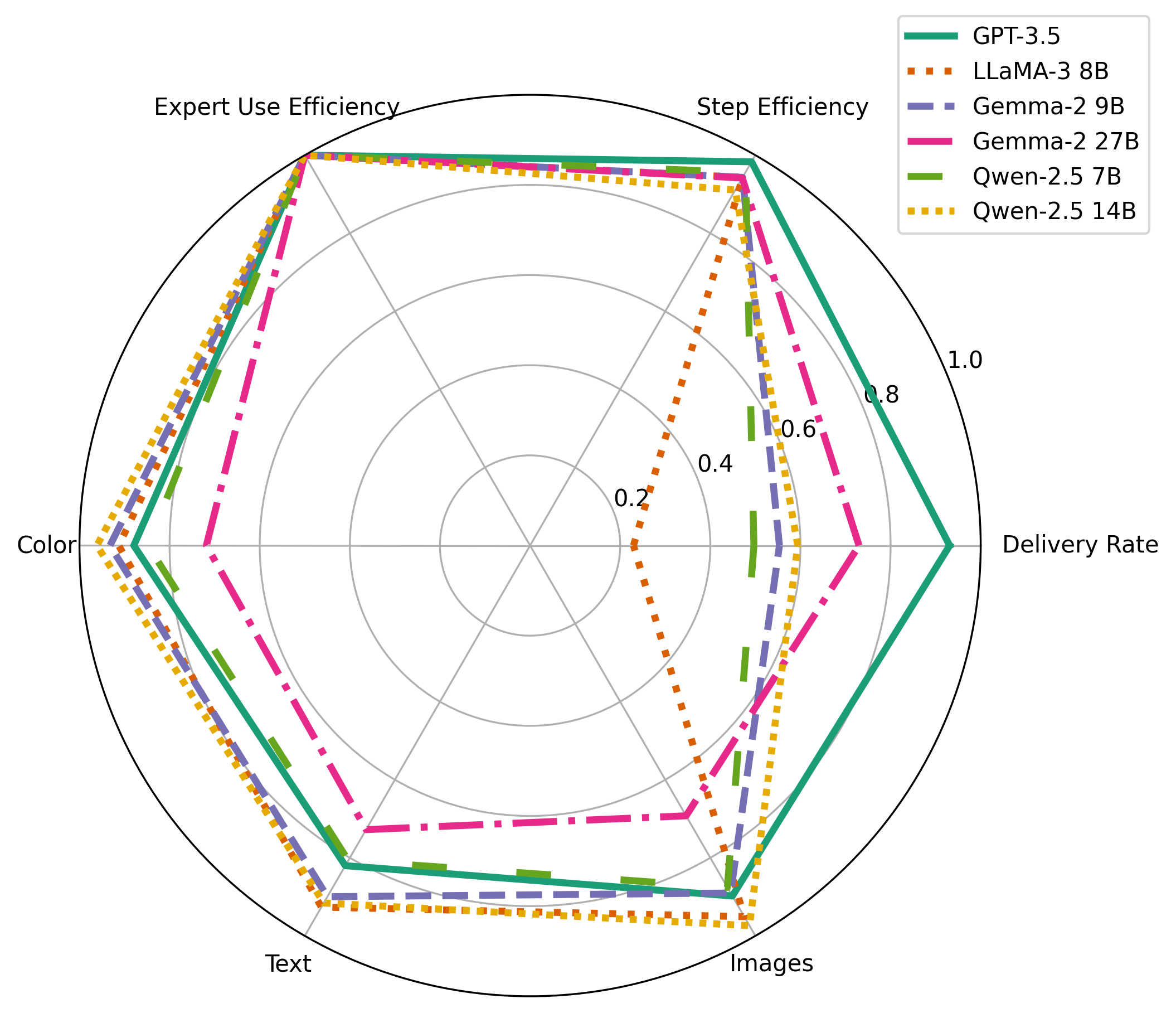}
    \caption{Workflow evaluation results for each model. We normalize color, text, and images pass rates to [0,1].}
    \label{fig:main_results}
\end{wrapfigure}

\paragraph{LLM agents can generate workflow plans that meet design constraints.}
As shown in Figure \ref{fig:main_results}, all tested models plan for design workflows that efficiently utilize expert agents, achieving an average efficiency score of 1.0 and a high average step efficiency of 0.947. Interestingly, larger models are not the top performers in terms of design pass rate. \textsc{GPT-3.5} and \textsc{Gemma-2 27b}, despite their size, exhibit relatively lower performance across all three design aspects compared to smaller models. In contrast, larger models outperform their smaller \textsc{7-9b} counterparts in terms of delivery rate. Overall, the models demonstrate strong performance across tested workflow evaluation metrics, indicating that their generated plans efficiently utilize expert agents, decompose high-level goals to distinct steps, and well-incorporate both explicit design constraints from user queries and implicit commonsense constraints. Full numerical results by model and design type are provided in Appendix \ref{appendix:detailed_res1}.

\paragraph{Specific expert agent sequences are preferred.}
On average, 2.05 expert agents are recruited per user query. All LLMs, except \textsc{Qwen-2.5 7b}, predominately use the Photo Editor and Layout Designer combination during planning, regardless of design type. This is likely due to the complementary expertise of these two agents and the specialized role of the Vector Graphic Editor, which focuses on creating vector illustrations. Since user queries in \textsc{GraphicBench} already have associated input images, only few require generating from scratch. The workload distribution also varies across expert agents, with the Layout Designer handling the most workflow steps on average (12.2), followed by the Vector Graphic Editor (9.87), and the Photo Editor (7.92). Additionally, LLMs tend to follow a preferred sequence of agents in their plans, most commonly Photo Editor → Layout Designer. Detailed results are provided in Appendix \ref{appendix:detailed_res2}.

\paragraph{Specific type and sequence of actions are preferred.}
In Appendix \ref{appendix:detailed_res3}, we present the distribution of retrieved actions across models. We observe that each expert agent tends to use only a limited set of actions, despite having access to a broader range: the Photo Editor agent primarily performs object manipulation (e.g., \texttt{ImportObject}, \texttt{ResizeObject}), while the Layout Designer agent frequently applies text-related operations (e.g., \texttt{AlignText}, \texttt{ColorText}). 
We further show that the most common action sequences in planned workflows closely mirror human-annotated workflows (Appendix \ref{appendix:reference_plan_annotation}), typically starting with document creation, setting the background color, importing and manipulating images, and concluding with text modifications. This suggests that prior findings that models resemble human reasoning process \citep{wei2023chainofthoughtpromptingelicitsreasoning} extend to more open-ended, creative tasks such as graphic design generation.

\paragraph{Planned workflows do not lead to successful design outcomes.}
We detail the execution results in Table \ref{tab:execution_results}, which shows that despite high scores on workflow evaluation metrics, the resulting design workflows actually fail to produce successful design outcomes. In most cases, models correctly import the required images but misplace them, frequently causing overflow beyond document boundaries, which contributes to low fidelity rate. This aligns with prior findings that LLMs struggle with spatial reasoning and object positioning within a given space \citep{yamada2024evaluating, wu2024minds}. Content similarity and VQA pass rates are also generally low, with \textsc{Gemma-2 27b} outperforming other models on both metrics, yet still achieving only 20.79 and 39.64, respectively. All models also exhibit similarly low creativity scores in both originality and elaboration, averaging 1.88 and 1.63. This shows that they struggle to introduce novel elements or expand on the provided details. We provide several case studies of failed executions in Appendix \ref{appendix:case_studies}. Taken together, these results suggest that while LLM agents effectively incorporate high-level design constraints in their planned workflows, they often fail to capture finer-grained details, such as spatial relationships between different design components.

\begin{table*}
\centering
\resizebox{\linewidth}{!}{%
    \begin{tabular}{lllllllll}
    \toprule
    \textbf{Model} & \textbf{Success Rate (\%)} & \textbf{Fidelity} & \textbf{Content Similarity} & \textbf{VQA Pass Rate} & \textbf{Creativity (O)} & \textbf{Creativity (E)} \\
    \toprule

    \textsc{\textbf{LLaMA-3.1 8b}} & 85.14 & 0.161 & 20.20 & 39.27 & 1.81 & 1.55 \\
    \textsc{\textbf{Gemma-2 9b}} & 82.23 & \textbf{0.178} & 19.71 & 37.60 & 1.93 & 1.60 \\
    \textsc{\textbf{Gemma-2 27b}} & 57.60 & 0.146 & \textbf{20.79} & \textbf{39.64} & \textbf{2.03} & 1.66 \\
    \textsc{\textbf{Qwen-2.5 7b}} & 46.47 & 0.159 & 19.13 & 39.10 & 1.81 & 1.51 \\
    \textsc{\textbf{Qwen-2.5 14b}} & \textbf{90.51} & 0.169 & 20.16 & 33.13 & 1.74 & 1.46 \\
    \textsc{\textbf{GPT-3.5}} & 69.68 & 0.167 & 19.92 & 27.52 & 1.96 & \textbf{2.00} \\

    \bottomrule
    \end{tabular}
}
\caption{Execution results for each model. Full results per design type are provided in Appendix \ref{appendix:detailed_res4}. Best scores for each column is \textbf{bold}. \textbf{Creativity (O):} Originality, \textbf{(E):} Elaboration.}
\label{tab:execution_results}
\end{table*}

% Fidelity rates remain low across all models, indicating that models fail to incorporate user-provided images into the final execution outputs. 
\section{Error Analysis}
\label{sec:error_analysis}

\begin{figure*}
    \centering
    \includegraphics[width=\linewidth]{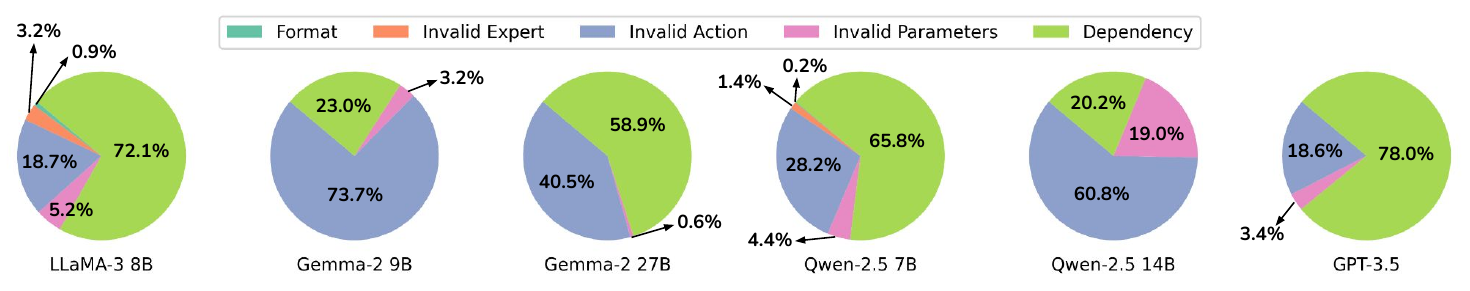}
    \caption{Error distribution per model. Most errors arise from models failing to resolve dependencies or retrieving invalid actions.}
    \label{fig:error_distribution}
\end{figure*}

In this section, we aim to understand the underlying reasons why the planned workflows often fail to produce successful design outcomes. We automatically categorize errors for each step in the workflow into the following types: \textbf{1) Format:} the workflow has formatting issues (e.g., not in proper JSON list format) and cannot be loaded for execution; \textbf{2) Invalid Expert:} the workflow step assigns an invalid expert agent (e.g., Text Editor); \textbf{3) Invalid Action:} the retrieved action is not defined in the available action set (e.g., \texttt{ApplyArialFont}); \textbf{4) Invalid Parameters:} the provided parameter keys do not match the expected inputs for the action (e.g., \texttt{doc} for \texttt{CreateDocument}); \textbf{5) Dependency:} the workflow step cannot be executed due to a broken dependency from a previous step (e.g., attempting to \texttt{ImportObject} a file that has not been previously saved). We show the distribution of error types per model in Figure \ref{fig:error_distribution}, from which we have the following observations:
\begin{enumerate}[leftmargin=*, itemsep=2pt, parsep=-1pt]
    \item The majority of errors stem from dependency issues, accounting for an average of 53.0\% of errors. Dependency errors can be categorized into \textbf{1)} local dependencies, which occur between steps within a single workflow, and \textbf{2)} global dependencies, which occur between expert agents. We observe that all models particularly struggle with handling global dependencies, such as correctly using an expert agent's output as input for the next agent or avoiding redundant steps across different expert agents.
    
    \item In models such as \textsc{Gemma-2 9b} and \textsc{Qwen-2.5 14b}, invalid functions contribute the most to errors. Despite having access to the full list of available actions and parameters during retrieval, agents still fail to use the correct function names in their workflow plans. This motivates further exploration into more reliable action retrieval methods.
    
    \item Other errors, invalid parameters, invalid expert assignments, and formatting issues, have smaller impact, contributing an average of 5.96\%, 0.77\%, and 0.17\% of errors, respectively.
\end{enumerate} 
As a whole, these results highlight the challenges LLM agents face in global planning and retrieving correct actions, which underscores the need for new strategies to enhance multi-step reasoning and dependency resolution during design planning.
\section{Related Work}
\paragraph{LLM-Based Agents.}
Leveraging the strengths of Large Language Models (LLMs), LLM-based agents have demonstrated strong performance in automating human tasks through tool use \citep{schick2023toolformer, qin2023toolllm, shen2023hugginggptsolvingaitasks, qin2024tool, liang2024taskmatrix} and reasoning \citep{yao2022react, shinn2023reflexionlanguageagentsverbal}. Further inspired by human society and the goal of improving work efficiency through collaboration \citep{reilly_phillips, woolley2015collective}, recent research has explored frameworks involving multiple agents \citep{ding2023designgptmultiagentcollaborationdesign, shen2023hugginggptsolvingaitasks, dong2024villageragentgraphbasedmultiagentframework, autoagents}. In particular, studies suggest that assigning specialized roles to agents improves their effectiveness in solving complex tasks \citep{camel, chen2023agentversefacilitatingmultiagentcollaboration, talebirad2023multiagentcollaborationharnessingpower, du2024multiagentsoftwaredevelopmentcrossteam, hong2024metagptmetaprogrammingmultiagent, qian-etal-2024-chatdev}. Similarly, we adapt an LLM-based agentic framework, but for a previously unexplored task in this research space: graphic design generation.

\paragraph{Graphic Design Generation.}
Graphic design is a form of visual art that combines multimodal elements (e.g., images, texts, and vector symbols) to create aesthetic compositions that effectively comunicate the intent of a user query \citep{cheng2024graphicdesignlargemultimodal}. Prior work has explored various design sub-tasks, including layout generation \citep{li2019layoutgan, gupta2021layouttransformer, jiang2023layoutformer++}, typography generation \citep{zhao2018modeling, jiang2019visual}, and colorization \citep{yuan2021infocolorizer, qiu2023multimodal}. However, limited attention has been given to planning the \textit{entire} graphic design process \citep{inoue2024opencolereproducibleautomaticgraphic}, particularly in the context of agents, and our work aims to fill this gap.

% writing support \citep{cowriting, chakrabarty-etal-2022-help},
%brainstorming ideas \citep{kulkarni2023wordworththousandpictures, wang2025aideation}, 

% Recent efforts have integrated LLMs into various stages of the creative design process, including generating standardized design procedures \citep{ding2023designgptmultiagentcollaborationdesign}, designing game components \citep{zhang2023creativeagentsempoweringagents, guo2024lubanbuildingopenendedcreative}, scientific plots \citep{wu2024plot2codecomprehensivebenchmarkevaluating}, animations \citep{keyframer}, presentation slides \citep{peng2024dreamstructunderstandingslidesuser, ge2025autopresentdesigningstructuredvisuals}, websites \citep{si2025design2codebenchmarkingmultimodalcode}, and logos \citep{liu2025logomotionvisuallygroundedcodesynthesis}. 
\section{Conclusion}
We introduce \includegraphics[height=1.0em]{figures/logo/pantone.png} \textsc{GraphicBench}, a benchmark for graphic design generation that evaluates the design planning abilities of LLM agents (\S \ref{sec:creativebench}). We further present \textsc{GraphicTown}, an LLM agent framework designed to emulate human group planning process for creative design tasks (\S \ref{sec:creativetown}). Our evaluation with six LLMs show that while models can plan for workflows that incorporate both explicit and implicit design constraints, these planned workflows fall short in \textbf{(1)} understanding spatial relationships and positioning of design components, \textbf{(2)} recognizing global dependencies between expert agents, and \textbf{(3)} retrieving appropriate actions at each step. We envision \textsc{GraphicBench} as a valuable stepping stone for future work on enhancing design planning and reasoning in LLM agents.
\section{Limitations}

\textsc{GraphicBench} assumes a scenario in which user queries explicitly specify the text and image content, as well as the precise attributes such as color and text position. However, in realistic settings, users may not always specify or even know exactly what images to include in a design, or they may express their requests at a very high-level \citep{ge2025autopresentdesigningstructuredvisuals}. Future works can explore scenarios where user input is limited, requiring models to seek clarification or request additional details through interactions with users \citep{qian-etal-2024-tell, li2024mediq}. Additionally, the number of actions available for \textsc{GraphicTown} agents is currently limited to a fixed set of 46, as each corresponding JavaScript code was manually written by the authors. This set is not exhaustive of all possible actions within the Adobe CC scripting environment. Future works could investigate automated methods for dynamically generating and retrieving actions \citep{yuan2024craftcustomizingllmscreating}.

% or integrate a Retrieval Augmented Generation (RAG) \citep{lewis2021retrievalaugmentedgenerationknowledgeintensivenlp} module into the action retrieval pipeline to enable agents make more informed decisions.

Our experiments primarily focus on evaluating the performance of LLMs with \textsc{GraphicBench}, as the amount of textual information involved in the planning process outweighs the image content. Therefore, a key complementary study still remains \---\ evaluating \textsc{GraphicBench} with visual language models. This would require modifying our current setup to directly prompt models with raw image files instead of using image captions as input, which we leave for future work.

% Additionally, \textsc{GraphicBench} is currently limited to four design types (book covers, business cards, postcards, and posters), and \textsc{GraphicTown} only considers three design expert agents, which does not cover the full range of design variants in graphic design generation tasks.

% In our preliminary studies, we compare \textsc{7b} LLMs, including \textsc{LLaMA-3.1 8b} and \textsc{Gemma-2 9b}, against Vision Language Models (VLMs), such as \textsc{LLaVA-1.5 7b} \citep{liu2023improvedllava} and \textsc{InternVL-2.5 8b} \citep{chen2025expandingperformanceboundariesopensource}. We find that VLMs often struggle to generate outputs in the expected format, failing as early as Step 1 in \textsc{GraphicTown}.
\section*{Acknowledgments}

We would like to thank our collaborators at Adobe Research for their valuable feedback, including Alexa Siu, Tong Sun, Saayan Mitra, and Stefano Petrangeli. Dayeon is especially grateful for the Adobe Research intern cohort for making the internship experience memorable, including Nishant Balepur, Dang Nguyen, Vishakh Padmakumar, Paiheng Xu, Hyunji Lee, and Yoonjoo Lee.

\bibliography{colm2025_conference}
\bibliographystyle{colm2025_conference}

\clearpage
\appendix
\section{Prompt Templates}
\label{appendix:prompts}
We show prompt templates used for constructing queries in \textsc{GraphicBench} (\S \ref{appendix:cb_prompts}), prompting LLMs for each step in \textsc{GraphicTown} (\S \ref{appendix:ct_prompts}), and evaluating for design pass rate using GPT-4 as a judge (\S \ref{appendix:eval_prompts}). 

\subsection{\textsc{GraphicBench} Prompts}
\label{appendix:cb_prompts}
\begin{prompt}[title={Prompt A.1.1: Query Construction (Book Cover)}]
\textbf{Task:} You are a design expert. Given the template for generating a book cover, fill in the placeholders with appropriate design components. Generate 5 diverse examples in a Python list of strings. Be as creative as possible. \\ \\
\textbf{Template:} Create a book cover design with a [background color] background, featuring [images]. The title [title] should be placed at [position] in [color] and the author name [author name] at [position] in [color]. \\
You may also include an optional [subtitle] or [tagline] if needed. \\ \\
**** Example Starts *** \\
Create a book cover design for a romance novel titled `Love \textbackslash n Story' featuring a silhouette illustration of a couple in a romantic pose against a pink moonlit background. The title should be at the top center, the author's name `A Novel By \textbackslash n Olivia Wilson' below the title, and the tagline `Best Selling Book of the Year' above the title, all in white. \\
**** Example Ends *** \\ \\
\textbf{Examples:}
\end{prompt}

\begin{prompt}[title={Prompt A.1.2: Query Construction (Business Card)}]
\textbf{Task:} You are a design expert. Given the template for generating a business card, fill in the placeholders with appropriate design components. Generate 5 diverse examples in a Python list of strings. Be as creative as possible. \\ \\
\textbf{Template:} Create a [side]-sided business card design with a [background color] background, featuring [images]. The name [brand name] should be placed at [position] in [color]. \\
You may also include an optional [contact details] or [tagline] if needed. \\ \\
**** Example Starts *** \\
Create a one-sided business card design with a light yellow background for the bookstore`CACTUS'. Replace the `T' in `CACTUS' with a cactus-shaped illustration in green font, centered and add a tagline `Livros Novos e Usados' in green font below the bookstore name. \\
**** Example Ends *** \\ \\
\textbf{Examples:}
\end{prompt}

\begin{prompt}[title={Prompt A.1.3: Query Construction (Postcard)}]
\textbf{Task:} You are a design expert. Given the template for generating a postcard, fill in the placeholders with appropriate design components. Generate 5 diverse examples in a Python list of strings. Be as creative as possible. \\ \\
\textbf{Template:} Create a postcard design with a [background color] background, featuring [images]. The message [message] should be placed at [position] in [color]. \\ \\
**** Example Starts *** \\
Create a postcard design with the message `Think Happy!' in a red, curly font on a floral background featuring a mix of warm-toned roses. Place a semi-transparent white box behind the message. \\
**** Example Ends *** \\ \\
\textbf{Examples:}
\end{prompt}

\begin{prompt}[title={Prompt A.1.4: Query Construction (Poster)}]
\textbf{Task:} You are a design expert. Given the template for generating a poster, fill in the placeholders with appropriate design components. Generate 5 diverse examples in a Python list of strings. Be as creative as possible. \\ \\
\textbf{Template:} Create a poster design with a [background color] background, featuring [images]. The title [title] should be placed at [position] in [color]. \\
You may also include an optional [tagline] if needed. \\ \\
**** Example Starts *** \\
Create a poster design with a light yellow background, featuring a large jellyfish illustration centered within a black rectangular box. Add a bold, black title `JELLYFISH' at the top and place a brief informative sentence about jellyfish in white font at the bottom left corner. \\
**** Example Ends *** \\ \\
\textbf{Examples:}
\end{prompt}
\begin{table*}
\centering
\resizebox{\linewidth}{!}{%
    \begin{tabular}{l p{0.6\textwidth} p{0.6\textwidth}}
    \toprule
    \textbf{Design Type} & \textbf{User Query} & \textbf{Questions} \\
    \toprule

    \textbf{Book Cover} & Please create a self-help book cover design titled `Achieve Your Dreams' with the author's name `Nathan White', featuring a person climbing a mountain centered in the lower half against a sunrise background. The title should be at the top center in white, the author's name below the title in white, and the tagline `Climb Higher, Dream Bigger.' below the author's name in white. 
    & [``Is there a text `Achieve Your Dreams'?'', ``Is there a text 'Nathan White'?'', ``Is there a person climbing a mountain?'', ``Is the background a sunrise?'', ``Is the text 'Achieve Your Dreams' at the top center in white?'', ``Is the text 'Nathan White' below the title in white?'', ``Is the text 'Climb Higher, Dream Bigger.' below the author's name in white?'', ``Is the person climbing a mountain centered in the lower half?''] \\
    \midrule

    \textbf{Business Card} & I need to create a business card design for `Sparkle Jewelry' with a royal blue background. Please include the company name in large gold font centered, and a medium-sized diamond icon placed above the company name. The contact details of `Phone: +1 987 654 3210 \textbackslash n Email: info@sparklejewelry.com \textbackslash n Address: 12 Gem St, Los Angeles, CA, USA' should be in small gold font, placed bottom right. & 
    [``Is the background of the business card royal blue?'', ``Is the company name `Sparkle Jewelry' in large gold font?'', ``Is the company name centered?'', ``Is there a medium-sized diamond icon?'', ``Is the diamond icon placed above the company name?'', ``Are the contact details in small gold font?'', ``Are the contact details placed at the bottom right?'', ``Is the phone number `+1 987 654 3210' included in the contact details?'', ``Is the email `info@sparklejewelry.com' included in the contact details?'', ``Is the address `12 Gem St, Los Angeles, CA, USA' included in the contact details?''] \\
    \midrule
    
    \textbf{Postcard} & Please create a motivational postcard design with the message `Stay Positive, Work Hard' at the top in red on a yellow background featuring a large lion roaring at the bottom. & 
    [``Is there a text 'Stay Positive, Work Hard'?'', ``Is the text 'Stay Positive, Work Hard' at the top?'', ``Is the text 'Stay Positive, Work Hard' in red?'', ``Is the background yellow?'', ``Is there an illustration of a lion?'', ``Is the lion roaring?'', ``Is the lion illustration large?'', ``Is the lion illustration at the bottom?''] \\
    \midrule
    
    \textbf{Poster} & Could you help create a promotional poster design for a jazz festival on a deep blue background, featuring a large image of a saxophonist playing in the center, a huge bold title `JAZZFEST' in gold at the top center, and event details `Jazz Festival | May 5-7, 2023 | Central Park, New York' in medium golden text at the bottom right? & 
    [``Is there a deep blue background?'', ``Is there a large image of a saxophonist playing in the center?'', ``Is there a huge bold title `JAZZFEST'?'', ``Is the title `JAZZFEST' in gold?'', ``Is the title 'JAZZFEST' at the top center?'', ``Is there event details `Jazz Festival | May 5-7, 2023 | Central Park, New York'?'', ``Is the event details in medium golden text?'', ``Is the event details at the bottom right?''] \\

    \bottomrule
    \end{tabular}
}
\caption{Examples of the generated questions using \textsc{GPT-4} per design type. We use the questions for computing the VQA pass rate as part of execution evaluation (\S \ref{sec:evaluation}).}
\label{tab:vqa_questions}
\end{table*}

\subsection{\textsc{GraphicTown} Prompts}
% We randomly select one human-annotated user query and workflow plan per design type from the training set as an in-context example.
\label{appendix:ct_prompts}
\begin{table*}
\centering
\resizebox{0.6\linewidth}{!}{%
    \begin{tabular}{ll}
    \toprule
    \textbf{Model} & \textbf{HuggingFace Name} \\
    \toprule

    \textsc{LLaMA-3.1 8b} & \texttt{meta-llama/Llama-3.1-8B-Instruct} \\
    \textsc{Gemma-2 9b} & \texttt{google/gemma-2-9b-it} \\
    \textsc{Gemma-2 27b} & \texttt{google/gemma-2-27b-it} \\
    \textsc{Qwen-2.5 7b} & \texttt{Qwen/Qwen2.5-7B-Instruct} \\
    \textsc{Qwen-2.5 14b} & \texttt{Qwen/Qwen2.5-14B-Instruct} \\

    \bottomrule
    \end{tabular}
}
\caption{HuggingFace model names for the tested open-weights models.}
\label{tab:huggingface_api}
\end{table*}

\begin{prompt}[title={Prompt A.2.1: (Step 1) Design Outline}]

\textbf{Task:} You are a proficient planner. Your task is to analyze the user's request and image file(s), identify the relevant design choices, and create a design outline in a JSON list format. \\ \\
\textbf{User request:} \texttt{\{user request\}} \\
\textbf{User image file(s):} \texttt{\{user image files\}} \\ \\
**** Output Format ***
\begin{lstlisting}
{
    "user_request": "Provide a brief summary of the user request.",
    "design_choices": {
        "background_color": "Identify the background color or style.",
        "text": {
            "content": "Specify the exact wording for each text to be included.",
            "position": "Identify the position of each text.",
            "color": "Identify the color of each text.",
            "size": "Identify the font size, either as an exact number or a description."
        },
        "image": {
            "content": "Specify the URL and caption for each image to be included in the design.",
            "position": "Identify the position of each image.",
            "size": "Identify the size of each image, either as exact dimensions or a description."
        }
    }
}
\end{lstlisting}

*** Key Requirements *** \\
- If there is any missing information from the user request, try to infer from the given information. \\
- Output should be in a list of JSON objects format. \\
- Do NOT include further explanation other than the JSON list. \\
- Be as concise and brief as possible. \\ \\
\textbf{Design Outline:}
\label{prompt:step1}
\end{prompt}
\begin{prompt}[title={Prompt A.2.2: (Step 2) Expert Recruitment}]

\textbf{Task:} Your task is to recruit the necessary experts to complete a design outlined in the user request. Create a recruitment status in JSON list format. \\ \\
\textbf{User request:} \texttt{\{user request\}} \\
\textbf{User image file(s):} \texttt{\{user image files\}} \\ \\
You can recruit from the three experts with the following profiles:
\begin{itemize}[leftmargin=10pt, itemsep=2pt, parsep=-1pt]
    \item \textbf{Photo Editor}
    \begin{itemize}[leftmargin=1em] % Indented sublist
        \item Job Responsibilities:
        \begin{itemize}[leftmargin=1em] % Further indentation
            \item Image editing: Cropping, adjusting composition, correcting lighting, and retouching images or illustrations.
            \item Color correction: Adjusting brightness and contrast or adjusting hue and saturation.
            \item Apply filters: Apply different filters (e.g., photo, glass, ocean ripple, watercolor) to images.
        \end{itemize}
    \end{itemize}

    \item \textbf{Vector Graphic Editor}
    \begin{itemize}[leftmargin=1em] % Indented sublist
        \item Job Responsibilities:
        \begin{itemize}[leftmargin=1em]
            \item Draw shapes: Drawing simple shapes (circle, polygon, square, star) on canvas.
        \end{itemize}
    \end{itemize}

    \item \textbf{Layout Designer}
    \begin{itemize}[leftmargin=1em] % Indented sublist
        \item Job Responsibilities:
        \begin{itemize}[leftmargin=1em]
            \item Customize layout templates: Create grid systems for books, brochures, cards, and magazines to organize the layout.
            \item Export files: Export documents to any format, in print or digital.
            \item Combine text and visual elements: Combine visual elements from other apps with text into a completed design.
        \end{itemize}
    \end{itemize}
\end{itemize}
*** Output Format *** \\
Each object in the JSON list should follow:
\begin{lstlisting}
{
    "expert": "Name of the expert (Photo Editor, Vector Graphic Editor, Layout Desinger).",
    "task": "High-level task that can be performed by the expert."
}
\end{lstlisting}
*** Example Starts ***
\begin{lstlisting}
[
    {"expert": "Photo Editor", "task": "Add the provided images to create a deep purple night sky background with a large dreamy moon centered, surrounded by small twinkling stars spread across the top half of the cover."}, 
    {"expert": "Layout Designer", "task": "Combine the edited image with the title 'Moonlit Fantasies', the author name 'J.K. Stellar', and the tagline 'A Journey Through the Night Sky.' to create the book cover design."}
]
\end{lstlisting}
*** Example Ends *** \\ \\
**** Key Requirements *** \\
- Only recruit each expert one. \\
- The name of the expert must match those in the expert profiles. \\
- For task description, explain how the expert can contribute towards the final product. Summarize in one sentence. \\
- In order to achieve the task in the design outline, experts should work together and their task will be dependent to each other. Arrange in the order of which expert should finish first. \\
- Output should be in a list of JSON objects format. \\
- Do NOT include further explanation other than in the JSON list. \\
- Be as concise and brief as possible. \\ \\
\textbf{Recruitment status:}
\label{prompt:step2}
\end{prompt}
\begin{prompt}[title={Prompt A.2.3: (Step 3) Workflow Generation}]

\textbf{Task:} You are a proficient \texttt{\{expert\}}. You are recruited to collaborate on a design project with other experts. \\ \\
The design choices collected from the user have been compiled into the following design outline. Use as reference: \\
\texttt{\{design outline\}} \\ \\
You are assigned to complete the following task as in the design outline: \texttt{\{task\}}. Please plan a sequence of detailed, low-level subtasks required to accomplish this task and output them as a JSON list. \\ \\
**** Output Format *** \\
Each object in the JSON list should follow:
\begin{lstlisting}
{
    "id": "ID of the subtask, starting from 1.",
    "expert": "Name of the expert.",
    "description": "Description of the subtask in one sentence."
}
\end{lstlisting}
*** Example Starts ***
\begin{lstlisting}
[
    {"id": 1, "expert": "Photo Editor", "description": "Create a new document with book cover dimensions."},
    {"id": 2, "expert": "Photo Editor", "description": "Set the background color to light pink."},
    {"id": 3, "expert": "Photo Editor", "description": "Import the pink moonlit image from 'static/pink_moonlit.png'."},
    {"id": 4, "expert": "Photo Editor", "description": "Resize the pink moonlit image to medium size, covering the bottom part of the document."},
    {"id": 5, "expert": "Photo Editor", "description": "Reposition the pink moonlit image to the bottom-center of the document."},
    {"id": 6, "expert": "Photo Editor", "description": "Import the couple silhouette illustration from 'static/couple_silhouette.png'."},
    {"id": 7, "expert": "Photo Editor", "description": "Resize the couple silhouette illustration to span across the lower half of the cover."},
    {"id": 8, "expert": "Photo Editor", "description": "Reposition the couple silhouette illustration to be centered in the bottom-middle part."},
    {"id": 9, "expert": "Photo Editor", "description": "Adjust the background colors to match the light pink moonlit theme."},
    {"id": 10, "expert": "Photo Editor", "description": "Save the document in a psd format suitable for further editing by the Layout Designer."},
]
\end{lstlisting}
*** Example Ends *** \\ \\
**** Key Requirements *** \\
- First step should always be creating a new document and the last step should always be saving the document in appropriate file format. \\
- Use the exact image URLs the user provided when importing images. \\
- Do not include very basic operations such as opening the software or closing the software. \\
- Do not include new expert in the plan. \\
- Output should be in a list of JSON objects format. \\
- Do NOT include further explanation other than in the JSON list. \\
- Be as concise and brief as possible. \\ \\
\textbf{Sequence of subtasks:}
\label{prompt:step3}
\end{prompt}
\begin{prompt}[title={Prompt A.2.4: (Step 4) Workflow Supervision}]

\textbf{Task:} You are the supervisor of a design project that requires collaboration among various design experts. \\ \\
The following experts have been recruited for the project. Use as reference: \\
\texttt{\{recruitment status\}} \\ \\
Each expert has submitted their proposed workflow plans: \\
\texttt{\{workflow plans\}} \\ \\
Your task is to combine these proposed workflow plans into a cohesive sequence of tasks in a JSON list format. \\ \\
**** Output Format *** \\
Each object in the JSON list should follow:
\begin{lstlisting}
{
    "id": "ID of the subtask, starting from 1.",
    "expert": "Name of the expert.",
    "description": "Description of the subtask in one sentence."
}
\end{lstlisting}
*** Example Starts ***
\begin{lstlisting}
[
    {"id": 1, "expert": "Photo Editor", "description": "Create a new document with book cover dimensions."},
    {"id": 2, "expert": "Photo Editor", "description": "Set the background color to light pink."},
    ...
    {"id": 11, "expert": "Layout Designer", "description": "Create a new document with book cover dimensions."},
    {"id": 12, "expert": "Layout Designer", "description": "Import the edited image from the Photo Editor: 'moonlit_illustration_edited.psd'."},
    {"id": 13, "expert": "Layout Designer", "description": "Resize the edited image to cover the entire document."},
    {"id": 14, "expert": "Layout Designer", "description": "Create text for the title 'LOVE\nSTORY'."},
    {"id": 15, "expert": "Layout Designer", "description": "Apply the Andale Mono font to the title text."},
    ...
    {"id": 31, "expert": "Layout Designer", "description": "Reposition the tagline text above the title."},
    {"id": 32, "expert": "Layout Designer", "description": "Export the final book cover design as a PDF file."}
]
\end{lstlisting}
*** Example Ends *** \\ \\
**** Key Requirements *** \\
- Do NOT repeat any steps that are already completed in previous step. \\
- For each expert, first step should always be creating a new document and the last step should always be saving the document in appropriate file format. \\
- When switching experts, use the output from the previous expert as input for the next. \\
- Once an expert is used and switched to another expert, it should not be used again. \\
- You should output only one list of workflow plan. \\
- Start the id from 1 to the number of steps in the workflow. \\
- Arrange each subtask in a chronological order. \\
- Output should be in a list of JSON objects format. \\
- Do NOT include further explanation other than in the JSON list. \\
- Be as concise and brief as possible. \\ \\
\textbf{Supervised sequence of subtasks:}
\label{prompt:step4}
\end{prompt}
\begin{prompt}[title={Prompt A.2.5: (Step 5) Action Retrieval}]

\textbf{Task:} You are a proficient \texttt{\{expert\}}. You are recruited to collaborate on a design project with other experts. Use your available list of actions to map each step in the sequence of subtasks to an action. \\ \\
\textbf{Sequence of subtasks:} \texttt{\{workflow plan\}} \\ \\
Your available actions are as below: \\
\texttt{\{list of actions\}} \\ \\
**** Output Format *** \\
Each object in the JSON list should follow:
\begin{lstlisting}
{
    "id": "ID of the subtask, starting from 1.",
    "expert": "Name of the expert.",
    "description": "Description of the subtask in one sentence.",
    "action": "Name of the mapped action.",
    "parameters": "Dictionary of parameter keys and corresponding values."
}
\end{lstlisting}
*** Example Starts ***
\begin{lstlisting}
[
    {"id": 1, "expert": "Photo Editor", "description": "Create a new document with book cover dimensions.", "skill": "CreateDocument", "parameters": {"docType": "book cover"}},
    {"id": 2, "expert": "Photo Editor", "description": "Set the background color to light pink.", "skill": "SetBackgroundColor", "parameters": {"red": 255, "green": 179, "blue": 238}},
    ...
    {"id": 8, "expert": "Photo Editor", "description": "Reposition the couple silhouette illustration to be centered in the bottom-middle part.", "parameters": {"layerName": "SilhouetteLayer", "posX": 267, "posY": 1052}},
    {"id": 9, "expert": "Photo Editor", "description": "Adjust the background colors to match the light pink moonlit theme.", "skill": "AdjustHSL", "parameters": {"layerName": "MoonlitLayer", "hue": 18, "saturation": -18, "light": 0}},
    {"id": 10, "expert": "Photo Editor", "description": "Save the document in a format suitable for further editing by the Layout Designer.", "skill": "SaveDocument", "parameters": {"fileName": "moonlit_illustration_edited", "format": "psd"}},
]
\end{lstlisting}
*** Example Ends *** \\ \\
**** Key Requirements *** \\
- For any file name that appears in the design outline, use exact file names in your sequence of subtasks. \\
- Each step should only be mapped to one action. If a step of the workflow is not able to be mapped to one action, it means the step can be decomposed further into multiple steps. You can reformat, reorder, add, edit steps of the workflow if needed to be directly mapped to actions. \\
- Each step should have an action and a dictionary of parameter values. \\
- For layerName, try to name it as to end as Layer (e.g., BackgroundLayer, TitleLayer). \\
- For detailed numeric values (e.g., height, width, x-axis position, y-axis position), consider the document's dimensions, imagine, and propose a likely value. \\
- Arrange each subtask in a chronological order. \\
- Output should be in a list of JSON objects format. \\
- Do NOT include further explanation other than in the JSON list. \\
- Be as concise and brief as possible. \\ \\
\textbf{Sequence of subtasks:}
\label{prompt:step5}
\end{prompt}

\subsection{Evaluation Prompts}
\label{appendix:eval_prompts}
\begin{prompt}[title={Prompt A.3.1: Design Pass Rate Evaluation (Color)}]
\textbf{Task:} Evaluate if the workflow plan (1) correctly applies the background color and (2) the background and the text color are contrasting. Return a score between 1 to 5 according to the scoring rubric. \\ \\
\textbf{Background color:} \texttt{\{background color\}} \\
\textbf{Text elements:} \texttt{\{text\}} \\
\textbf{Workflow plan:} \texttt{\{workflow plan\}} \\ \\
**** Scoring Rubric *** \\
- 1: Workflow plan fails to reflect all of the color constraints specified. \\
- 3: Workflow plan reflects approximately half of the color constraints specified. \\
- 5: Workflow plan reflects all of the color constraints specified. \\ \\
Score should strictly be a number between 1 to 5. Do not include any further explanation other than the score. \\
\textbf{Score:}
\end{prompt}

\begin{prompt}[title={Prompt A.3.2: Design Pass Rate Evaluation (Text)}]
\textbf{Task:} Evaluate if the workflow plan adequately applies the text elements (e.g., title, tagline) specified. Return a score between 1 to 5 according to the scoring rubric. \\ \\
\textbf{Text elements:} \texttt{\{text\}} \\
\textbf{Workflow plan:} \texttt{\{workflow plan\}} \\ \\
**** Scoring Rubric *** \\
- 1: Workflow plan fails to reflect all of the text elements specified. \\
- 3: Workflow plan reflects approximately half of the text elements specified. \\
- 5: Workflow plan reflects all of the text elements specified. \\ \\
Score should strictly be a number between 1 to 5. Do not include any further explanation other than the score. \\
\textbf{Score:}
\end{prompt}

\begin{prompt}[title={Prompt A.3.3: Design Pass Rate Evaluation (Image)}]
\textbf{Task:} Evaluate if the workflow plan adequately applies the image elements (e.g., size, position) specified. Return a score between 1 to 5 according to the scoring rubric. \\ \\
\textbf{Image elements:} \texttt{\{image\}} \\
\textbf{Workflow plan:} \texttt{\{workflow plan\}} \\ \\
**** Scoring Rubric *** \\
- 1: Workflow plan fails to reflect all of the image elements specified. \\
- 3: Workflow plan reflects approximately half of the image elements specified. \\
- 5: Workflow plan reflects all of the image elements specified. \\ \\
Score should strictly be a number between 1 to 5. Do not include any further explanation other than the score. \\
\textbf{Score:}
\end{prompt}
\begin{prompt}[title={Prompt A.3.4: VQA Evaluation}]
\textbf{Instruction:} Look at the image and answer the question with `Yes' or `No'. \\ \\
\textbf{Question:} \texttt{\{quetion\}} \\
\textbf{Answer:}
\end{prompt}
\begin{prompt}[title={Prompt A.3.5: Creativity Evaluation (Originality)}]
\textbf{Instruction:} Evaluate the originality of the image generated based on the user query. Originality measures the uniqueness of the ideas generated. Original ideas are those that are rare or unconventional, differing from the norm. Return a score between 1 to 5 according to the scoring rubric. \\ \\
\textbf{User query:} \texttt{\{user query\}} \\ \\
**** Scoring Rubric *** \\
- 1: Image is highly conventional and predictable. No significant signs of creative thinking is shown. \\
- 2: Image shows minimal originality and mostly align with typical or common responses. Few novel elements are present. \\
- 3: Image is somewhat original, with a mix of conventional and unique elements. \\
- 4: Image is noticeable original and uncommon. It shows creative thinking and depart meaningfully from conventional norms. \\
- 5: Image is highly unique, rare, and stand out as unconventional. They demonstrate a strong departure from typical or expected approaches. \\ \\
Score should strictly be a number between 1 to 5. Do not include any further explanation other than the score. \\
\textbf{Score:}
\end{prompt}

\begin{prompt}[title={Prompt A.3.6: Creativity Evaluation (Elaboration)}]
\textbf{Instruction:} Evaluate the elaboration of the image generated based on the user query. Elaboration refers to the ability to expand upon, refine, and embellish an idea. It involves adding details, developing nuances, and building upon a basic concept to make it more intricate or complex. Return a score between 1 to 5 according to the scoring rubric. \\ \\
\textbf{User query:} \texttt{\{user query\}} \\ \\
**** Scoring Rubric *** \\
- 1: Image is presented in a simpler or vague manner with no meaningful development or supporting detail. \\
- 2: Image is minimally expanded, with few details or refinements added. \\
- 3: Image includes expansion of some details, but elaboration is somewhat surface-level. \\
- 4: Image well-expands the user query with several added details and refinements. \\
- 5: Image thoroughly expands the user query with rich, specific details or refinements added beyond the core concept. \\ \\
Score should strictly be a number between 1 to 5. Do not include any further explanation other than the score. \\
\textbf{Score:}
\end{prompt}

\section{Details on \textsc{GraphicBench}}
\label{appendix:DesignBench}

\subsection{Design Concept Distribution}
\label{appendix:concept_distribution}
\begin{figure*}[t]
    \centering
    \begin{subfigure}{0.5\textwidth}
        \centering
        \includegraphics[height=4.5cm]{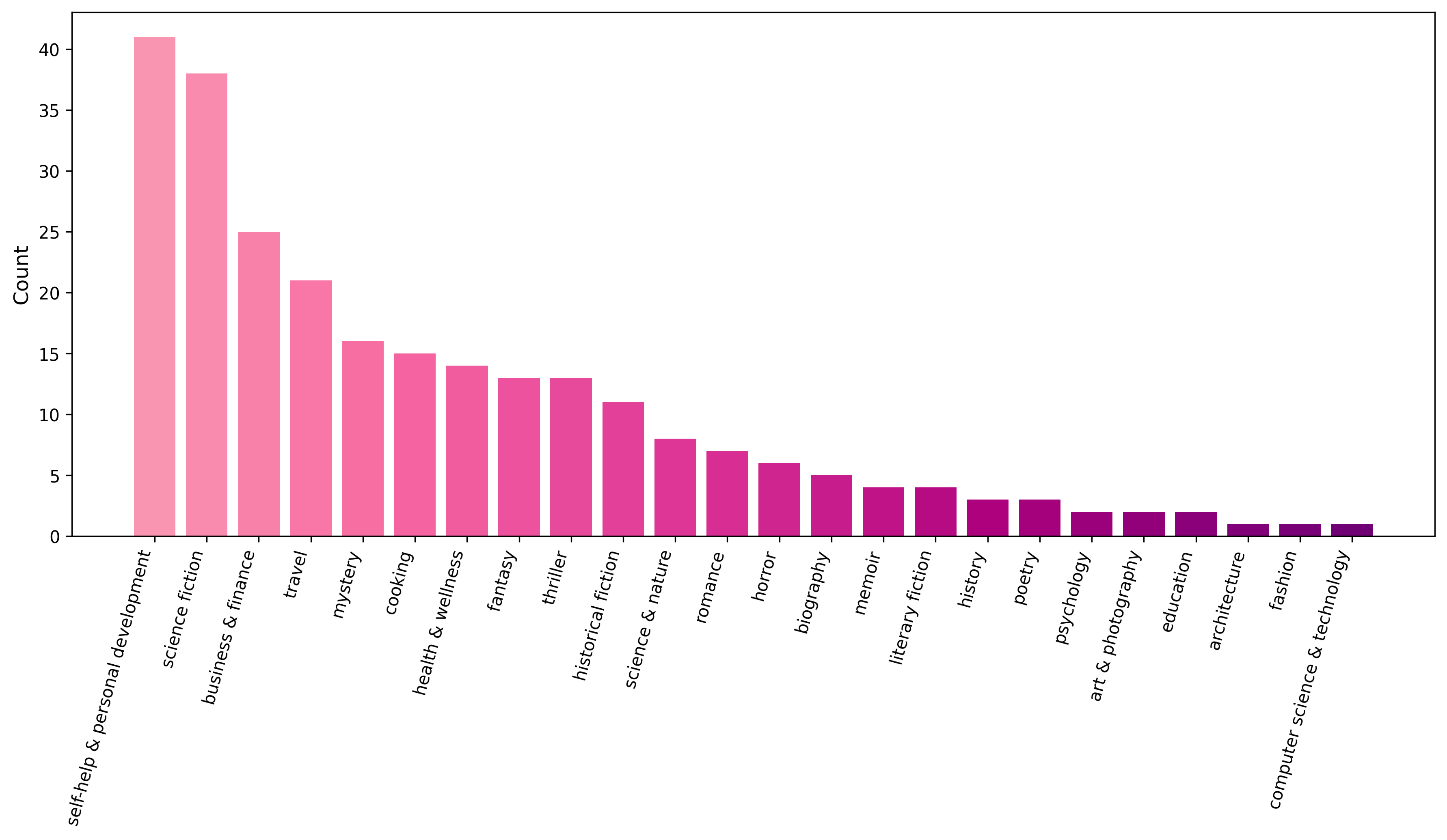}
        \caption{Book Cover}
    \end{subfigure}

    \begin{subfigure}{0.5\textwidth}
        \centering
        \includegraphics[height=4.5cm]{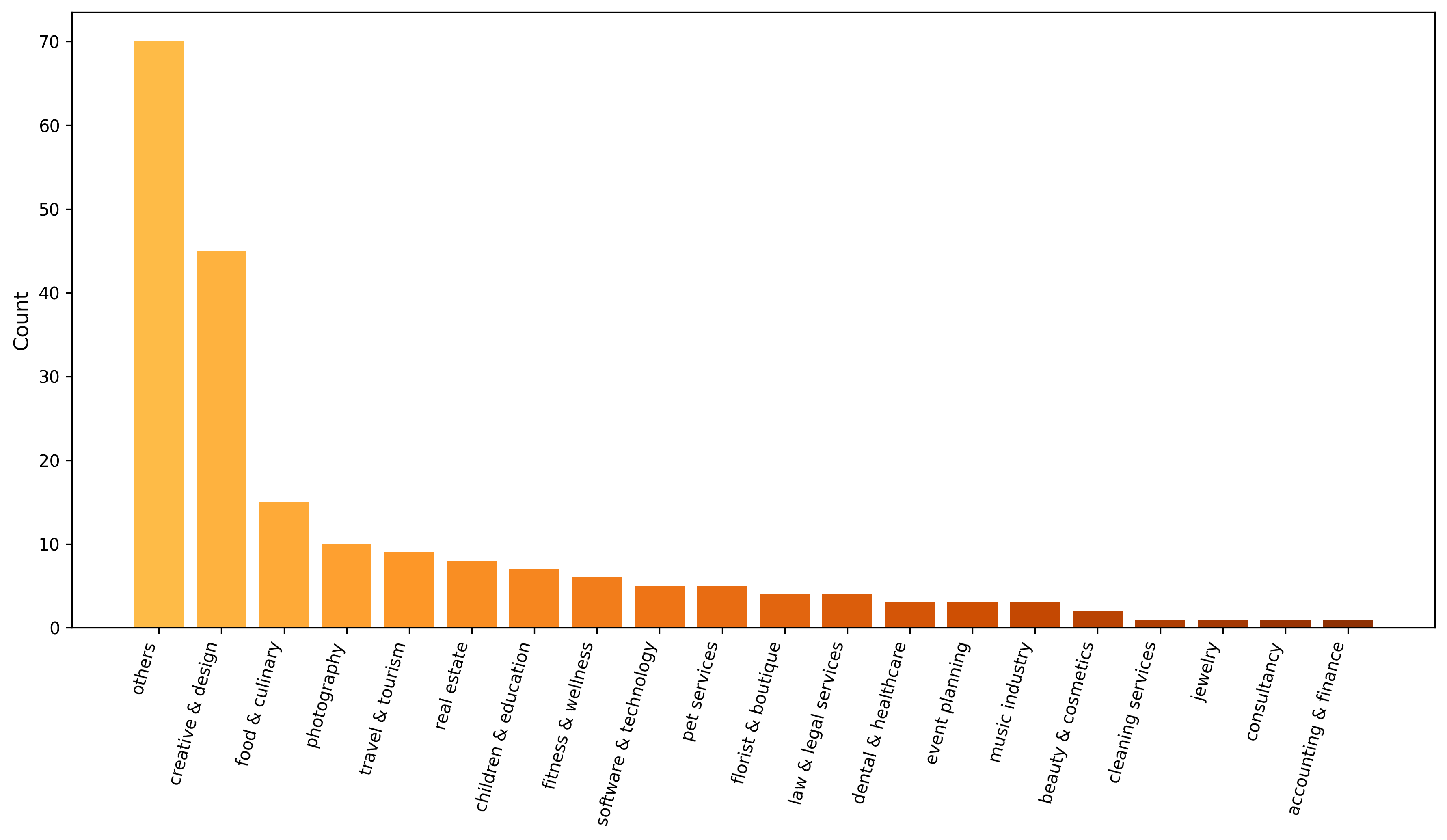}
        \caption{Business Card}
    \end{subfigure}

    \begin{subfigure}{0.5\textwidth}
        \centering
        \includegraphics[height=4.5cm]{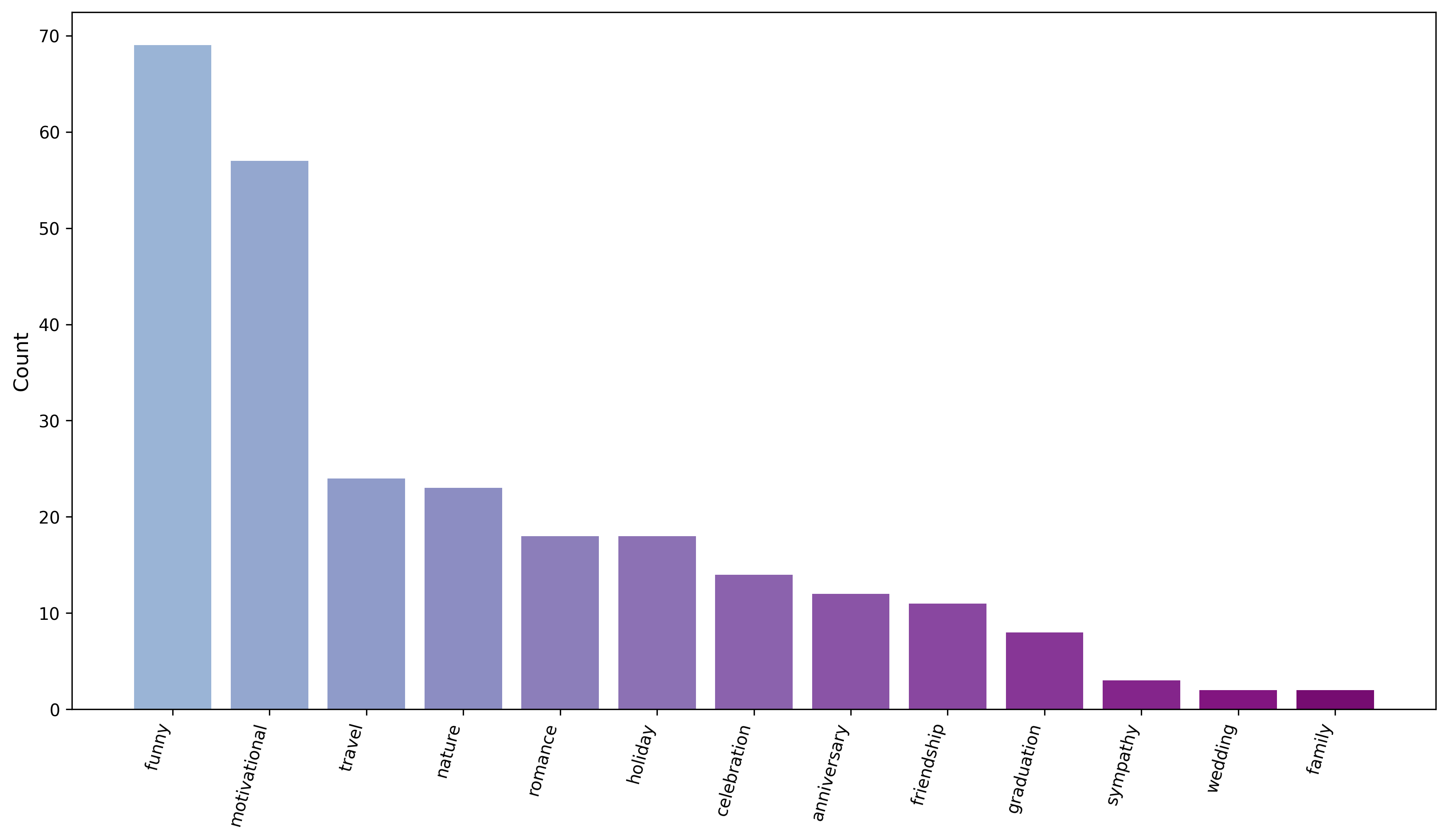}
        \caption{Postcard}
    \end{subfigure}

    \begin{subfigure}{0.5\textwidth}
        \centering
        \includegraphics[height=4.5cm]{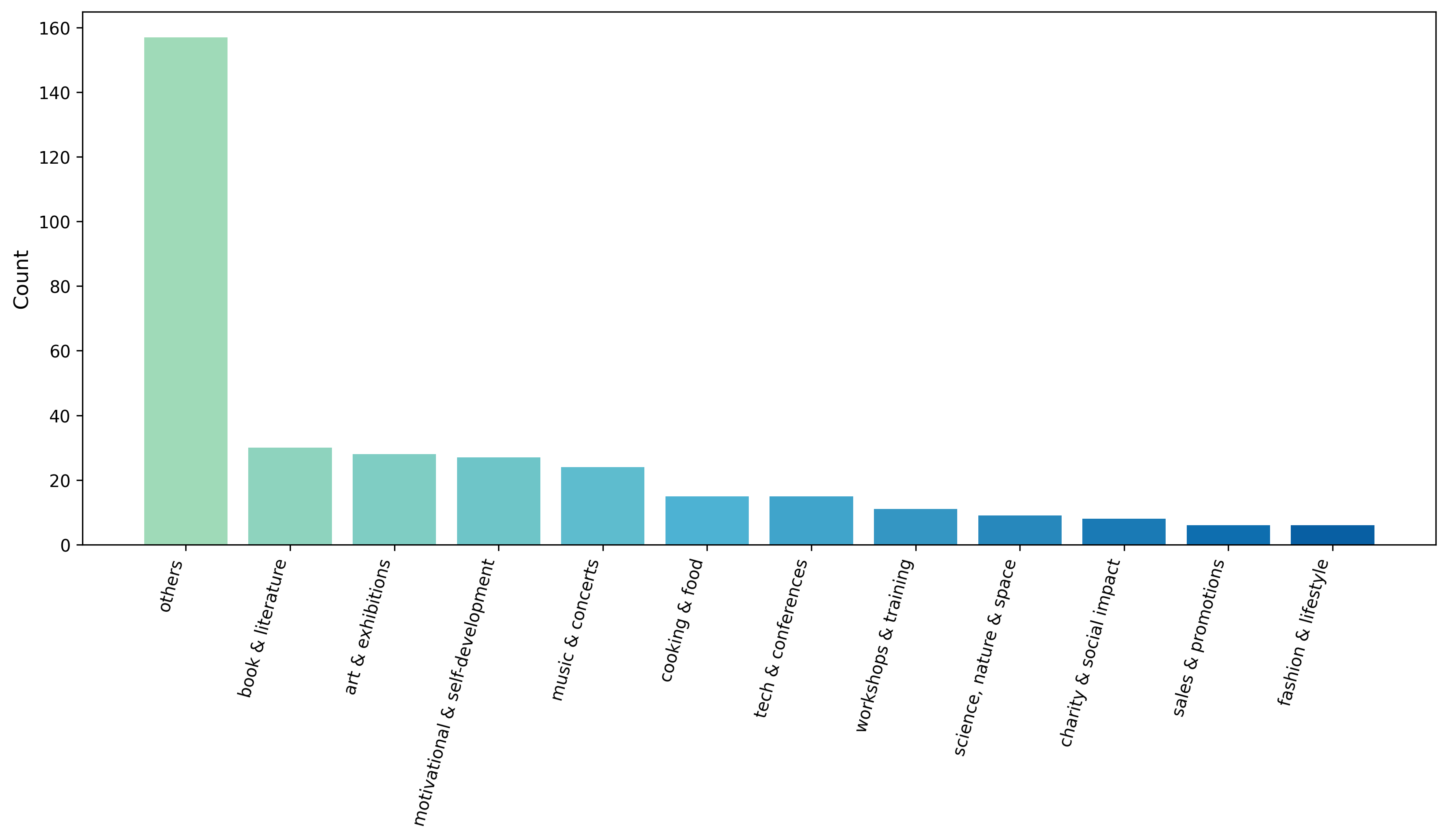}
        \caption{Poster}
    \end{subfigure}

    \caption{Distribution of design concepts by design type in \textsc{GraphicBench}. We automatically extract the main theme (e.g., postcard) or event (e.g., poster) from each user query and categorize it based on predefined categories.}
    \label{fig:concept_distribution}
\end{figure*}

We provide a detailed distribution of design concepts by design type in Figure \ref{fig:concept_distribution}. We prompt \textsc{GPT-4} to identify the main theme or event outlined in the user query and categorize into predefined categories. We show that the distribution spans a diverse range of categories, which highlights the benchmark's breadth and variety.

\subsection{Reference Plan Annotation}
\label{appendix:reference_plan_annotation}
\begin{figure*}
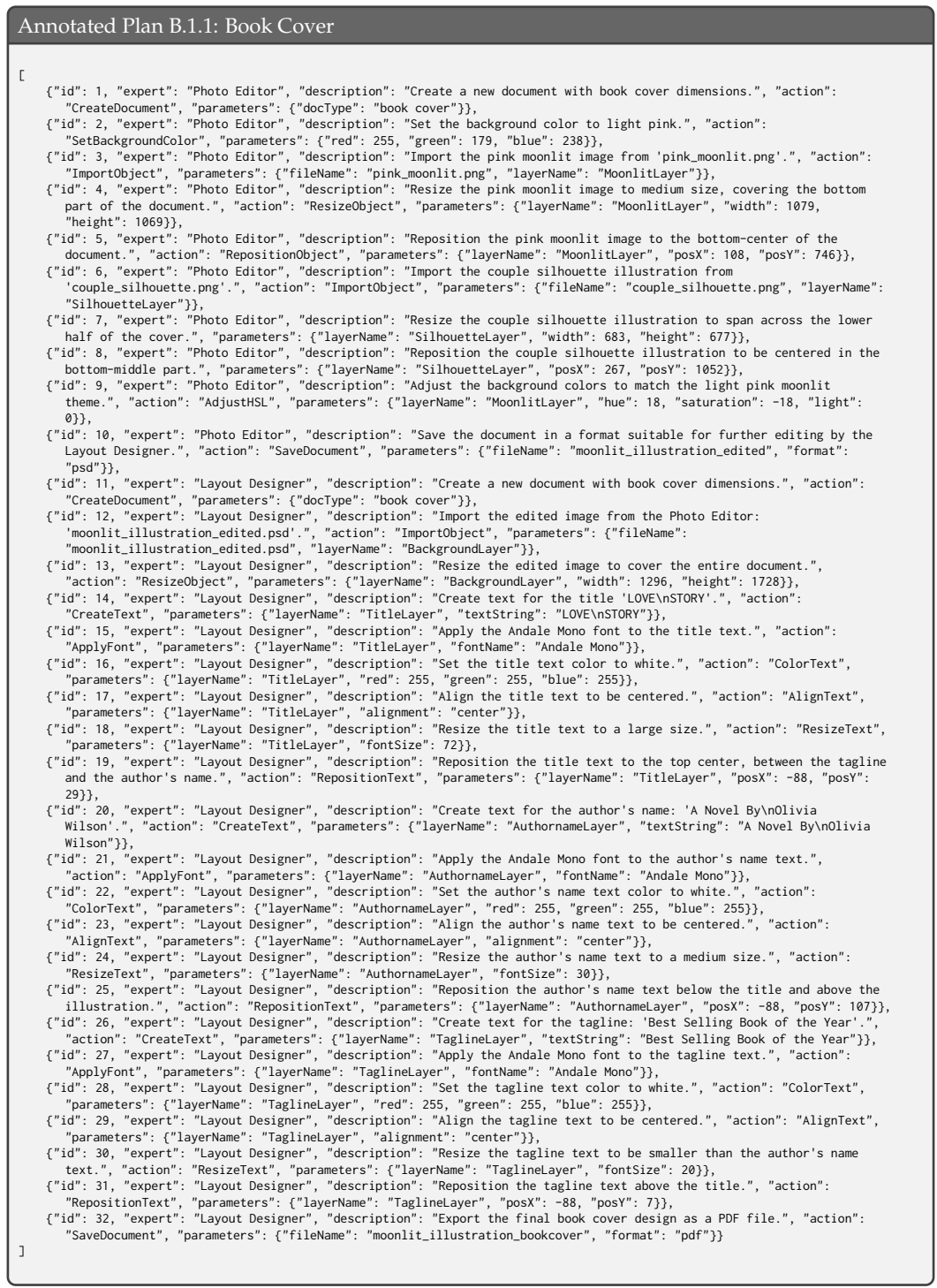


\begin{prompt}[title={Annotated Plan B.1.1: Book Cover}]
\begin{lstlisting}[basicstyle=\ttfamily\tiny]
[
    {"id": 1, "expert": "Photo Editor", "description": "Create a new document with book cover dimensions.", "action": "CreateDocument", "parameters": {"docType": "book cover"}},
    {"id": 2, "expert": "Photo Editor", "description": "Set the background color to light pink.", "action": "SetBackgroundColor", "parameters": {"red": 255, "green": 179, "blue": 238}},
    {"id": 3, "expert": "Photo Editor", "description": "Import the pink moonlit image from 'pink_moonlit.png'.", "action": "ImportObject", "parameters": {"fileName": "pink_moonlit.png", "layerName": "MoonlitLayer"}},
    {"id": 4, "expert": "Photo Editor", "description": "Resize the pink moonlit image to medium size, covering the bottom part of the document.", "action": "ResizeObject", "parameters": {"layerName": "MoonlitLayer", "width": 1079, "height": 1069}},
    {"id": 5, "expert": "Photo Editor", "description": "Reposition the pink moonlit image to the bottom-center of the document.", "action": "RepositionObject", "parameters": {"layerName": "MoonlitLayer", "posX": 108, "posY": 746}},
    {"id": 6, "expert": "Photo Editor", "description": "Import the couple silhouette illustration from 'couple_silhouette.png'.", "action": "ImportObject", "parameters": {"fileName": "couple_silhouette.png", "layerName": "SilhouetteLayer"}},
    {"id": 7, "expert": "Photo Editor", "description": "Resize the couple silhouette illustration to span across the lower half of the cover.", "parameters": {"layerName": "SilhouetteLayer", "width": 683, "height": 677}},
    {"id": 8, "expert": "Photo Editor", "description": "Reposition the couple silhouette illustration to be centered in the bottom-middle part.", "parameters": {"layerName": "SilhouetteLayer", "posX": 267, "posY": 1052}},
    {"id": 9, "expert": "Photo Editor", "description": "Adjust the background colors to match the light pink moonlit theme.", "action": "AdjustHSL", "parameters": {"layerName": "MoonlitLayer", "hue": 18, "saturation": -18, "light": 0}},
    {"id": 10, "expert": "Photo Editor", "description": "Save the document in a format suitable for further editing by the Layout Designer.", "action": "SaveDocument", "parameters": {"fileName": "moonlit_illustration_edited", "format": "psd"}},
    {"id": 11, "expert": "Layout Designer", "description": "Create a new document with book cover dimensions.", "action": "CreateDocument", "parameters": {"docType": "book cover"}},
    {"id": 12, "expert": "Layout Designer", "description": "Import the edited image from the Photo Editor: 'moonlit_illustration_edited.psd'.", "action": "ImportObject", "parameters": {"fileName": "moonlit_illustration_edited.psd", "layerName": "BackgroundLayer"}},
    {"id": 13, "expert": "Layout Designer", "description": "Resize the edited image to cover the entire document.", "action": "ResizeObject", "parameters": {"layerName": "BackgroundLayer", "width": 1296, "height": 1728}},
    {"id": 14, "expert": "Layout Designer", "description": "Create text for the title 'LOVE\nSTORY'.", "action": "CreateText", "parameters": {"layerName": "TitleLayer", "textString": "LOVE\nSTORY"}},
    {"id": 15, "expert": "Layout Designer", "description": "Apply the Andale Mono font to the title text.", "action": "ApplyFont", "parameters": {"layerName": "TitleLayer", "fontName": "Andale Mono"}},
    {"id": 16, "expert": "Layout Designer", "description": "Set the title text color to white.", "action": "ColorText", "parameters": {"layerName": "TitleLayer", "red": 255, "green": 255, "blue": 255}},
    {"id": 17, "expert": "Layout Designer", "description": "Align the title text to be centered.", "action": "AlignText", "parameters": {"layerName": "TitleLayer", "alignment": "center"}},
    {"id": 18, "expert": "Layout Designer", "description": "Resize the title text to a large size.", "action": "ResizeText", "parameters": {"layerName": "TitleLayer", "fontSize": 72}},
    {"id": 19, "expert": "Layout Designer", "description": "Reposition the title text to the top center, between the tagline and the author's name.", "action": "RepositionText", "parameters": {"layerName": "TitleLayer", "posX": -88, "posY": 29}},
    {"id": 20, "expert": "Layout Designer", "description": "Create text for the author's name: 'A Novel By\nOlivia Wilson'.", "action": "CreateText", "parameters": {"layerName": "AuthornameLayer", "textString": "A Novel By\nOlivia Wilson"}},
    {"id": 21, "expert": "Layout Designer", "description": "Apply the Andale Mono font to the author's name text.", "action": "ApplyFont", "parameters": {"layerName": "AuthornameLayer", "fontName": "Andale Mono"}},
    {"id": 22, "expert": "Layout Designer", "description": "Set the author's name text color to white.", "action": "ColorText", "parameters": {"layerName": "AuthornameLayer", "red": 255, "green": 255, "blue": 255}},
    {"id": 23, "expert": "Layout Designer", "description": "Align the author's name text to be centered.", "action": "AlignText", "parameters": {"layerName": "AuthornameLayer", "alignment": "center"}},
    {"id": 24, "expert": "Layout Designer", "description": "Resize the author's name text to a medium size.", "action": "ResizeText", "parameters": {"layerName": "AuthornameLayer", "fontSize": 30}},
    {"id": 25, "expert": "Layout Designer", "description": "Reposition the author's name text below the title and above the illustration.", "action": "RepositionText", "parameters": {"layerName": "AuthornameLayer", "posX": -88, "posY": 107}},
    {"id": 26, "expert": "Layout Designer", "description": "Create text for the tagline: 'Best Selling Book of the Year'.", "action": "CreateText", "parameters": {"layerName": "TaglineLayer", "textString": "Best Selling Book of the Year"}},
    {"id": 27, "expert": "Layout Designer", "description": "Apply the Andale Mono font to the tagline text.", "action": "ApplyFont", "parameters": {"layerName": "TaglineLayer", "fontName": "Andale Mono"}},
    {"id": 28, "expert": "Layout Designer", "description": "Set the tagline text color to white.", "action": "ColorText", "parameters": {"layerName": "TaglineLayer", "red": 255, "green": 255, "blue": 255}},
    {"id": 29, "expert": "Layout Designer", "description": "Align the tagline text to be centered.", "action": "AlignText", "parameters": {"layerName": "TaglineLayer", "alignment": "center"}},
    {"id": 30, "expert": "Layout Designer", "description": "Resize the tagline text to be smaller than the author's name text.", "action": "ResizeText", "parameters": {"layerName": "TaglineLayer", "fontSize": 20}},
    {"id": 31, "expert": "Layout Designer", "description": "Reposition the tagline text above the title.", "action": "RepositionText", "parameters": {"layerName": "TaglineLayer", "posX": -88, "posY": 7}},
    {"id": 32, "expert": "Layout Designer", "description": "Export the final book cover design as a PDF file.", "action": "SaveDocument", "parameters": {"fileName": "moonlit_illustration_bookcover", "format": "pdf"}}
]
\end{lstlisting}
\end{prompt}
\caption{Human-annotated workflow plan for the book cover example in Table \ref{tab:reference_plan}.}
\label{fig:plan_1}
\end{figure*}

We detail examples of human-annotated user queries and design outputs in Table \ref{tab:reference_plan} and their corresponding workflow plans in Figures \ref{fig:plan_1} to \ref{fig:plan_4}. We use the same Adobe CC design tool combinations as specified by the designers of the references from the Behance platform. We show the identified key design components for each design type in Table \ref{tab:design_components}. The average number of text and image elements included in each query are 3.05, 1.33 for book covers, 2.15 and 1.00 for business cards, 1.03 and 1.28 for postcards, and 1.99 and 1.04 for posters.

\begin{table*}
\centering
\resizebox{0.7\linewidth}{!}{%
    \begin{tabular}{lll}
    \toprule
    \textbf{Design Type} & \textbf{Design Components} & \textbf{Required?} \\
    \toprule
    \multirow{6}{*}{\textbf{Book Cover}} & Background color & \ding{51} \\
    & Title \small{(content, size, color, position)} & \ding{51} \\
    & Author Name \small{(content, size, color, position)} & \ding{51} \\
    & Subtitle \small{(content, size, color, position)} & \ding{55} \\
    & Tagline \small{(content, size, color, position)} & \ding{55} \\
    & Image \small{(size, position, image URL, caption)} & \ding{51} \\
    \midrule

    \multirow{5}{*}{\textbf{Business Card}} & Background color & \ding{51} \\
    & Brand Name \small{(content, size, color, position)} & \ding{51} \\
    & Tagline \small{(content, size, color, position)} & \ding{55} \\
    & Contact Details \small{(content, size, color, position)} & \ding{55} \\
    & Image \small{(size, position, image URL, caption)} & \ding{51} \\
    \midrule

    \multirow{3}{*}{\textbf{Postcard}} & Background color & \ding{51} \\
    & Message \small{(content, size, color, position)} & \ding{51} \\
    & Image \small{(size, position, image URL, caption)} & \ding{51} \\
    \midrule

    \multirow{4}{*}{\textbf{Poster}} & Background color & \ding{51} \\
    & Title \small{(content, size, color, position)} & \ding{51} \\
    & Tagline \small{(content, size, color, position)} & \ding{55} \\
    & Image \small{(size, position, image URL, caption)} & \ding{51} \\

    \bottomrule
    \end{tabular}
}
\caption{Design components for each design type (book cover, business card, postcard, poster), which are associated with sub-components listed in parentheses. \textbf{Required?}: Indicates whether the component is explicitly required in the user query during query construction. Note that even when a component is required, its sub-component details may be missing from the user query.}
\label{tab:design_components}
\end{table*}
\begin{table*}
\centering
\resizebox{\linewidth}{!}{%
    \begin{tabular}{l p{0.6\textwidth} p{0.15\textwidth} p{0.15\textwidth} c} % Adjust column widths
    \toprule
    \textbf{Design Type} & \textbf{User Query} & \textbf{Reference Design} & \textbf{Generated Design} & \textbf{App} \\
    \toprule
    \textbf{Book Cover} & 
    \begin{minipage}{0.6\textwidth}
        Create a book cover design for a romance novel titled `Love \textbackslash n Story' featuring a silhouette illustration of a couple in a romantic pose against a pink moonlit background. The title should be at the top center, the author's name `A Novel By \textbackslash n  Olivia Wilson' below the title, and the tagline `Best Selling Book of the Year' above the title, all in white.
    \end{minipage} & 
    \begin{minipage}{0.15\textwidth}
        \centering
        \makebox[\textwidth]{ % Ensures both images fit inside the column
            \includegraphics[width=\textwidth]{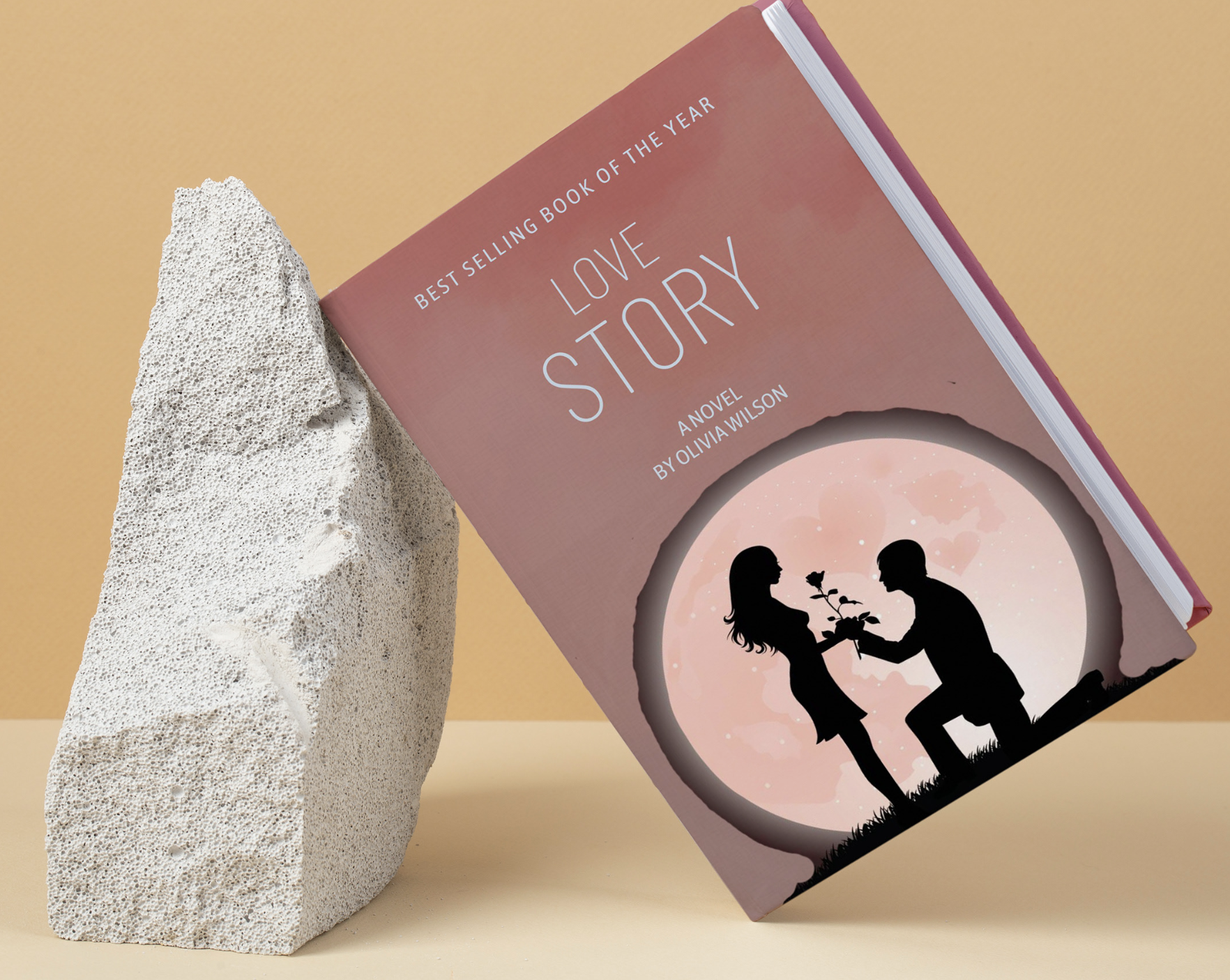}
        }
    \end{minipage} & 
    \begin{minipage}{0.15\textwidth}
        \centering
        \makebox[\textwidth]{ % Ensures both images fit inside the column
            \includegraphics[width=0.7\textwidth]{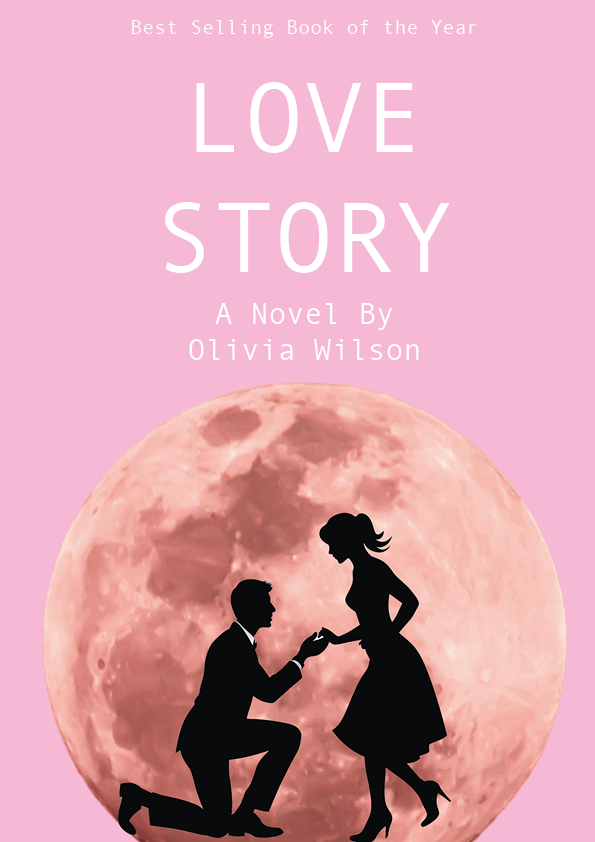}
        }
    \end{minipage} &
    \includegraphics[height=1.0em]{figures/logo/photoshop.png} \includegraphics[height=1.0em]{figures/logo/indesign.png} \\
    \midrule

    \textbf{Business Card} & 
    \begin{minipage}{0.6\textwidth}
        Create a one-sided business card design with a light yellow background for the bookstore`CACTUS'. Replace the `T' in `CACTUS' with a cactus-shaped illustration in green font, centered and add a tagline `Livros Novos e Usados' in green font below the bookstore name.
    \end{minipage} & 
    \begin{minipage}{0.15\textwidth}
        \centering
        \makebox[\textwidth]{ % Ensures both images fit inside the column
            \includegraphics[width=\textwidth]{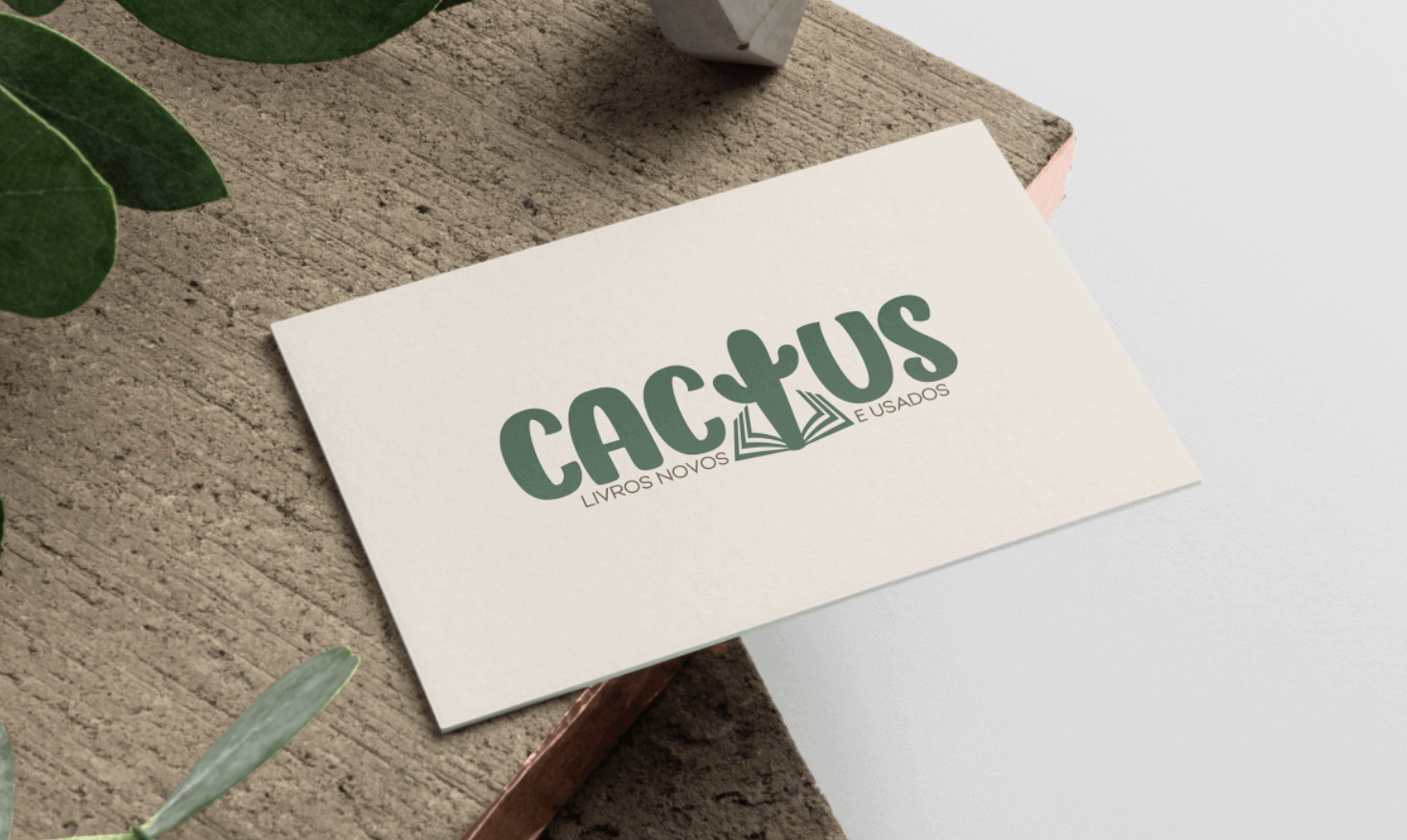}
        }
    \end{minipage} & 
    \begin{minipage}{0.15\textwidth}
        \centering
        \makebox[\textwidth]{ % Ensures both images fit inside the column
            \includegraphics[width=\textwidth]{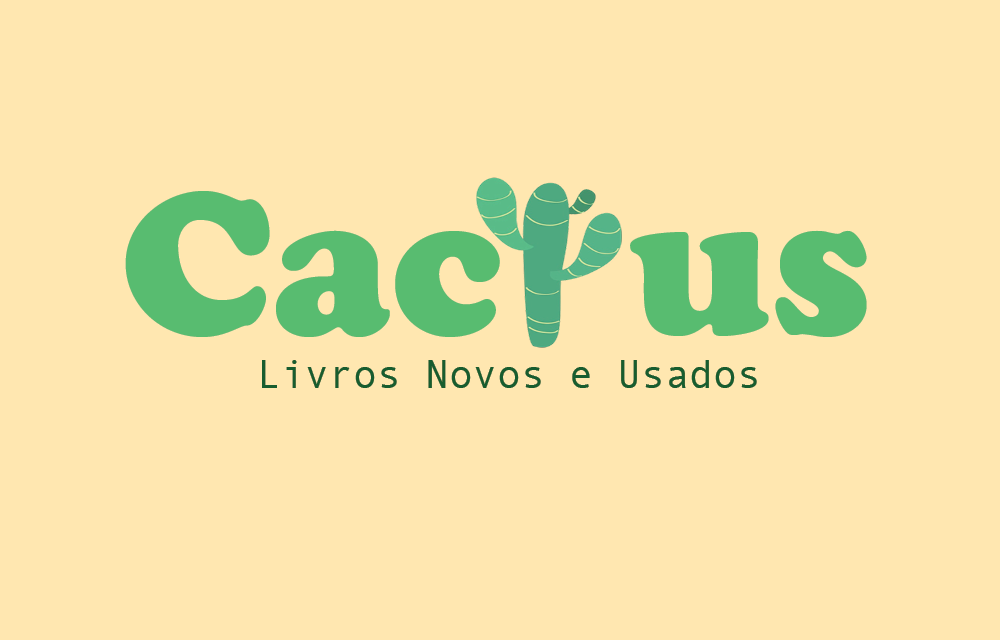}
        }
    \end{minipage} &
    \includegraphics[height=1.0em]{figures/logo/indesign.png} \\
    \midrule

    \textbf{Postcard} & 
    \begin{minipage}{0.6\textwidth}
        Create a postcard design with the message `Think Happy!' in a red, curly font on a floral background featuring a mix of warm-toned roses. Place a semi-transparent white box behind the message.
    \end{minipage} & 
    \begin{minipage}{0.15\textwidth}
        \centering
        \makebox[\textwidth]{ % Ensures both images fit inside the column
            \includegraphics[width=0.7\textwidth]{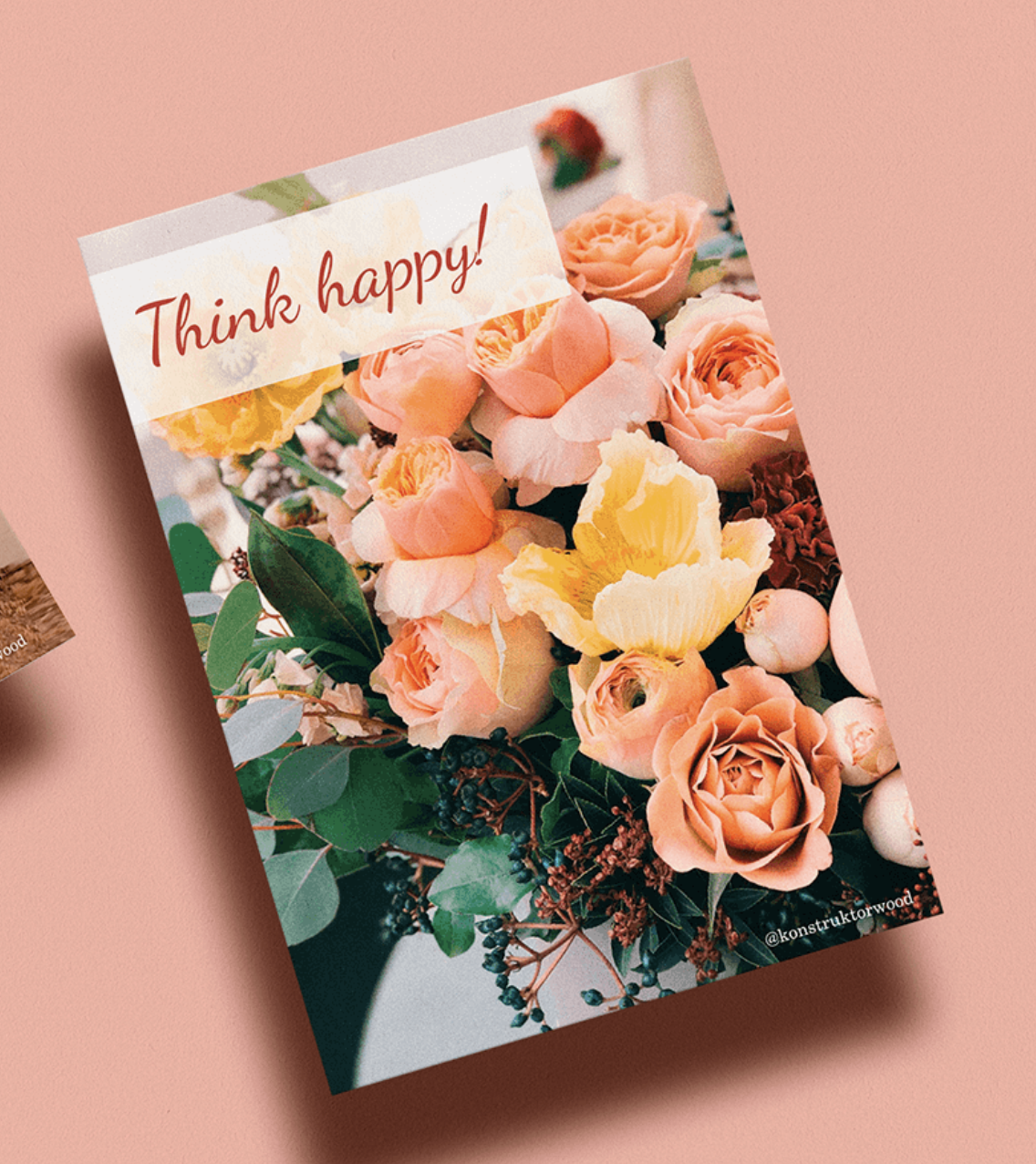}
        }
    \end{minipage} & 
    \begin{minipage}{0.15\textwidth}
        \centering
        \makebox[\textwidth]{ % Ensures both images fit inside the column
            \includegraphics[width=0.6\textwidth]{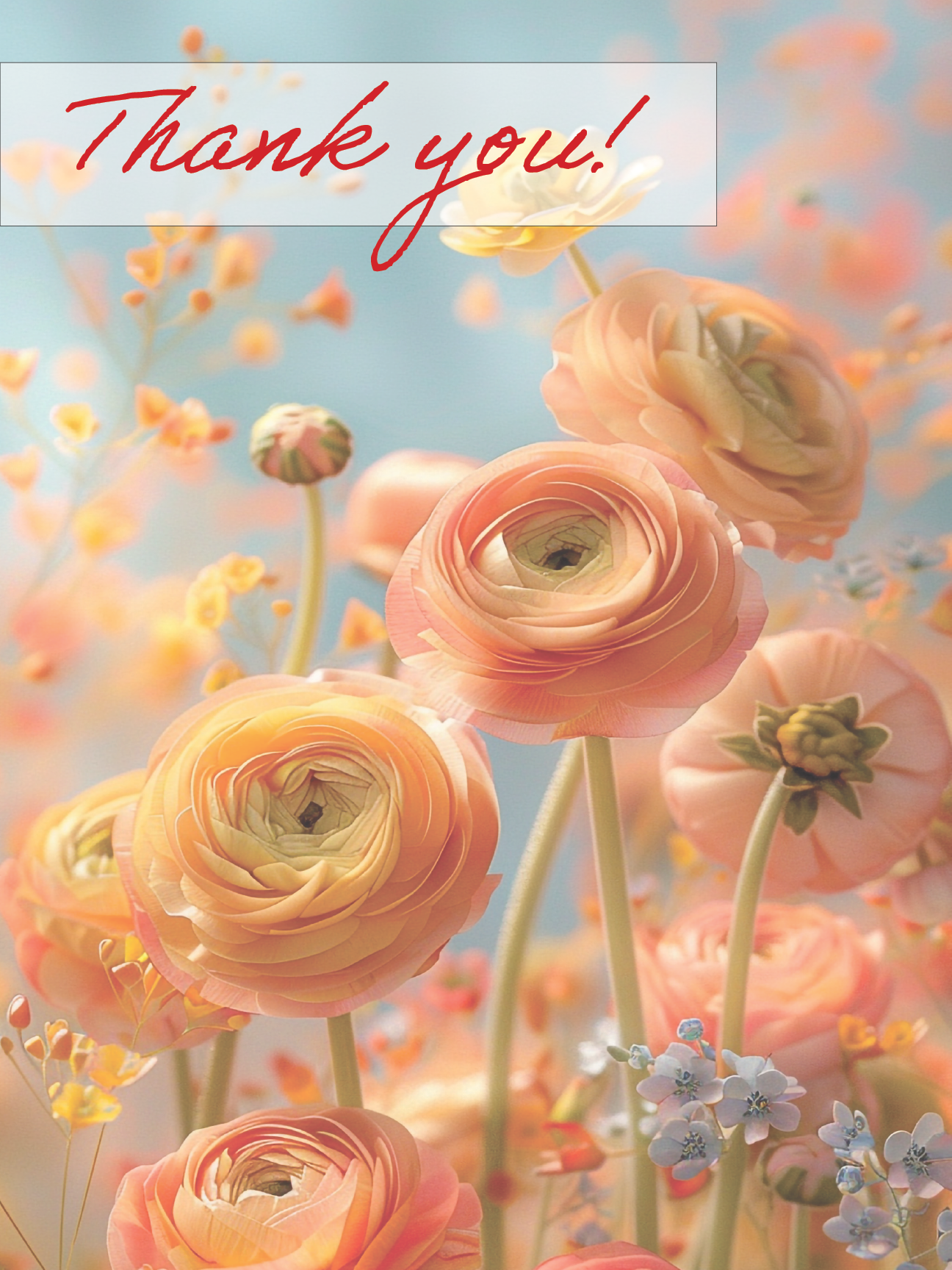}
        }
    \end{minipage} &
    \includegraphics[height=1.0em]{figures/logo/illustrator.png} \includegraphics[height=1.0em]{figures/logo/indesign.png} \\
    \midrule

    \textbf{Poster} & 
    \begin{minipage}{0.6\textwidth}
        Create a poster design with a light yellow background, featuring a large jellyfish illustration centered within a black rectangular box. Add a bold, black title `JELLYFISH' at the top and place a brief informative sentence about jellyfish in white font at the bottom left corner.
    \end{minipage} & 
    \begin{minipage}{0.15\textwidth}
        \centering
        \makebox[\textwidth]{ % Ensures both images fit inside the column
            \includegraphics[width=0.7\textwidth]{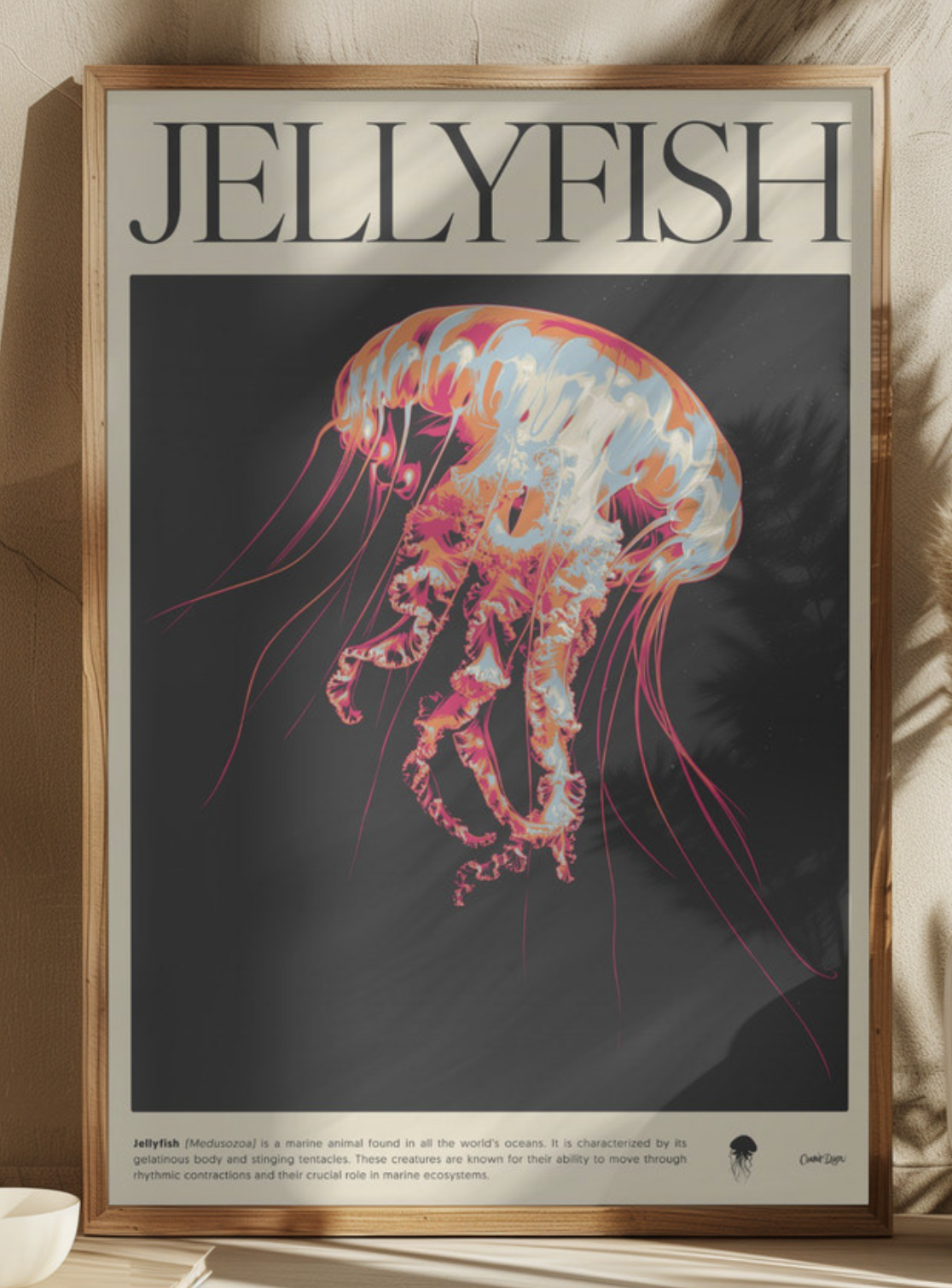}
        }
    \end{minipage} & 
    \begin{minipage}{0.15\textwidth}
        \centering
        \makebox[\textwidth]{ % Ensures both images fit inside the column
            \includegraphics[width=0.7\textwidth]{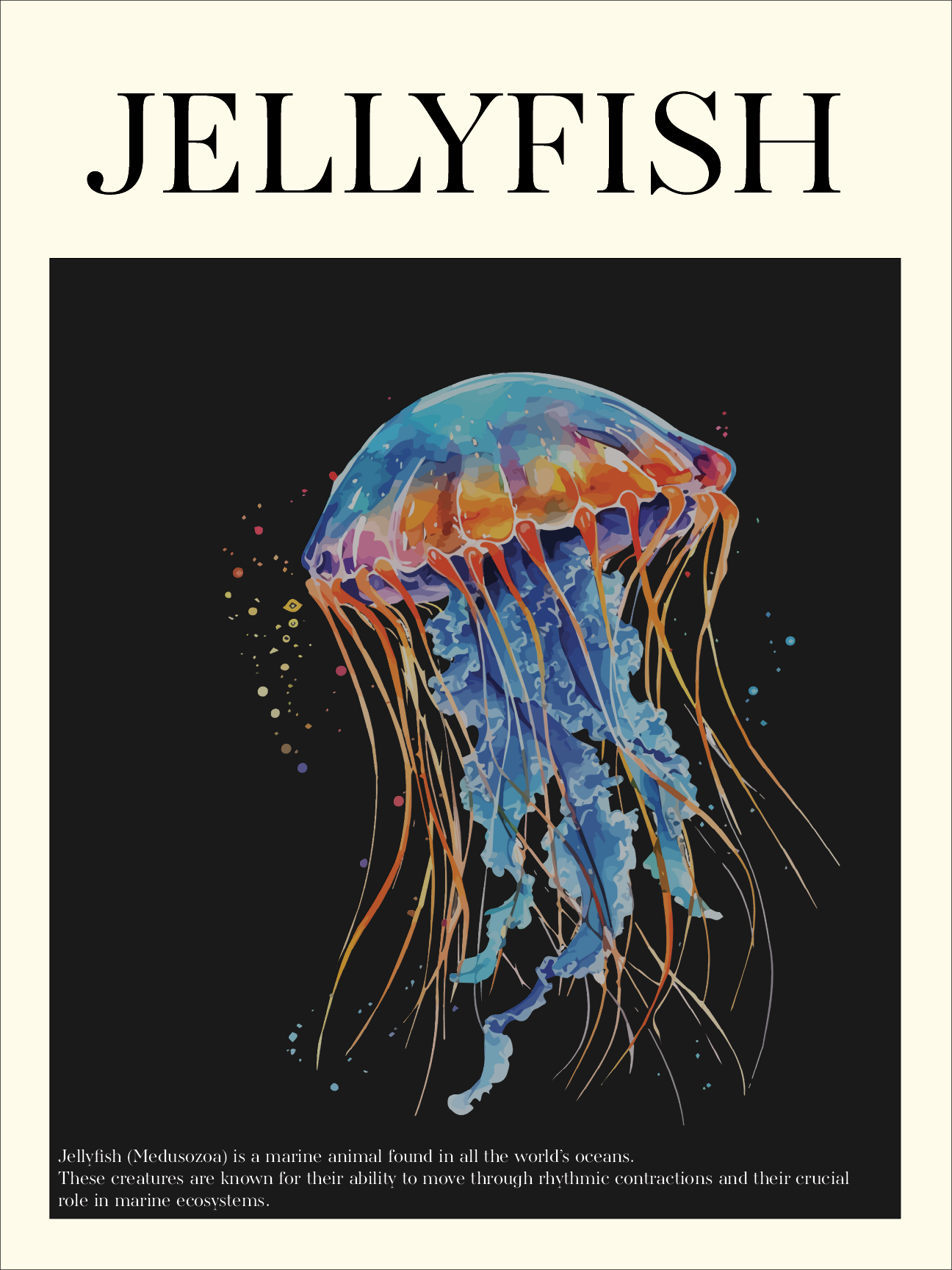}
        }
    \end{minipage} &
    \includegraphics[height=1.0em]{figures/logo/illustrator.png} \includegraphics[height=1.0em]{figures/logo/indesign.png} \\

    \bottomrule
    \end{tabular}
}
\caption{Example of human-annotated user queries and corresponding design outputs along with their reference outputs for each design type. \textbf{Tool:} Adobe CC design tool(s) used to generate the design output. We use the same combination of tools specified by the designers of the reference designs from the Behance platform.}
\label{tab:reference_plan}
\end{table*}

\begin{figure*}

\begin{prompt}[title={Annotated Plan B.1.2: Business Card}]
\begin{lstlisting}[basicstyle=\ttfamily\tiny]
[
    {"id": 1, "expert": "Layout Designer", "description": "Create a new document with business card dimensions.", "action": "CreateDocument", "parameters": {"docType": "business card"}},
    {"id": 2, "expert": "Layout Designer", "description": "Set the background color to ivory.", "action": "SetBackgroundColor", "parameters": {"red": 255, "green": 231, "blue": 176}},
    {"id": 3, "expert": "Layout Designer", "description": "Create text for the first bookstore name: 'Cac  us'.", "action": "CreateText", "parameters": {"layerName": "NameLayer1", "textString": "Cac  us"}},
    {"id": 4, "expert": "Layout Designer", "description": "Apply the Cooper Std Black font to the first name text.", "action": "ApplyFont", "parameters": {"layerName": "NameLayer1", "fontName": "Cooper Std Black"}},
    {"id": 5, "expert": "Layout Designer", "description": "Set the first name text color to green.", "action": "ColorText", "parameters": {"layerName": "NameLayer1", "red": 88, "green": 188, "blue": 112}},
    {"id": 6, "expert": "Layout Designer", "description": "Align the first name text to the center.", "action": "AlignText", "parameters": {"layerName": "NameLayer1", "alignment": "center"}},
    {"id": 7, "expert": "Layout Designer", "description": "Resize the first name text to a large size.", "action": "ResizeText", "parameters": {"layerName": "NameLayer1", "fontSize": 200}},
    {"id": 8, "expert": "Layout Designer", "description": "Reposition the first name text to the center of the document.", "action": "RepositionText", "parameters": {"layerName": "NameLayer1", "posX": -55, "posY": 192}},
    {"id": 9, "expert": "Layout Designer", "description": "Create text for the second bookstore name: 'Cac  us'.", "action": "CreateText", "parameters": {"layerName": "NameLayer2", "textString": "Cac  us"}},
    {"id": 10, "expert": "Layout Designer", "description": "Apply the Cooper Std Black font to the second name text.", "action": "ApplyFont", "parameters": {"layerName": "NameLayer2", "fontName": "Cooper Std Black"}},
    {"id": 11, "expert": "Layout Designer", "description": "Set the second name text color to white.", "action": "ColorText", "parameters": {"layerName": "NameLayer2", "red": 255, "green": 255, "blue": 255}},
    {"id": 12, "expert": "Layout Designer", "description": "Align the second name text to the center.", "action": "AlignText", "parameters": {"layerName": "NameLayer2", "alignment": "center"}},
    {"id": 13, "expert": "Layout Designer", "description": "Resize the second name text to a large size.", "action": "ResizeText", "parameters": {"layerName": "NameLayer2", "fontSize": 200}},
    {"id": 14, "expert": "Layout Designer", "description": "Reposition the second name text to the center of the document, slightly overlapping with the first name text.", "action": "RepositionText", "parameters": {"layerName": "NameLayer2", "posX": -69, "posY": 178}},
    {"id": 15, "expert": "Layout Designer", "description": "Import the cactus-shaped 'T' image from 'cactus_shaped_T.png' and integrate it into the name text.", "action": "ImportObject", "parameters": {"fileName": "cactus_shaped_T.png", "layerName": "CactusTLayer"}},
    {"id": 16, "expert": "Layout Designer", "description": "Resize the cactus-shaped 'T' image to a small size to fit between 'Cac' and 'us'.", "action": "ResizeObject", "parameters": {"layerName": "CactusTLayer", "width": 190, "height": 210}},
    {"id": 17, "expert": "Layout Designer", "description": "Reposition the cactus-shaped 'T' image to be centered, right between 'Cac' and 'us'.", "action": "RepositionObject", "parameters": {"layerName": "CactusTLayer", "posX": 458, "posY": 152}},
    {"id": 18, "expert": "Layout Designer", "description": "Create text for the tagline: 'Livros Novos e Usados'.", "action": "CreateText", "parameters": {"layerName": "TaglineLayer", "textString": "Livros Novos e Usados"}},
    {"id": 19, "expert": "Layout Designer", "description": "Apply the Andale Mono font to the tagline text.", "action": "ApplyFont", "parameters": {"layerName": "TaglineLayer", "fontName": "Andale Mono"}},
    {"id": 20, "expert": "Layout Designer", "description": "Set the tagline text color to green.", "action": "ColorText", "parameters": {"layerName": "TaglineLayer", "red": 88, "green": 188, "blue": 112}},
    {"id": 21, "expert": "Layout Designer", "description": "Align the tagline text to the center.", "action": "AlignText", "parameters": {"layerName": "TaglineLayer", "alignment": "center"}},
    {"id": 22, "expert": "Layout Designer", "description": "Resize the tagline text to a medium size.", "action": "ResizeText", "parameters": {"layerName": "TaglineLayer", "fontSize": 40}},
    {"id": 23, "expert": "Layout Designer", "description": "Reposition the tagline text to the center of the document.", "action": "RepositionText", "parameters": {"layerName": "TaglineLayer", "posX": 160, "posY": 360}},
    {"id": 24, "expert": "Layout Designer", "description": "Export the final business card design as a PNG file.", "action": "SaveDocument", "parameters": {"fileName": "cactus_business_card", "format": "png"}}
]
\end{lstlisting}
\end{prompt}
\caption{Human-annotated workflow plan for the business card example in Table \ref{tab:reference_plan}.}
\label{fig:plan_2}
\end{figure*}
\begin{figure*}

\begin{prompt}[title={Annotated Plan B.1.3: Postcard}]
\begin{lstlisting}[basicstyle=\ttfamily\tiny]
[
    {"id": 1, "expert": "Vector Graphic Editor", "description": "Create a new document with postcard dimensions.", "action": "CreateDocument", "parameters": {"docType": "postcard"}},
    {"id": 2, "expert": "Vector Graphic Editor", "description": "Import the floral background image from 'floral_background.png'.", "action": "ImportObject", "parameters": {"fileName": "floral_background.png", "layerName": "BackgroundLayer"}},
    {"id": 3, "expert": "Vector Graphic Editor", "description": "Resize the floral background image to cover the full background.", "action": "ResizeObject", "parameters": {"layerName": "BackgroundLayer", "width": 1296, "height": 2129}},
    {"id": 4, "expert": "Vector Graphic Editor", "description": "Decrease the opacity of the floral background image to enhance text visibility.", "action": "OpacityObject", "parameters": {"layerName": "BackgroundLayer", "opacity": 80}},
    {"id": 5, "expert": "Vector Graphic Editor", "description": "Draw a white rectangle box.", "action": "DrawRectangle", "parameters": {"layerName": "RectangleLayer", "width": 1034, "height": 233, "red": 255, "green": 255, "blue": 255}},
    {"id": 6, "expert": "Vector Graphic Editor", "description": "Reposition the rectangle to the top left of the document.", "action": "RepositionDrawing", "parameters": {"layerName": "RectangleLayer", "posX": 488, "posY": 196}},
    {"id": 7, "expert": "Vector Graphic Editor", "description": "Decrease the opacity of the rectangle to make it more subtle.", "action": "OpacityDrawing", "parameters": {"layerName": "RectangleLayer", "opacity": 60}},
    {"id": 8, "expert": "Vector Graphic Editor", "description": "Save the document in a format suitable for further editing by the Layout Designer.", "action": "SaveDocument", "parameters": {"fileName": "floral_image_edited", "format": "ai"}},
    {"id": 9, "expert": "Layout Designer", "description": "Create a new document with postcard dimensions.", "action": "CreateDocument", "parameters": {"docType": "postcard"}},
    {"id": 10, "expert": "Layout Designer", "description": "Import the edited image from the Vector Graphic Editor 'floral_image_edited.ai'.", "action": "ImportObject", "parameters": {"fileName": "floral_image_edited.ai", "layerName": "BackgroundLayer"}},
    {"id": 11, "expert": "Layout Designer", "description": "Create text for the message 'Thank you!'.", "action": "CreateText", "parameters": {"layerName": "MessageLayer", "textString": "Thank you!"}},
    {"id": 12, "expert": "Layout Designer", "description": "Apply the Adobe Handwriting Ernie Pro font to the message text.", "action": "ApplyFont", "parameters": {"layerName": "MessageLayer", "fontName": "Adobe Handwriting Ernie Pro"}},
    {"id": 13, "expert": "Layout Designer", "description": "Color the message text to red.", "action": "ColorText", "parameters": {"layerName": "MessageLayer", "red": 210, "green": 35, "blue": 42}},
    {"id": 14, "expert": "Layout Designer", "description": "Align the message text to the left.", "action": "AlignText", "parameters": {"layerName": "MessageLayer", "alignment": "left"}},
    {"id": 15, "expert": "Layout Designer", "description": "Resize the message text to a large size.", "action": "ResizeText", "parameters": {"layerName": "MessageLayer", "fontSize": 160}},
    {"id": 16, "expert": "Layout Designer", "description": "Reposition the message text to the top right corner of the document.", "action": "RepositionText", "parameters": {"layerName": "MessageLayer", "posX": 33, "posY": 40}},
    {"id": 17, "expert": "Layout Designer", "description": "Export the postcard design as a PNG file.", "action": "SaveDocument", "parameters": {"fileName": "floral_postcard", "format": "png"}}
]
\end{lstlisting}
\end{prompt}
\caption{Human-annotated workflow plan for the postcard example in Table \ref{tab:reference_plan}.}
\label{fig:plan_3}
\end{figure*}
\begin{figure*}

\begin{prompt}[title={Annotated Plan B.1.4: Poster}]
\begin{lstlisting}[basicstyle=\ttfamily\tiny]
[
    {"id": 1, "expert": "Vector Graphic Editor", "description": "Create a new document with poster dimensions.", "action": "CreateDocument", "parameters": {"docType": "poster"}},
    {"id": 2, "expert": "Vector Graphic Editor", "description": "Set background color as ivory.", "action": "SetBackgroundColor", "parameters": {"red": 255, "green": 251, "blue": 233}},
    {"id": 3, "expert": "Vector Graphic Editor", "description": "Draw a dark gray rectangle.", "action": "DrawRectangle", "parameters": {"layerName": "RectangleLayer", "width": 1158, "height": 1300, "red": 26, "green": 26, "blue": 26}},
    {"id": 4, "expert": "Vector Graphic Editor", "description": "Reposition the rectangle to be slightly bottom centered.", "action": "RepositionDrawing", "parameters": {"layerName": "RectangleLayer", "posX": 500, "posY": 325}},
    {"id": 5, "expert": "Vector Graphic Editor", "description": "Import the jellyfish illustration image from 'jellyfish_illustration.png'.", "action": "ImportObject", "parameters": {"fileName": "jellyfish_illustration.png", "layerName": "JellyfishLayer"}},
    {"id": 6, "expert": "Vector Graphic Editor", "description": "Resize the jellyfish illustration image to large size.", "action": "ResizeObject", "parameters": {"layerName": "JellyfishLayer", "width": 1110, "height": 1060}},
    {"id": 7, "expert": "Vector Graphic Editor", "description": "Reposition the jellyfish illustration image to be centered in the middle.", "action": "RepositionObject", "parameters": {"layerName": "JellyfishLayer", "posX": 720, "posY": 1000}},
    {"id": 8, "expert": "Vector Graphic Editor", "description": "Decrease the opacity of the jellyfish illustration image to be more dimming.", "action": "OpacityObject", "parameters": {"layerName": "JellyfishLayer", "opacity": 80}},
    {"id": 9, "expert": "Vector Graphic Editor", "description": "Save the document in a format suitable for further editing by the Layout Designer.", "action": "SaveDocument", "parameters": {"fileName": "jellyfish_edited", "format": "ai"}},
    {"id": 10, "expert": "Layout Designer", "description": "Create a new document with poster dimensions.", "action": "CreateDocument", "parameters": {"docType": "poster"}},
    {"id": 11, "expert": "Layout Designer", "description": "Import the edited image from the Vector Graphic Editor 'jellyfish_edited.ai'.", "action": "ImportObject", "parameters": {"fileName": "jellyfish_edited.ai", "layerName": "BackgroundLayer"}},
    {"id": 12, "expert": "Layout Designer", "description": "Create text for the title 'JELLYFISH'.", "action": "CreateText", "parameters": {"layerName": "TitleLayer", "textString": "JELLYFISH"}},
    {"id": 13, "expert": "Layout Designer", "description": "Apply the AWConqueror Std Didot font to the title text.", "action": "ApplyFont", "parameters": {"layerName": "TitleLayer", "fontName": "AWConqueror Std Didot"}},
    {"id": 14, "expert": "Layout Designer", "description": "Color the title text to black.", "action": "ColorText", "parameters": {"layerName": "TitleLayer", "red": 0, "green": 0, "blue": 0}},
    {"id": 15, "expert": "Layout Designer", "description": "Align the title text to be centered.", "action": "AlignText", "parameters": {"layerName": "TitleLayer", "alignment": "center"}},
    {"id": 16, "expert": "Layout Designer", "description": "Resize the title text to a large size.", "action": "ResizeText", "parameters": {"layerName": "TitleLayer", "fontSize": 200}},
    {"id": 17, "expert": "Layout Designer", "description": "Reposition the title text to be in top center.", "action": "RepositionText", "parameters": {"layerName": "TitleLayer", "posX": 28, "posY": 45}},
    {"id": 18, "expert": "Layout Designer", "description": "Create text for the description 'Jellyfish (Medusozoa) is a marine animal found in all the world\\'s oceans.\nThese creatures are known for their ability to move through rhythmic contractions and their crucial role in marine ecosystems.'.", "action": "CreateText", "parameters": {"layerName": "DescriptionLayer", "textString": "Jellyfish (Medusozoa) is a marine animal found in all the world\\'s oceans.\nThese creatures are known for their ability to move through rhythmic contractions and their crucial role in marine ecosystems."}},
    {"id": 19, "expert": "Layout Designer", "description": "Apply the AWConqueror Std Didot font to the description text.", "action": "ApplyFont", "parameters": {"layerName": "DescriptionLayer", "fontName": "AWConqueror Std Didot"}},
    {"id": 20, "expert": "Layout Designer", "description": "Color the description text to white.", "action": "ColorText", "parameters": {"layerName": "DescriptionLayer", "red": 255, "green": 255, "blue": 255}},
    {"id": 21, "expert": "Layout Designer", "description": "Align the description text to be left.", "action": "AlignText", "parameters": {"layerName": "DescriptionLayer", "alignment": "left"}},
    {"id": 22, "expert": "Layout Designer", "description": "Resize the description text to a small size.", "action": "ResizeText", "parameters": {"layerName": "DescriptionLayer", "fontSize": 25}},
    {"id": 23, "expert": "Layout Designer", "description": "Reposition the description text to be in the bottom left corner.", "action": "RepositionText", "parameters": {"layerName": "DescriptionLayer", "posX": 28, "posY": 551}},
    {"id": 24, "expert": "Layout Designer", "description": "Export the poster design as a PDF file.", "action": "SaveDocument", "parameters": {"fileName": "jellyfish_poster", "format": "pdf"}}
]
\end{lstlisting}
\end{prompt}
\caption{Human-annotated workflow plan for the poster example in Table \ref{tab:reference_plan}.}
\label{fig:plan_4}
\end{figure*}

\subsection{Human Annotation}
\label{appendix:human_quality_check}
\begin{figure*}
    \centering
    \includegraphics[width=0.88\linewidth]{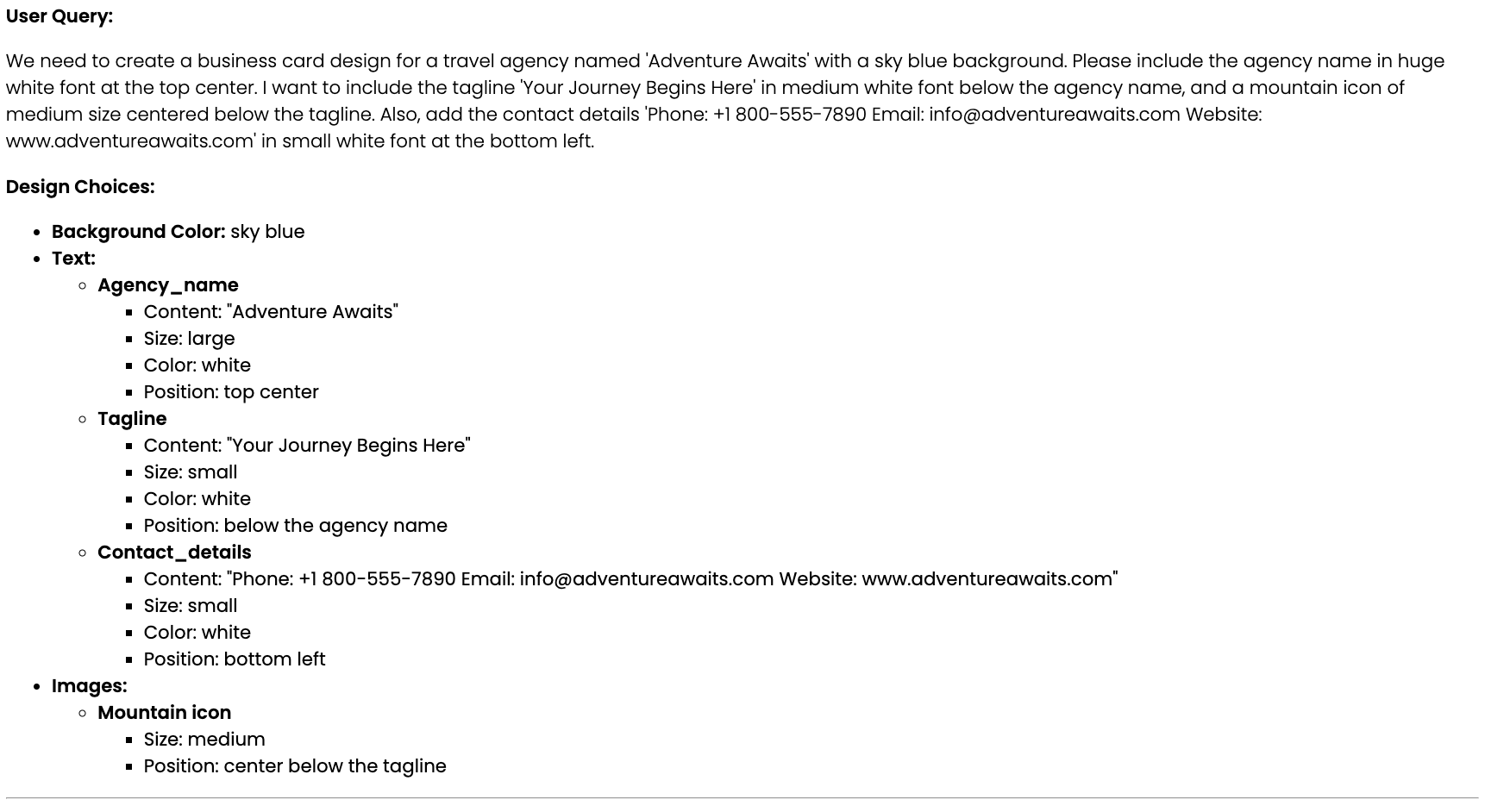}
    \includegraphics[width=0.88\linewidth]{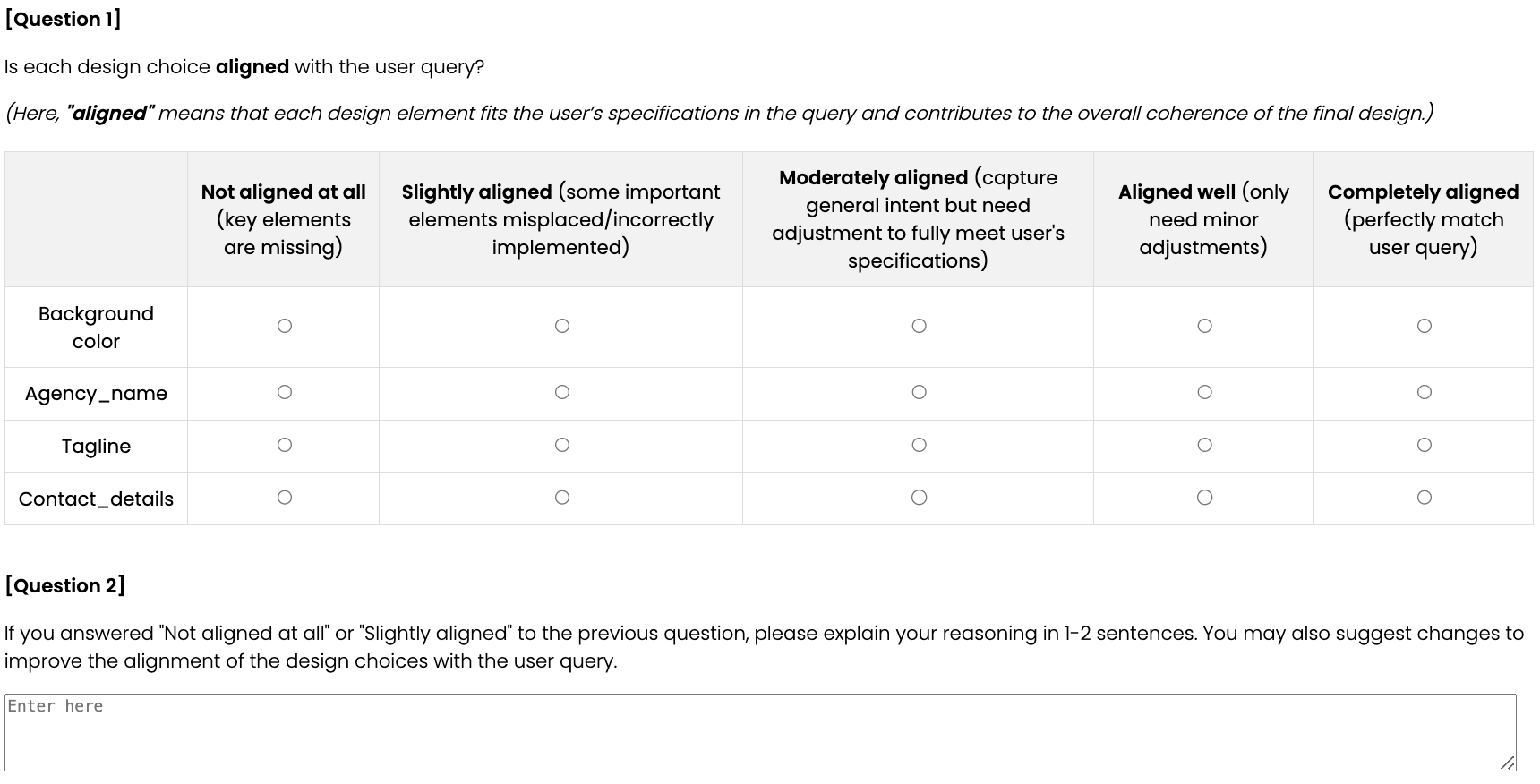}
    \includegraphics[width=0.88\linewidth]{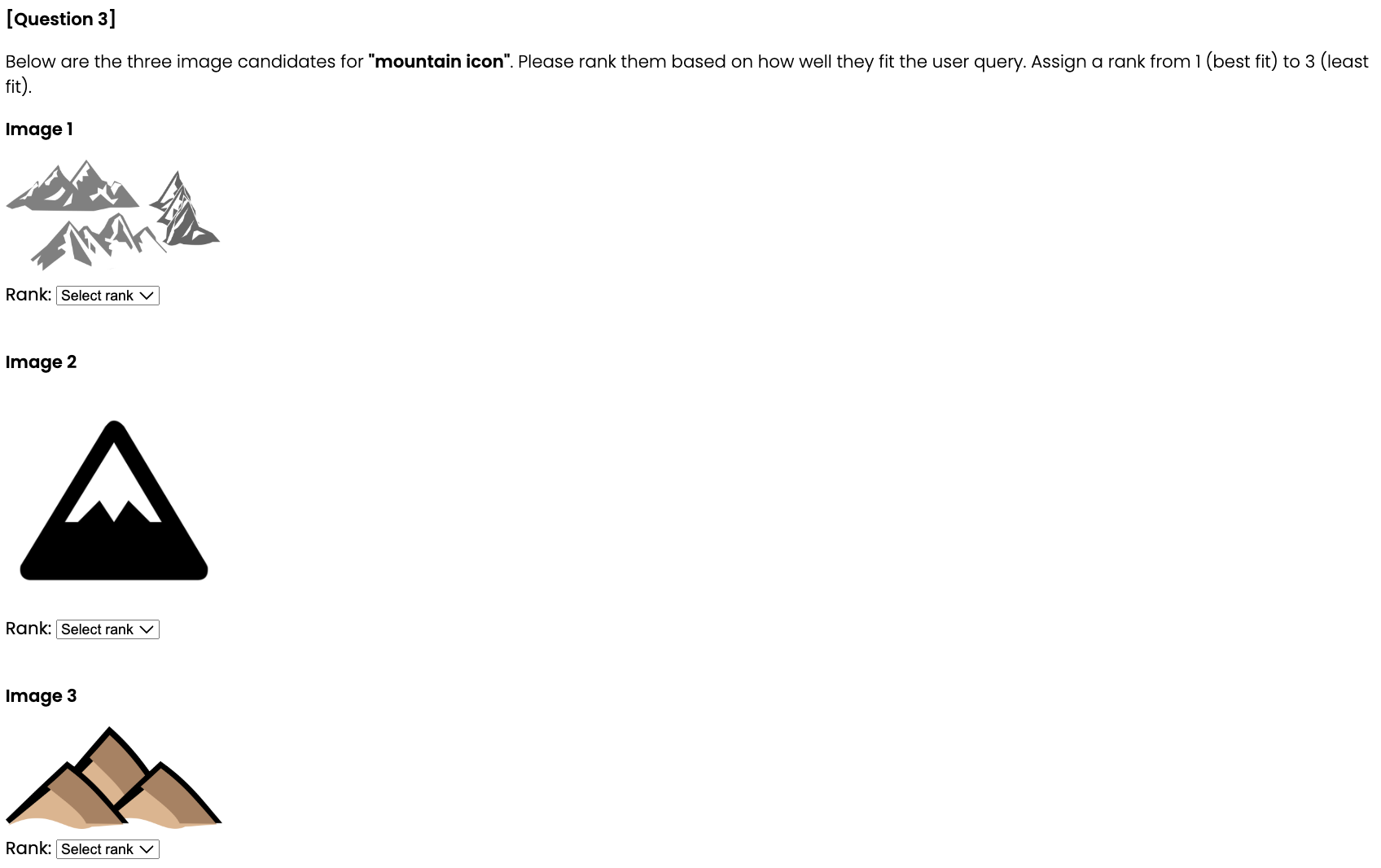}
    \caption{Screenshot of human annotation interface. For each query, annotators are asked to \textbf{1)} evaluate how well each design component aligns with the user query on a 5-point Likert scale and \textbf{2)} rank the 3 image from 1 to 3 based on their relevance to the query. Additionally, they have the option to provide free-form feedback.}
    \label{fig:annotation_interface}
\end{figure*}

We detail the human annotation process as part of constructing \textsc{GraphicBench}. We built our custom annotation interface as illustrated in Figure \ref{fig:annotation_interface}. We invited 8 students to participate and provide a compensation of \$10 gift card each. Before the survey, we show examples of both successful and failed cases to provide some context of annotation standards to annotators.

As part of the pre-survey, annotators were asked two questions on a 5-point Likert scale: \textbf{(1) Design tool usage:} How often do you use design tools in daily work and life? (1: Never, 5: Always) and \textbf{(2) Adobe Creative Cloud application usage:} How familiar are you in using Adobe Creative Cloud Applications (e.g., Photoshop, Illustrator)? (1: Not familiar at all, 5: Extremely familiar). Of the 8 annotators, for design tool usage, 3 responded ``Never'' (Never in the past month), 4 ``Rarely'' (Fewer than once a week), and 1 ``Sometimes'' (two or three times a week). For Adobe Creative Cloud application usage, 3 were ``Not familiar at all'' (have never used it before), 2 ``Slightly familiar'' (have some basic knowledge but have rarely used it), and 3 ``Moderately familiar'' (can perform simple tasks but may need guidance for more complex features).

\begin{table*}
\centering
\resizebox{\linewidth}{!}{%
    \begin{tabular}{ll p{0.5\textwidth} p{0.6\textwidth} ll}
    \toprule
    \textbf{Category} & \textbf{Action} & \textbf{Parameters} & \textbf{Description} & \textbf{Experts}\\
    \toprule
    \multirow{5}{*}{\textbf{Basic}} & \texttt{CreateDocument} & \texttt{docType} & Create new document with pre-defined dimensions. & \includegraphics[height=1.0em]{figures/logo/photoshop.png} \includegraphics[height=1.0em]{figures/logo/illustrator.png} \includegraphics[height=1.0em]{figures/logo/indesign.png} \\

    & \texttt{CreateDocumentCustom} & \texttt{width, height} & Create new document with desired width and height values. & \includegraphics[height=1.0em]{figures/logo/photoshop.png} \includegraphics[height=1.0em]{figures/logo/illustrator.png} \includegraphics[height=1.0em]{figures/logo/indesign.png} \\

    & \texttt{SetBackgroundColor} & \texttt{red, green, blue} & Set the background color to desired RGB color. & \includegraphics[height=1.0em]{figures/logo/photoshop.png} \includegraphics[height=1.0em]{figures/logo/illustrator.png} \includegraphics[height=1.0em]{figures/logo/indesign.png} \\

    & \texttt{SaveDocument} & \texttt{fileName, format} & Save the current document into desired format. & \includegraphics[height=1.0em]{figures/logo/photoshop.png} \includegraphics[height=1.0em]{figures/logo/illustrator.png} \includegraphics[height=1.0em]{figures/logo/indesign.png} \\
    \midrule

    \multirow{22}{*}{\textbf{Drawing}} & \texttt{DrawCircle} & \texttt{layerName, radius, red, green, blue} & Draw a circle of desired radius and RGB color. & \includegraphics[height=1.0em]{figures/logo/illustrator.png} \\

    & \texttt{DrawEllipse} & \texttt{layerName, majorRadius, minorRadius, red, green, blue} & Draw an ellipse of desired radius and RGB color. & \includegraphics[height=1.0em]{figures/logo/illustrator.png} \\

    & \texttt{DrawLine} & \texttt{layerName, startX, startY, endX, endY, strokeWidth, red, green, blue} & Draw a line of desired length, stroke, and RGB color. & \includegraphics[height=1.0em]{figures/logo/illustrator.png} \\

    & \texttt{DrawPolygon} & \texttt{layerName, sides, radius, red, green, blue} & Draw a polygon of desired number of sides, radius, and RGB color. & \includegraphics[height=1.0em]{figures/logo/illustrator.png} \\

    & \texttt{DrawRectangle} & \texttt{layerName, width, height, red, green, blue} & Draw a rectangle of desired size and RGB color. & \includegraphics[height=1.0em]{figures/logo/illustrator.png} \\

    & \texttt{DrawStar} & \texttt{layerName, numPoints, radius, red, green, blue} & Draw a star of desired number of points, radius, and RGB color. & \includegraphics[height=1.0em]{figures/logo/illustrator.png} \\

    & \texttt{DrawTriangle} & \texttt{layerName, base, height, red, green, blue} & Draw a triangle of desired size and RGB color. & \includegraphics[height=1.0em]{figures/logo/illustrator.png} \\
    
    & \texttt{OpacityDrawing} & \texttt{layerName, opacity} & Adjust opacity of a drawing. & \includegraphics[height=1.0em]{figures/logo/illustrator.png} \\

    & \texttt{RemoveDrawing} & \texttt{layerName} & Remove a drawing. & \includegraphics[height=1.0em]{figures/logo/illustrator.png} \\

    & \texttt{RepositionDrawing} & \texttt{layerName, posX, posY} & Reposition a drawing to desired x and y-axis position. & \includegraphics[height=1.0em]{figures/logo/illustrator.png} \\

    & \texttt{ResizeDrawing} & \texttt{layerName, width, height} & Resize a drawing to desired width and height. & \includegraphics[height=1.0em]{figures/logo/illustrator.png} \\

    & \texttt{RotateDrawing} & \texttt{layerName, angle} & Rotate a drawing to desired angle. & \includegraphics[height=1.0em]{figures/logo/illustrator.png} \\

    & \texttt{StorkeDrawing} & \texttt{layerName, strokeWidth, red, green, blue} & Adjust stroke of a drawing with desired width and RGB color. & \includegraphics[height=1.0em]{figures/logo/illustrator.png} \\
    \midrule

    \multirow{15}{*}{\textbf{Text}} & \texttt{AlignText} & \texttt{layerName, alignment} & Align text to desired alignment (left, center, right). & \includegraphics[height=1.0em]{figures/logo/photoshop.png} \includegraphics[height=1.0em]{figures/logo/illustrator.png} \includegraphics[height=1.0em]{figures/logo/indesign.png} \\

    & \texttt{ApplyFont} & \texttt{layerName, fontName} & Apply font to text. & \includegraphics[height=1.0em]{figures/logo/photoshop.png} \includegraphics[height=1.0em]{figures/logo/illustrator.png} \includegraphics[height=1.0em]{figures/logo/indesign.png} \\

    & \texttt{ArrangeText} & \texttt{layerName, arrangement} & Arrange text to desired arrangement (front, frontward, back, backward). & \includegraphics[height=1.0em]{figures/logo/photoshop.png} \includegraphics[height=1.0em]{figures/logo/illustrator.png} \includegraphics[height=1.0em]{figures/logo/indesign.png} \\

    & \texttt{ColorText} & \texttt{layerName, red, green, blue} & Color text to desired RGB color. & \includegraphics[height=1.0em]{figures/logo/photoshop.png} \includegraphics[height=1.0em]{figures/logo/illustrator.png} \includegraphics[height=1.0em]{figures/logo/indesign.png} \\
    
    & \texttt{CreateText} & \texttt{layerName, textString} & Create a new text (default to Arial font). & \includegraphics[height=1.0em]{figures/logo/photoshop.png} \includegraphics[height=1.0em]{figures/logo/illustrator.png} \includegraphics[height=1.0em]{figures/logo/indesign.png} \\

    & \texttt{OpacityText} & \texttt{layerName, opacity} & Adjust opacity of text. & \includegraphics[height=1.0em]{figures/logo/photoshop.png} \includegraphics[height=1.0em]{figures/logo/illustrator.png} \\

    & \texttt{RemoveText} & \texttt{layerName} & Remove text. & \includegraphics[height=1.0em]{figures/logo/photoshop.png} \includegraphics[height=1.0em]{figures/logo/illustrator.png} \includegraphics[height=1.0em]{figures/logo/indesign.png} \\

    & \texttt{RepositionText} & \texttt{layerName, posX, posY} & Reposition text to desired x and y-axis position. & \includegraphics[height=1.0em]{figures/logo/photoshop.png} \includegraphics[height=1.0em]{figures/logo/illustrator.png} \includegraphics[height=1.0em]{figures/logo/indesign.png} \\

    & \texttt{ResizeText} & \texttt{layerName, fontSize} & Resize text to desired font size. & \includegraphics[height=1.0em]{figures/logo/photoshop.png} \includegraphics[height=1.0em]{figures/logo/illustrator.png} \includegraphics[height=1.0em]{figures/logo/indesign.png} \\

    & \texttt{RotateText} & \texttt{layerName, angle} & Rotate text to desired angle. & \includegraphics[height=1.0em]{figures/logo/photoshop.png} \includegraphics[height=1.0em]{figures/logo/illustrator.png} \includegraphics[height=1.0em]{figures/logo/indesign.png} \\

    & \texttt{StrokeText} & \texttt{layerName, strokeWidth, red, green, blue} & Adjust stroke of text with desired width and RGB color. & \includegraphics[height=1.0em]{figures/logo/illustrator.png} \includegraphics[height=1.0em]{figures/logo/indesign.png} \\
    \midrule

    \multirow{25}{*}{\textbf{Object}} & \texttt{ImportObject} & \texttt{fileName, layerName} & Import an image or object from file path. & \includegraphics[height=1.0em]{figures/logo/photoshop.png} \includegraphics[height=1.0em]{figures/logo/illustrator.png} \includegraphics[height=1.0em]{figures/logo/indesign.png} \\

    & \texttt{OpacityObject} & \texttt{fileName, opacity} & Adjust opacity of an object. & \includegraphics[height=1.0em]{figures/logo/photoshop.png} \includegraphics[height=1.0em]{figures/logo/illustrator.png} \\

    & \texttt{RemoveObject} & \texttt{fileName} & Remove an object. & \includegraphics[height=1.0em]{figures/logo/photoshop.png} \includegraphics[height=1.0em]{figures/logo/illustrator.png} \includegraphics[height=1.0em]{figures/logo/indesign.png} \\

    & \texttt{RepositionObject} & \texttt{fileName, posX, posY} & Reposition an object to desired x and y-axis position. & \includegraphics[height=1.0em]{figures/logo/photoshop.png} \includegraphics[height=1.0em]{figures/logo/illustrator.png} \includegraphics[height=1.0em]{figures/logo/indesign.png} \\

    & \texttt{ResizeObject} & \texttt{fileName, width, height} & Resize an object to desired width and height. & \includegraphics[height=1.0em]{figures/logo/photoshop.png} \includegraphics[height=1.0em]{figures/logo/illustrator.png} \includegraphics[height=1.0em]{figures/logo/indesign.png} \\

    & \texttt{RotateObject} & \texttt{fileName, angle} & Rotate an object to desired angle. & \includegraphics[height=1.0em]{figures/logo/photoshop.png} \includegraphics[height=1.0em]{figures/logo/illustrator.png} \includegraphics[height=1.0em]{figures/logo/indesign.png} \\

    & \texttt{GenerateQRObject} & \texttt{layerName, linkURL} & Generate a QR code with desired URL embedded. & \includegraphics[height=1.0em]{figures/logo/indesign.png} \\

    & \texttt{AdjustBC} & \texttt{layerName, brightness, contrast} & Adjust brightness and contrast level of an object. & \includegraphics[height=1.0em]{figures/logo/photoshop.png} \\

    & \texttt{AdjustBW} & \texttt{layerName} & Change an object to black \& white. & \includegraphics[height=1.0em]{figures/logo/photoshop.png} \\

    & \texttt{AdjustHSL} & \texttt{layerName} & Adjust hue, saturation, and lightness level of an object. & \includegraphics[height=1.0em]{figures/logo/photoshop.png} \\

    & \texttt{BlurObject} & \texttt{layerName, blurAmount} & Blur an object to desired amount. & \includegraphics[height=1.0em]{figures/logo/photoshop.png} \\

    & \texttt{PhotoFilter} & \texttt{layerName, filterType, density} & Apply a photo filter to an object with desired density. & \includegraphics[height=1.0em]{figures/logo/photoshop.png} \\

    & \texttt{GlassFilter} & \texttt{layerName, distortion, smoothness, scaling} & Apply a glass filter to an object with the specified parameters. & \includegraphics[height=1.0em]{figures/logo/photoshop.png} \\

    & \texttt{GlowFilter} & \texttt{layerName, graininess, glowAmount, clearAmount} & Apply a glow filter to an object with the specified parameters. & \includegraphics[height=1.0em]{figures/logo/photoshop.png} \\

    & \texttt{OceanRippleFilter} & \texttt{layerName, rippleSize, rippleMagnitude} & Apply an ocean ripple filter to an object with the specified parameters. & \includegraphics[height=1.0em]{figures/logo/photoshop.png} \\

    & \texttt{StainedGlassFilter} & \texttt{layerName, cellSize, borderThickness, lightIntensity} & Apply a stained glass filter to an object with the specified parameters. & \includegraphics[height=1.0em]{figures/logo/photoshop.png} \\

    & \texttt{PatchWorkFilter} & \texttt{layerName, squareSize, relief} & Apply a patchwork filter to an object with the specified parameters. & \includegraphics[height=1.0em]{figures/logo/photoshop.png} \\

    & \texttt{WatercolorFilter} & \texttt{layerName, brushDetail, shadowIntensity, texture} & Apply a watercolor filter to an object with the specified parameters. & \includegraphics[height=1.0em]{figures/logo/photoshop.png} \\

    \bottomrule
    \end{tabular}
}
\caption{Complete list of available actions in \textsc{GraphicTown}. Each action requires specific parameters for execution. \textbf{Experts:} The expert agent(s) which supports the execution of a specific action. For numerical parameters, we provide reference ranges (e.g., \texttt{angle} as [0, 360], \texttt{brightness} as [-150, +150]). For parameters in filter-related functions, we provide a short description for each (e.g., \texttt{rippleSize} in \texttt{OcenRippleFilter} means ``\textit{the size of the ripples created, where lower means smaller and finer ripples that creates more subtle water effect}'').}
\label{tab:action_full_list}
\end{table*}

\section{Detailed Results}
\subsection{Workflow Plan Evaluation}
\label{appendix:detailed_res1}
We provide the full numerical results by model and design type in Table \ref{tab:detailed_res1}. Each component of the design pass rate (color, text, and images) is later normalized to the range [0,1] in Figure \ref{fig:main_figure}. Our results show that most tested models perform well in expert use efficiency and design pass rate, while larger models outperform smaller ones in delivery rate.

\begin{table*}
\centering
\resizebox{\linewidth}{!}{%
    \begin{tabular}{lllllllllllll}
    \toprule
    \textbf{Model} & \textbf{Design Type} & \textbf{Delivery Rate} & \textbf{Step Eff.} & \textbf{Expert Use Eff.} & \textbf{Design Pass Rate} & \textbf{Color} & \textbf{Text} & \textbf{Image} \\
    \toprule
    
    \multirow{4}{*}{\textbf{\textsc{LLaMA-3 8b}}} & Book Cover & 0.092 & 0.941 & 1.00 & 0.938 & 4.508 & 4.773 & 4.792 \\
    & Business Card & 0.148 & 0.922 & 1.00 & 0.948 & 4.813 & 4.542 & 4.867 \\
    & Postcard & 0.473 & 0.946 & 1.00 & 0.943 & 4.650 & 4.769 & 4.727 \\
    & Poster & 0.196 & 0.955 & 1.00 & 0.947 & 4.667 & 4.702 & 4.830 \\
    \midrule
    
    \multirow{4}{*}{\textbf{\textsc{Gemma-2 9b}}} & Book Cover & 0.173 & 0.939 & 1.00 & 0.940 & 4.765 & 4.812 & 4.519 \\
    & Business Card & 0.714 & 0.963 & 1.00 & 0.866 & 4.433 & 4.084 & 4.473 \\
    & Postcard & 0.835 & 0.936 & 1.00 & 0.954 & 4.877 & 4.850 & 4.585 \\
    & Poster & 0.533 & 0.942 & 1.00 & 0.930 & 4.765 & 4.551 & 4.628 \\
    \midrule
    
    \multirow{4}{*}{\textbf{\textsc{Gemma-2 27b}}} & Book Cover & 0.842 & 0.951 & 1.00 & 0.332 & 1.623 & 1.688 & 1.665 \\
    & Business Card & 0.478 & 0.936 & 1.00 & 0.917 & 4.670 & 4.626 & 4.453 \\
    & Postcard & 0.954 & 0.926 & 1.00 & 0.933 & 4.677 & 4.746 & 4.577 \\
    & Poster & 0.622 & 0.959 & 1.00 & 0.895 & 4.512 & 4.554 & 4.360 \\
    \midrule
    
    \multirow{4}{*}{\textbf{\textsc{Qwen-2.5 7b}}} & Book Cover & 0.358 & 0.947 & 1.00 & 0.859 & 4.212 & 4.342 & 4.331 \\
    & Business Card & 0.345 & 0.948 & 1.00 & 0.888 & 4.650 & 3.990 & 4.685 \\
    & Postcard & 0.669 & 0.982 & 1.00 & 0.873 & 4.285 & 4.388 & 4.419 \\
    & Poster & 0.562 & 0.949 & 1.00 & 0.881 & 4.461 & 4.158 & 4.592 \\
    \midrule
    
    \multirow{4}{*}{\textbf{\textsc{Qwen-2.5 14b}}} & Book Cover & 0.262 & 0.879 & 1.00 & 0.968 & 4.700 & 4.900 & 4.923 \\
    & Business Card & 0.695 & 0.909 & 1.00 & 0.930 & \textbf{4.926} & 4.172 & 4.847 \\
    & Postcard & 0.792 & 0.933 & 1.00 & \textbf{0.972} & 4.850 & \textbf{4.885} & 4.838 \\
    & Poster & 0.637 & 0.920 & 1.00 & 0.964 & 4.905 & 4.604 & \textbf{4.946} \\
    \midrule
    
    \multirow{4}{*}{\textbf{\textsc{GPT-3.5}}} & Book Cover & 0.873 & \textbf{0.985} & 1.00 & 0.917 & 4.527 & 4.654 & 4.573 \\
    & Business Card & \textbf{0.966} & \textbf{0.985} & 1.00 & 0.859 & 4.433 & 3.764 & 4.695 \\
    & Postcard & 0.954 & 0.980 & 1.00 & 0.909 & 4.596 & 4.419 & 4.623 \\
    & Poster & 0.938 & 0.983 & 1.00 & 0.881 & 4.500 & 4.196 & 4.518 \\
    
    \bottomrule
    \end{tabular}
}
\caption{Workflow plan evaluation results per model and design type. The range for Delivery Rate, Expert Use Efficiency, and Design Pass Rate are [0,1] while the ranges for Color, Text, Visual Pass Rates are [1,5]. Best scores for each column is \textbf{bold}.}
\label{tab:detailed_res1}
\end{table*}

\subsection{Expert Distribution}
\label{appendix:detailed_res2}
In Table \ref{tab:detailed_res2}, we show detailed results on expert recruitment ratios and workload distribution across models and design types. The expert recruitment ratio is computed as the number of times an expert is recruited per user query, divided by the total number of queries. Workload distribution represented the average number of steps assigned to each expert agent in the workflow plan $W_s$. Additionally, we present the distribution of expert agent usage sequences in Table \ref{tab:detailed_res22}. The most common order of sequence across models is Photo Editor → Layout Designer with the highest number of occurrence (4,207 in total), followed by Layout Designer → Photo Editor (504) and Photo Editor → Vector Graphic Editor (262).

\begin{table*}
\centering
\resizebox{0.9\linewidth}{!}{%
    \begin{tabular}{l l l lllllllll} 
    \toprule
    \multirow{2}{*}{\textbf{Model}} & \multirow{2}{*}{\textbf{Design Type}} & \multicolumn{3}{c}{\textbf{Ratio}} & \multicolumn{3}{c}{\textbf{Workload}} & \multirow{2}{*}{\textbf{Avg. \# Agents}} & \multirow{2}{*}{\textbf{Avg. \# Steps}} \\
    \cmidrule[1pt]{3-5}
    \cmidrule[1pt]{6-8}
    & & \includegraphics[height=1.0em]{figures/logo/photoshop.png} & \includegraphics[height=1.0em]{figures/logo/illustrator.png} & \includegraphics[height=1.0em]{figures/logo/indesign.png} & \includegraphics[height=1.0em]{figures/logo/photoshop.png} & \includegraphics[height=1.0em]{figures/logo/illustrator.png} & \includegraphics[height=1.0em]{figures/logo/indesign.png} & & \\
    \toprule
    
    \multirow{4}{*}{\textbf{\textsc{LLaMA-3 8b}}} & Book Cover & 1.00 & 0.02 & 1.00 & 11.8 & 9.67 & 18.3 & 2.15 & 31.3 \\
    & Business Card & 1.00 & 0.01 & 1.00 & 10.6 & 18.0 & 14.6 & 2.46 & 26.0 \\
    & Postcard & 1.00 & 0.00 & 1.00 & 10.8 & 6.00 & 11.1 & 2.12 & 20.2 \\
    & Poster & 1.00 & 0.00 & 1.00 & 9.84 & 10.0 & 14.1 & 2.12 & 24.5 \\
    \midrule
    
    \multirow{4}{*}{\textbf{\textsc{Gemma-2 9b}}} & Book Cover & 1.00 & 0.00 & 1.00 & 8.55 & - & 18.3 & 2.00 & 23.6 \\
    & Business Card & 1.00 & 0.03 & 1.00 & 5.18 & 10.6 & 12.9 & 2.04 & 17.2 \\
    & Postcard & 1.00 & 0.00 & 1.00 & 7.63 & - & 11.7 & 2.00 & 16.5 \\
    & Poster & 1.00 & 0.03 & 1.00 & 6.44 & 9.86 & 13.6 & 2.03 & 18.3 \\
    \midrule
    
    \multirow{4}{*}{\textbf{\textsc{Gemma-2 27b}}} & Book Cover & 1.00 & 0.00 & 1.00 & 7.31 & - & 21.0 & 2.00 & 23.2 \\
    & Business Card & 1.00 & 0.00 & 1.00 & 5.15 & - & 15.9 & 2.01 & 18.4 \\
    & Postcard & 1.00 & 0.00 & 1.00 & 7.04 & - & 11.5 & 2.00 & 13.8 \\
    & Poster & 1.00 & 0.00 & 1.00 & 6.29 & - & 15.4 & 2.00 & 14.6 \\
    \midrule
    
    \multirow{4}{*}{\textbf{\textsc{Qwen-2.5 7b}}} & Book Cover & 0.85 & 0.52 & 0.54 & 10.9 & 11.9 & 12.3 & 1.92 & 23.6 \\
    & Business Card & 0.61 & 0.47 & 0.61 & 8.75 & 10.2 & 11.1 & 1.68 & 17.8 \\
    & Postcard & 0.37 & 0.15 & 0.85 & 9.36 & 10.5 & 9.77 & 1.38 & 11.2 \\
    & Poster & 0.80 & 0.75 & 0.43 & 8.17 & 10.2 & 10.4 & 1.97 & 18.0 \\
    \midrule
    
    \multirow{4}{*}{\textbf{\textsc{Qwen-2.5 14b}}} & Book Cover & 1.00 & 0.00 & 1.00 & 10.4 & - & 13.9 & 2.00 & 21.7 \\
    & Business Card & 1.00 & 0.00 & 1.00 & 7.09 & - & 12.3 & 2.00 & 17.8 \\
    & Postcard & 1.00 & 0.00 & 1.00 & 8.71 & - & 10.4 & 2.00 & 16.2 \\
    & Poster & 1.00 & 0.00 & 1.00 & 7.57 & - & 12.2 & 2.00 & 19.0 \\
    \midrule
    
    \multirow{4}{*}{\textbf{\textsc{GPT-3.5}}} & Book Cover & 0.99 & 0.01 & 1.00 & 6.31 & 7.50 & 8.73 & 2.01 & 15.0 \\
    & Business Card & 1.00 & 0.01 & 1.00 & 5.17 & 6.00 & 7.26 & 2.01 & 12.1 \\
    & Postcard & 0.99 & 0.01 & 1.00 & 5.88 & 6.00 & 6.51 & 2.01 & 11.8 \\
    & Poster & 1.00 & 0.01 & 1.00 & 5.10 & 6.17 & 7.20 & 2.02 & 12.5 \\
    
    \bottomrule
    \end{tabular}
}
\caption{Expert recruitment ratios (\textbf{Ratio}) and workload distribution (\textbf{Workload}) per model and design type. \textbf{Avg. \# Agents:} Average number of expert agents recruited per user query; \textbf{Avg. \# Steps:} Average number of steps in the workflow plan per user query; \includegraphics[height=1.0em]{figures/logo/photoshop.png}: Photo Editor; \includegraphics[height=1.0em]{figures/logo/illustrator.png}: Vector Graphic Editor; \includegraphics[height=1.0em]{figures/logo/indesign.png}: Layout Designer.}
\label{tab:detailed_res2}
\end{table*}
\begin{table*}
\centering
\resizebox{\linewidth}{!}{%
    \begin{tabular}{l l l lllllllll} 
    \toprule
    \textbf{\# of Agents} & \textbf{Sequence} & \textbf{\textsc{LLaMA-3 8b}} & \textbf{\textsc{Gemma-2 9b}} & \textbf{\textsc{Gemma-2 27b}} & \textbf{\textsc{Qwen-2.5 7b}} & \textbf{\textsc{Qwen-2.5 14b}} & \textbf{\textsc{GPT-3.5}} \\
    \toprule

    \multirow{3}{*}{\textbf{1}} & \includegraphics[height=1em]{figures/logo/photoshop.png} & 0 & 0 & 1 & 59 & 30 & 10 \\
    & \includegraphics[height=1em]{figures/logo/illustrator.png} & 0 & 0 & 0 & 27 & 0 & 0 \\
    & \includegraphics[height=1em]{figures/logo/indesign.png} & 0 & 1 & 0 & 193 & 18 & 3 \\
    \midrule
    
    \multirow{5}{*}{\textbf{2}} & \includegraphics[height=1em]{figures/logo/photoshop.png} → \includegraphics[height=1em]{figures/logo/illustrator.png} & 0 & 0 & 0 & 262 & 0 & 0 \\
    & \includegraphics[height=1em]{figures/logo/photoshop.png} → \includegraphics[height=1em]{figures/logo/indesign.png} & 647 & 922 & 622 & 171 & 914 & 931 \\
    & \includegraphics[height=1em]{figures/logo/illustrator.png} → \includegraphics[height=1em]{figures/logo/photoshop.png} & 0 & 0 & 1 & 24 & 0 & 0 \\
    & \includegraphics[height=1em]{figures/logo/illustrator.png} → \includegraphics[height=1em]{figures/logo/indesign.png} & 0 & 0 & 0 & 9 & 0 & 0 \\
    & \includegraphics[height=1em]{figures/logo/indesign.png} → \includegraphics[height=1em]{figures/logo/photoshop.png} & 305 & 4 & 1 & 88 & 75 & 31 \\
    \midrule
    
    \multirow{5}{*}{\textbf{3}} & \includegraphics[height=1em]{figures/logo/photoshop.png} → \includegraphics[height=1em]{figures/logo/illustrator.png} → \includegraphics[height=1em]{figures/logo/indesign.png} & 1 & 0 & 0 & 35 & 0 & 0 \\
    & \includegraphics[height=1em]{figures/logo/photoshop.png} → \includegraphics[height=1em]{figures/logo/indesign.png} → \includegraphics[height=1em]{figures/logo/illustrator.png} & 0 & 11 & 0 & 0 & 0 & 11 \\
    & \includegraphics[height=1em]{figures/logo/illustrator.png} → \includegraphics[height=1em]{figures/logo/photoshop.png} → \includegraphics[height=1em]{figures/logo/indesign.png} & 5 & 0 & 1 & 4 & 0 & 0 \\
    & \includegraphics[height=1em]{figures/logo/illustrator.png} → \includegraphics[height=1em]{figures/logo/indesign.png} → \includegraphics[height=1em]{figures/logo/photoshop.png} & 0 & 0 & 0 & 1 & 0 & 0 \\
    & \includegraphics[height=1em]{figures/logo/indesign.png} → \includegraphics[height=1em]{figures/logo/photoshop.png} → \includegraphics[height=1em]{figures/logo/illustrator.png} & 0 & 0 & 0 & 0 & 0 & 1 \\
    
    \bottomrule
    \end{tabular}
}
\caption{Distribution of expert agent usage sequence per model. \includegraphics[height=1em]{figures/logo/photoshop.png}: Photo Editor; \includegraphics[height=1em]{figures/logo/illustrator.png}: Vector Graphic Editor; \includegraphics[height=1em]{figures/logo/indesign.png}: Layout Designer.}
\label{tab:detailed_res22}
\end{table*}

\subsection{Action Distribution}
\label{appendix:detailed_res3}
\begin{figure*}[t]
    \centering
    \begin{subfigure}{0.3\textwidth}
        \centering
        \includegraphics[height=3cm]{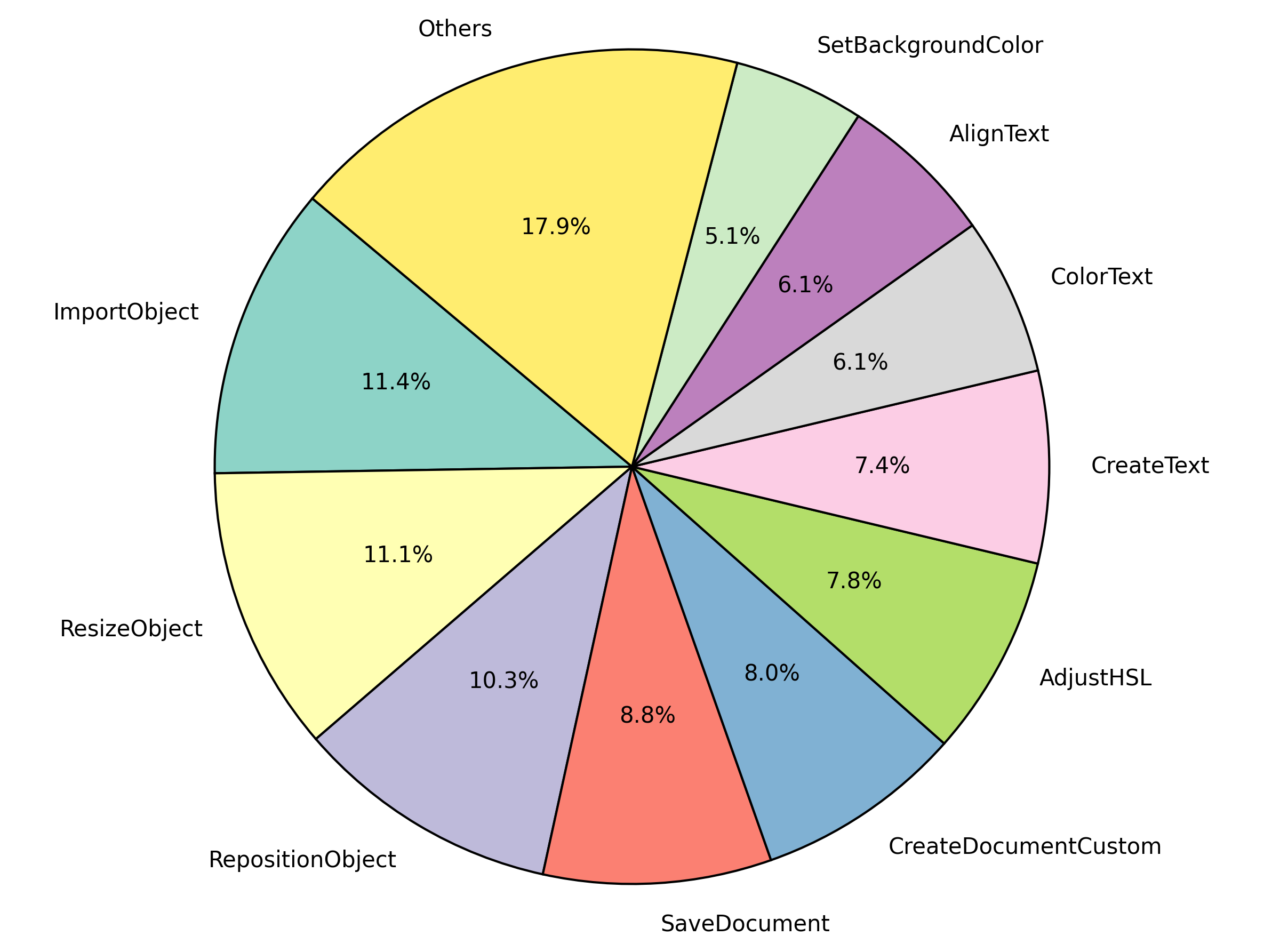}
        \caption{\textsc{LLaMA-3 8b} \includegraphics[height=1em]{figures/logo/photoshop.png}}
    \end{subfigure}
    \hfill
    \begin{subfigure}{0.3\textwidth}
        \centering
        \includegraphics[height=3cm]{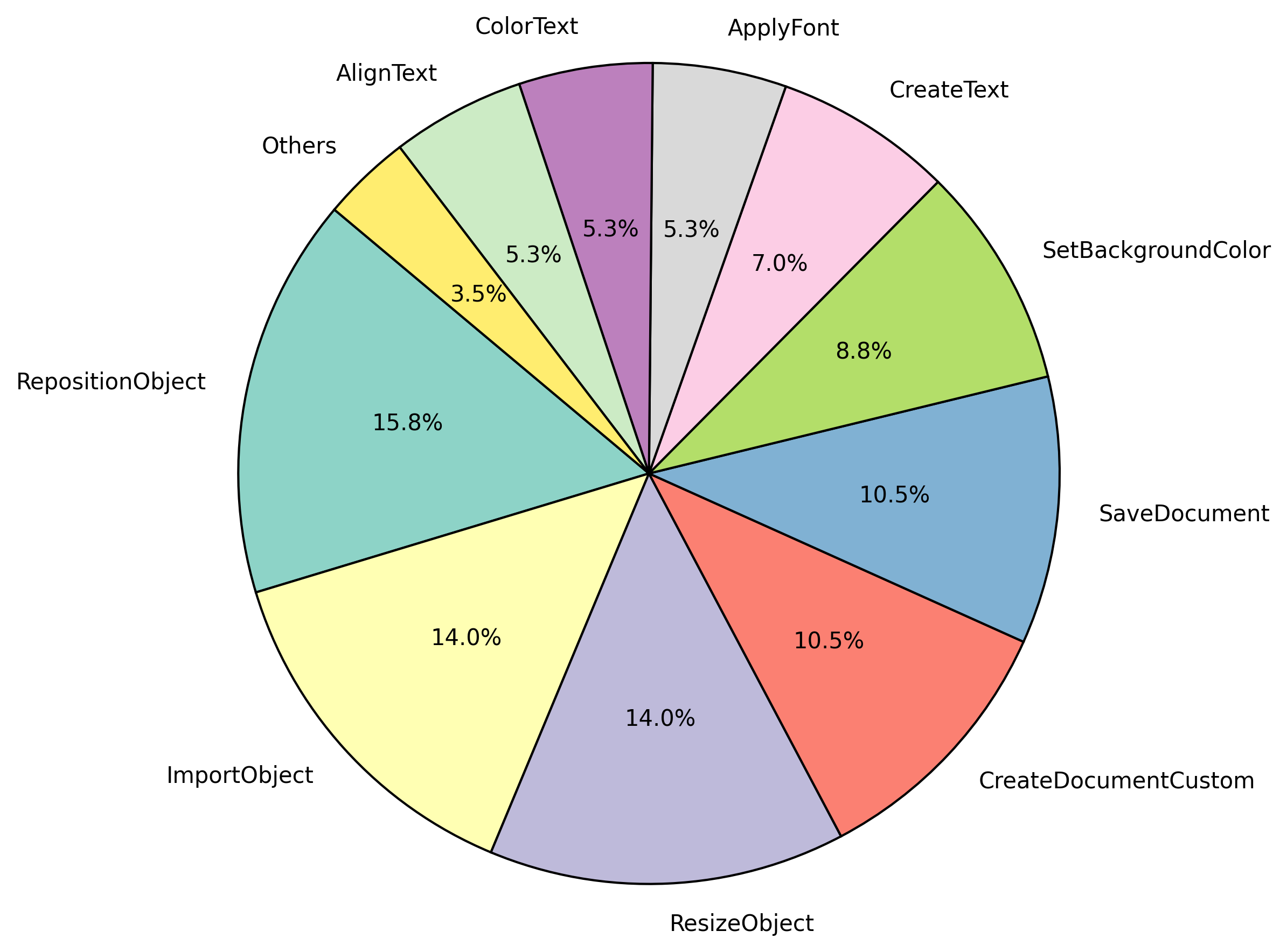}
        \caption{\textsc{LLaMA-3 8b} \includegraphics[height=1em]{figures/logo/illustrator.png}}
    \end{subfigure}
    \hfill
    \begin{subfigure}{0.3\textwidth}
        \centering
        \includegraphics[height=3cm]{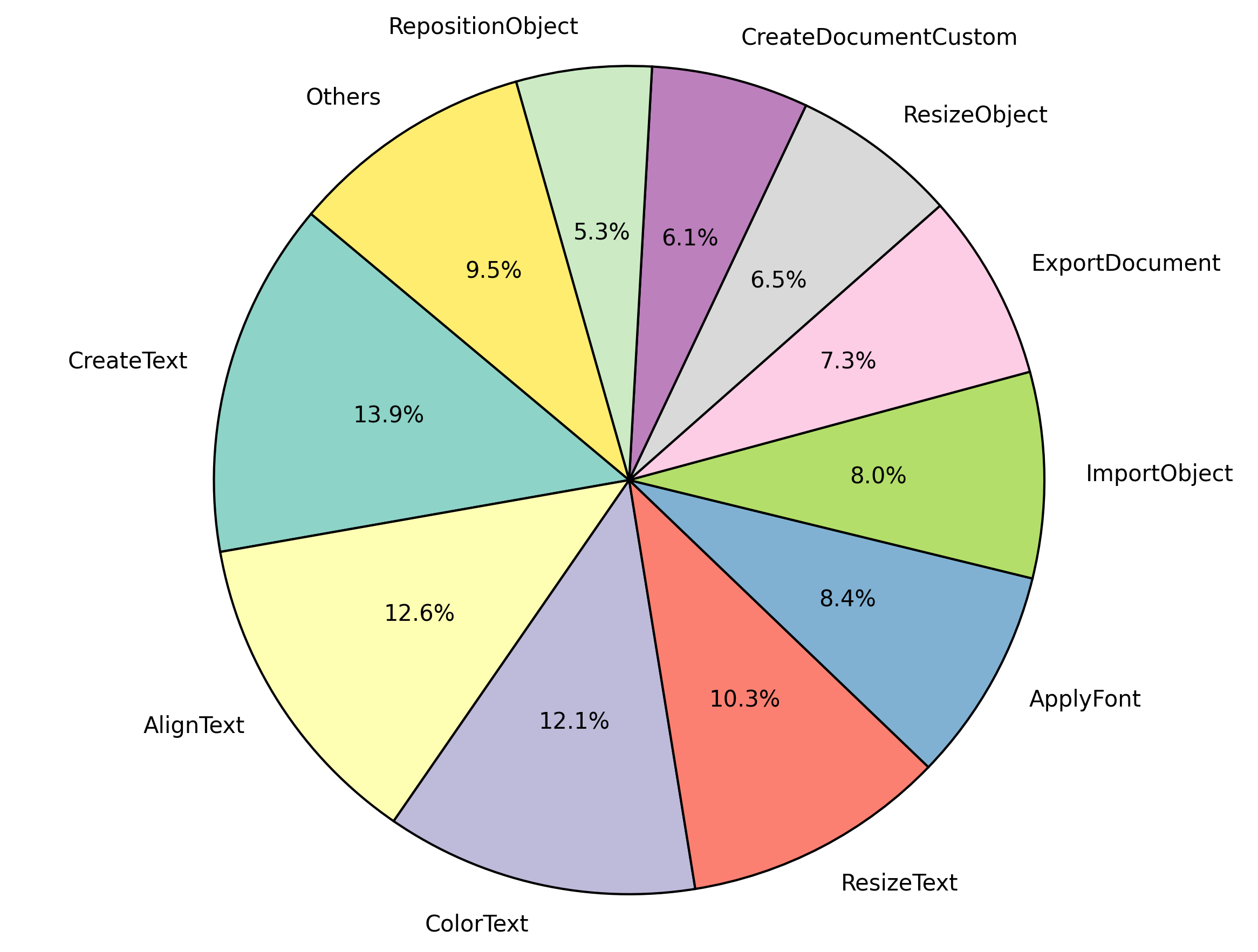}
        \caption{\textsc{LLaMA-3 8b} \includegraphics[height=1em]{figures/logo/indesign.png}}
    \end{subfigure}

    \begin{subfigure}{0.3\textwidth}
        \centering
        \includegraphics[height=3cm]{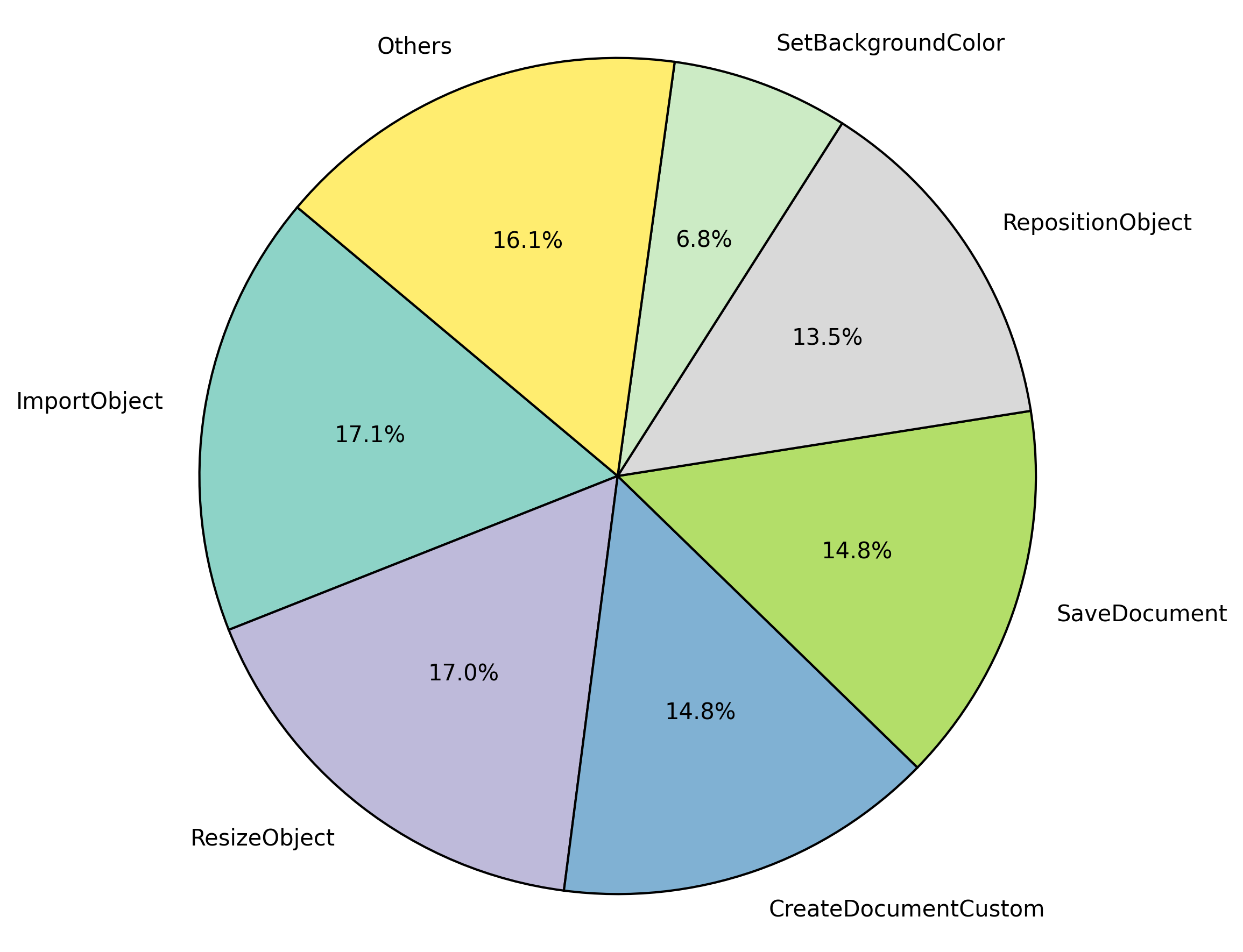}
        \caption{\textsc{Gemma-2 9b} \includegraphics[height=1em]{figures/logo/photoshop.png}}
    \end{subfigure}
    \hfill
    \begin{subfigure}{0.3\textwidth}
        \centering
        \includegraphics[height=3cm]{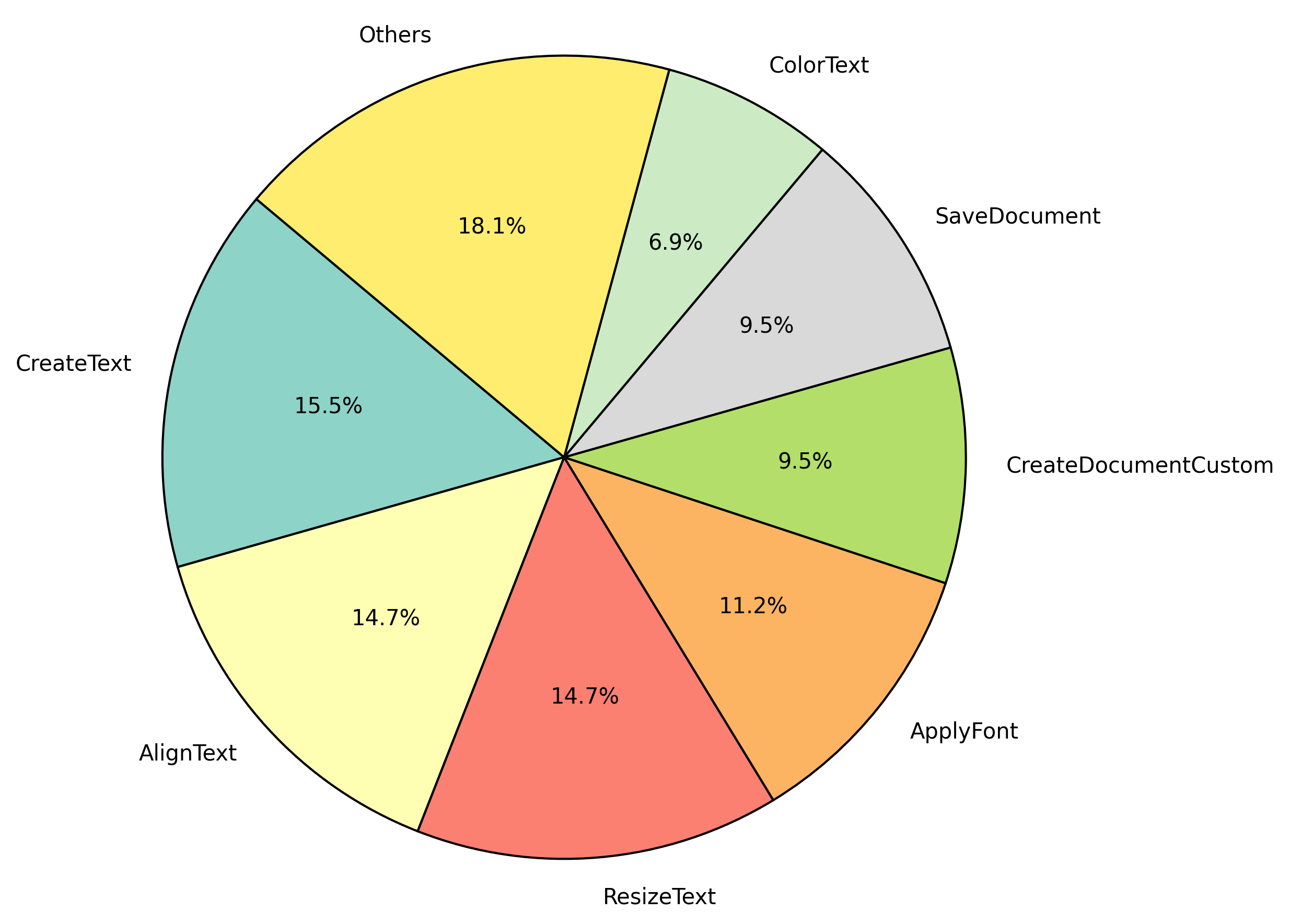}
        \caption{\textsc{Gemma-2 9b} \includegraphics[height=1em]{figures/logo/illustrator.png}}
    \end{subfigure}
    \hfill
    \begin{subfigure}{0.3\textwidth}
        \centering
        \includegraphics[height=3cm]{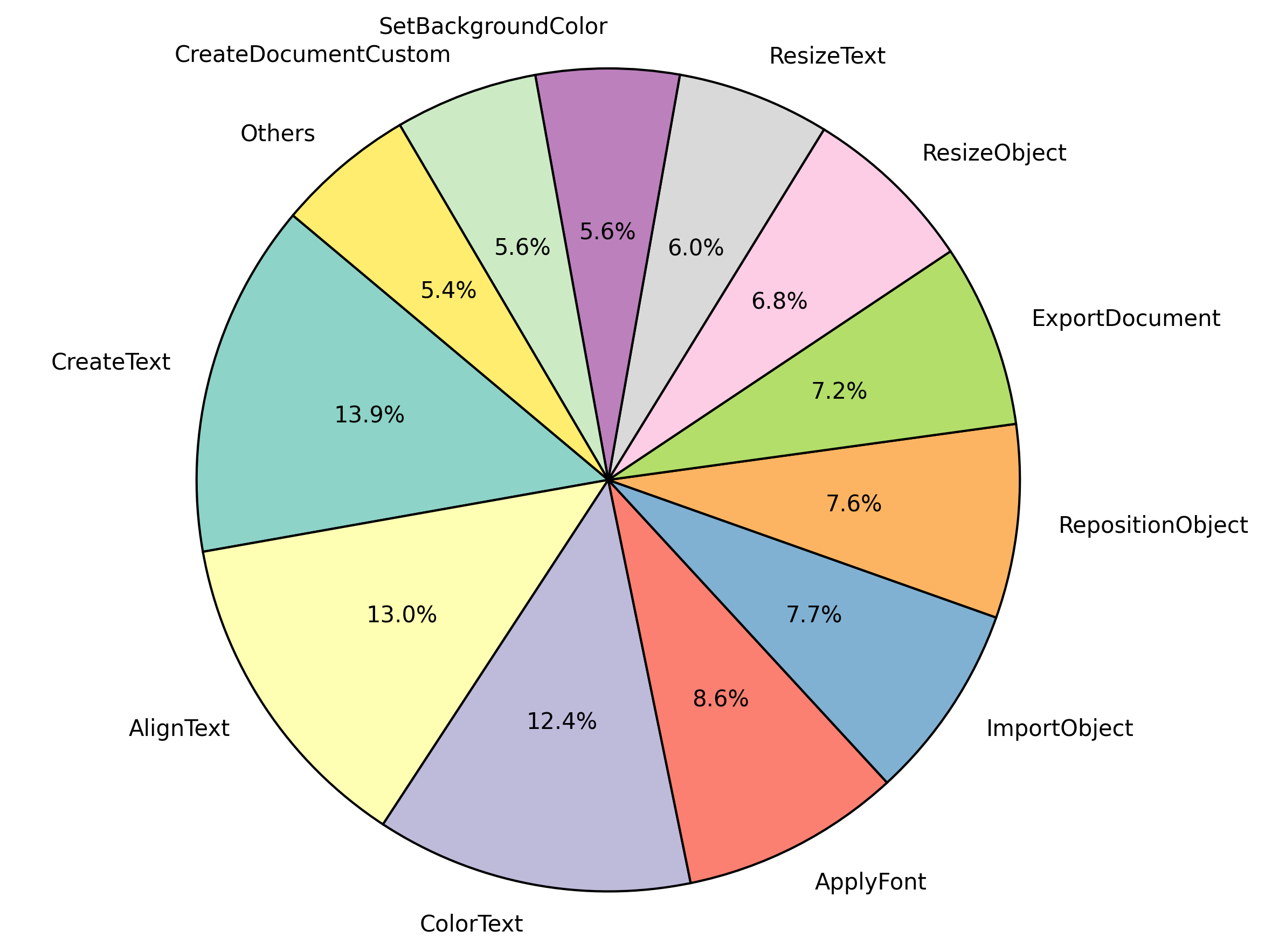}
        \caption{\textsc{Gemma-2 9b} \includegraphics[height=1em]{figures/logo/indesign.png}}
    \end{subfigure}

    \begin{subfigure}{0.3\textwidth}
        \centering
        \includegraphics[height=3cm]{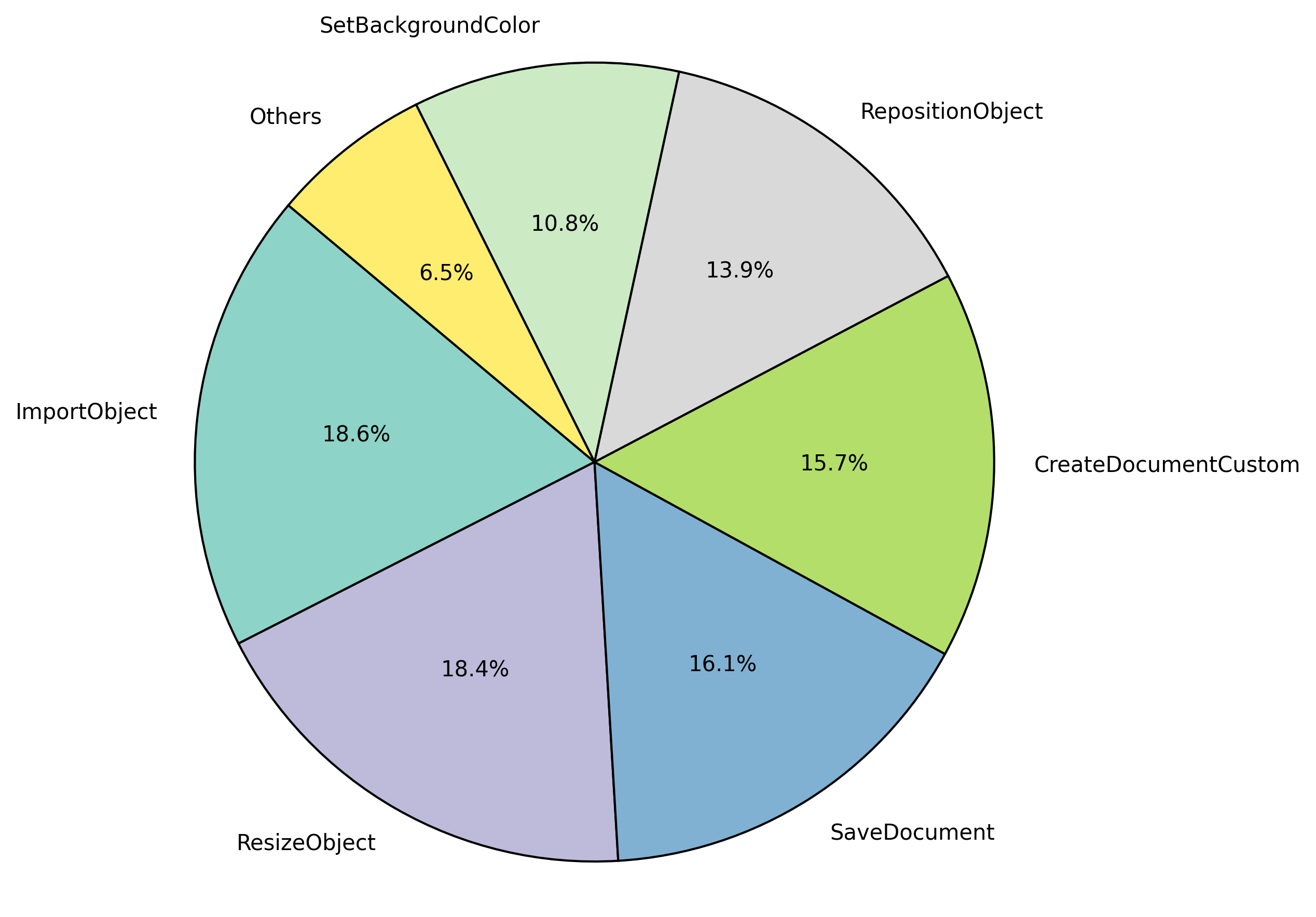}
        \caption{\textsc{Gemma-2 27b} \includegraphics[height=1em]{figures/logo/photoshop.png}}
    \end{subfigure}
    \hfill
    \begin{subfigure}{0.3\textwidth}
        \centering
        \includegraphics[height=3cm]{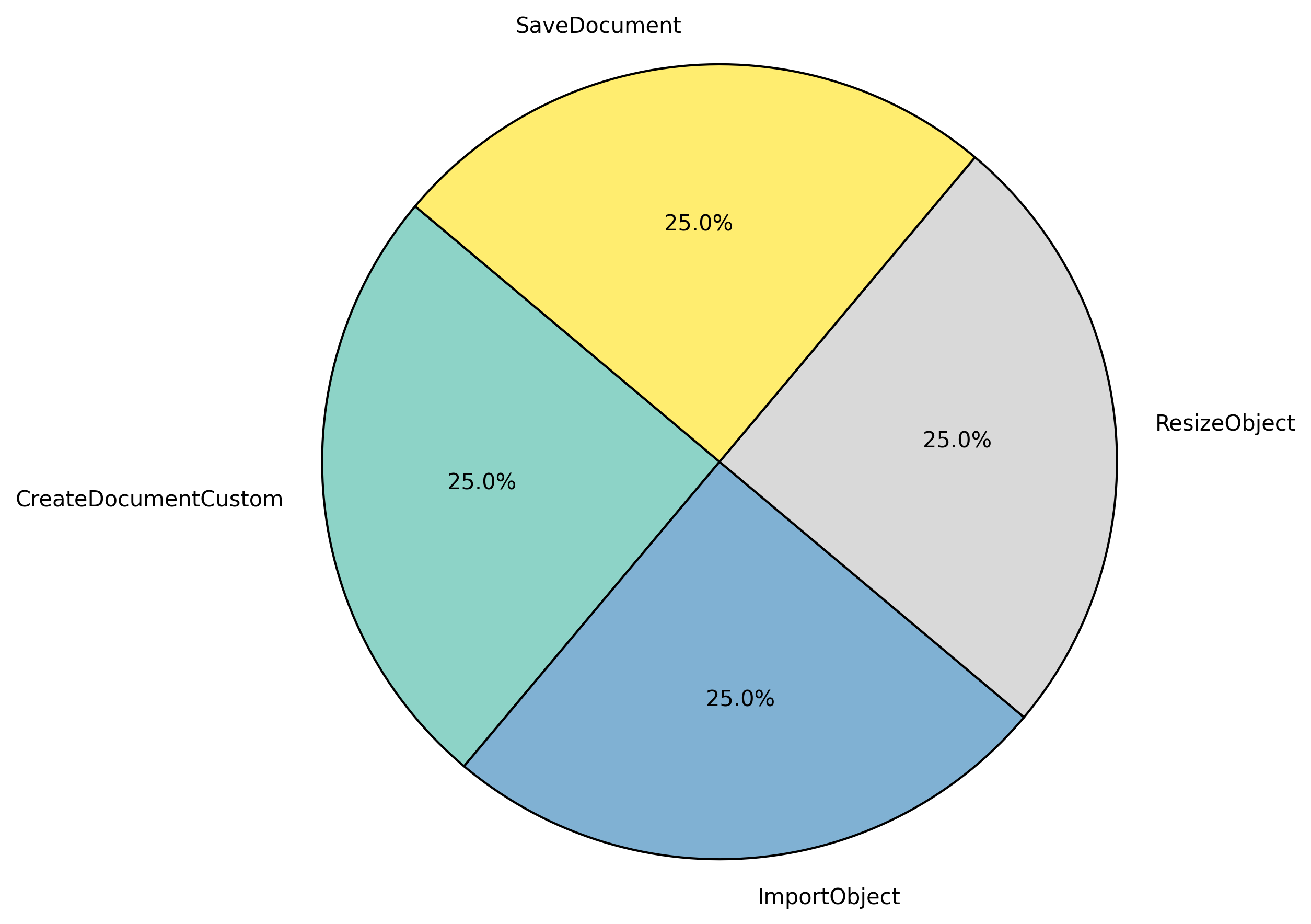}
        \caption{\textsc{Gemma-2 27b} \includegraphics[height=1em]{figures/logo/illustrator.png}}
    \end{subfigure}
    \hfill
    \begin{subfigure}{0.3\textwidth}
        \centering
        \includegraphics[height=3cm]{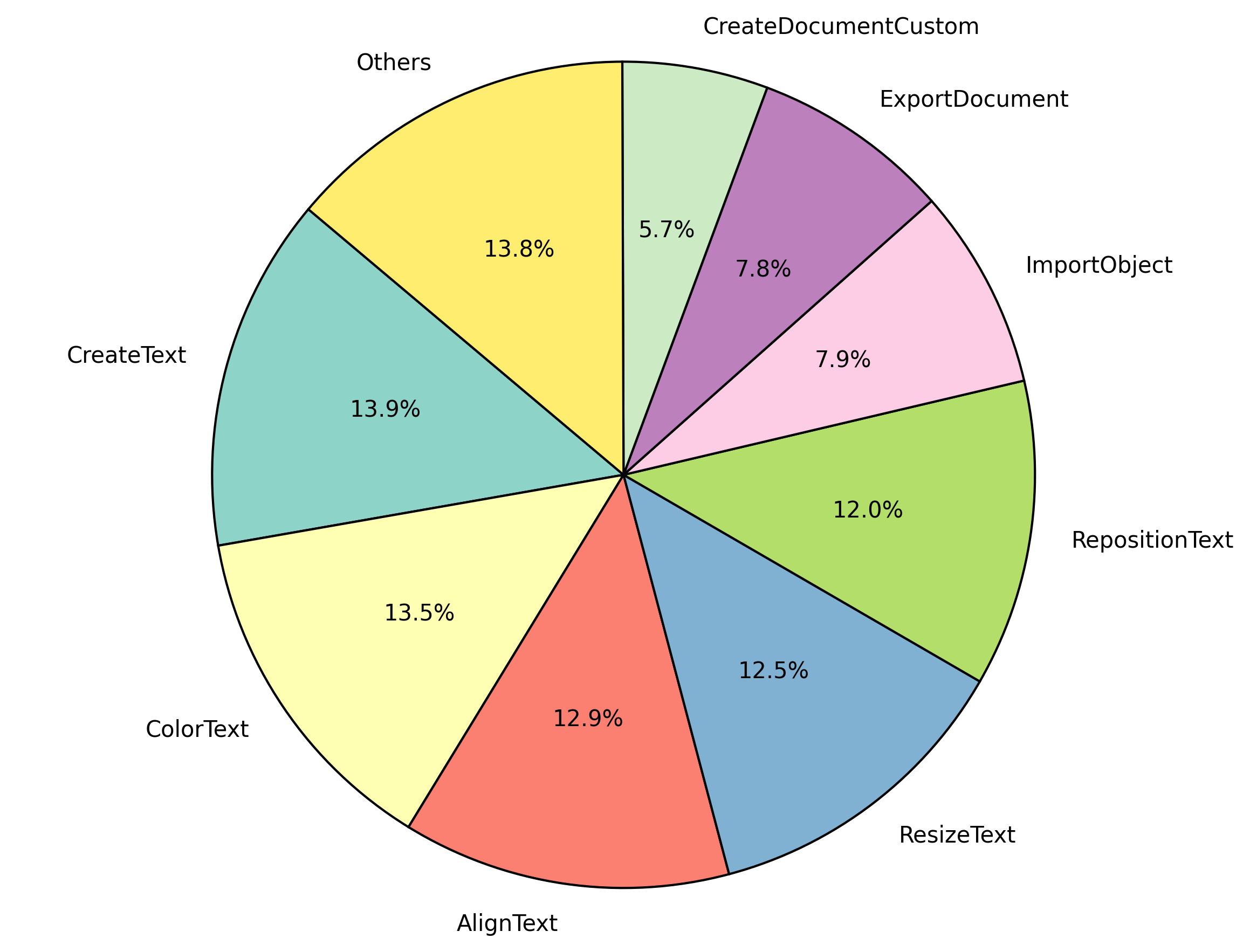}
        \caption{\textsc{Gemma-2 27b} \includegraphics[height=1em]{figures/logo/indesign.png}}
    \end{subfigure}

    \begin{subfigure}{0.3\textwidth}
        \centering
        \includegraphics[height=3cm]{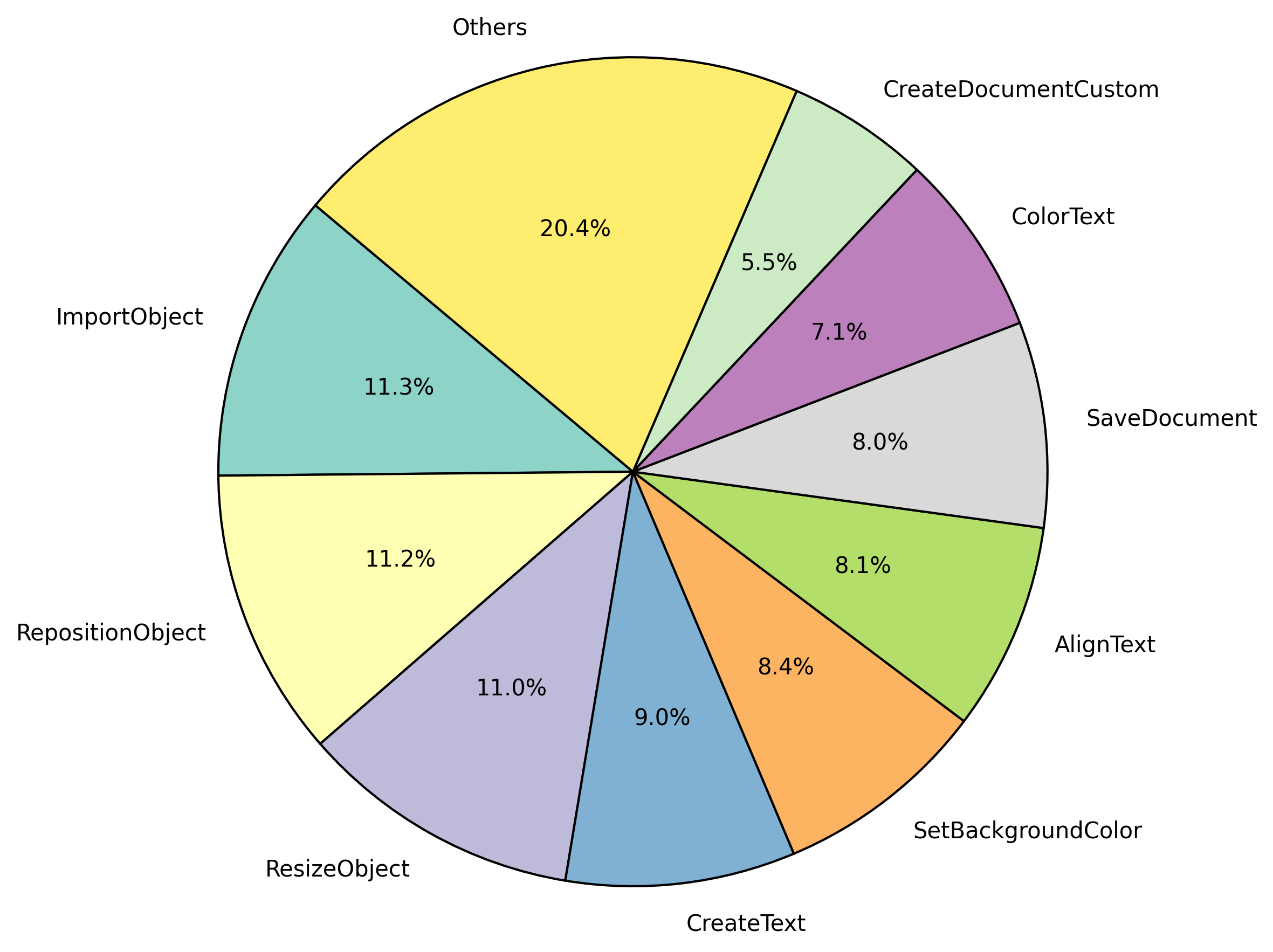}
        \caption{\textsc{Qwen-2.5 7b} \includegraphics[height=1em]{figures/logo/photoshop.png}}
    \end{subfigure}
    \hfill
    \begin{subfigure}{0.3\textwidth}
        \centering
        \includegraphics[height=3cm]{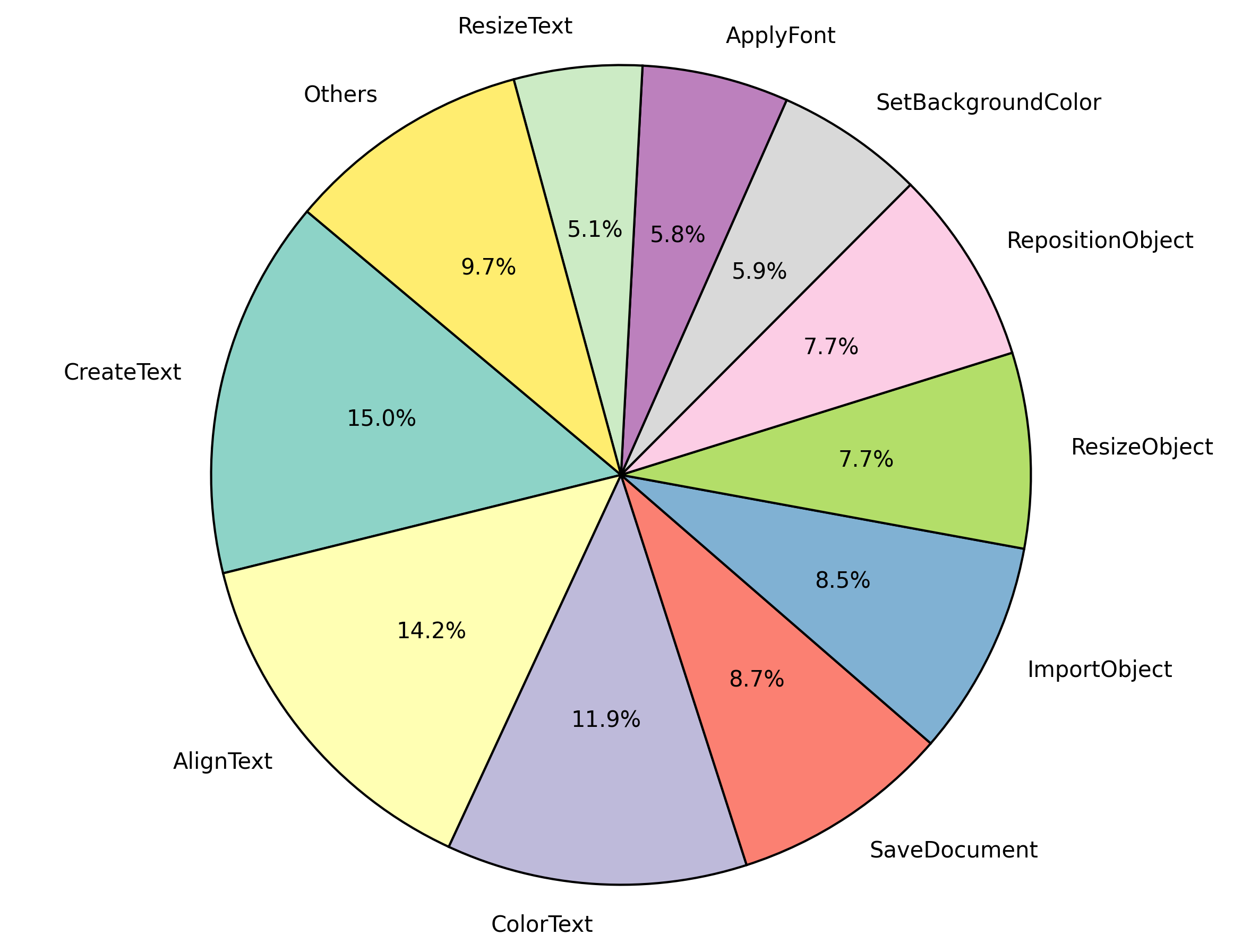}
        \caption{\textsc{Qwen-2.5 7b} \includegraphics[height=1em]{figures/logo/illustrator.png}}
    \end{subfigure}
    \hfill
    \begin{subfigure}{0.3\textwidth}
        \centering
        \includegraphics[height=3cm]{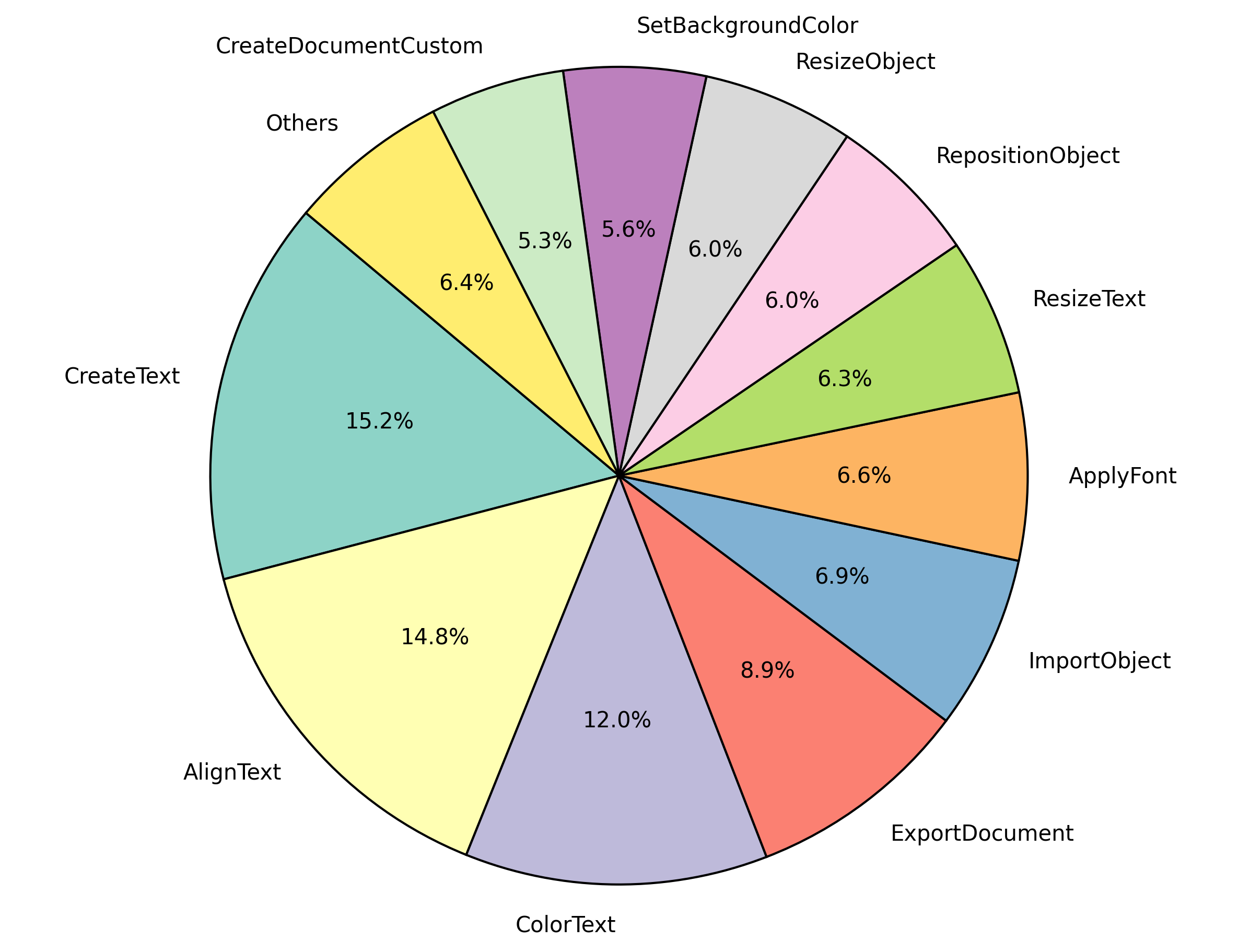}
        \caption{\textsc{Qwen-2.5 7b} \includegraphics[height=1em]{figures/logo/indesign.png}}
    \end{subfigure}

    \begin{subfigure}{0.3\textwidth}
        \centering
        \includegraphics[height=3cm]{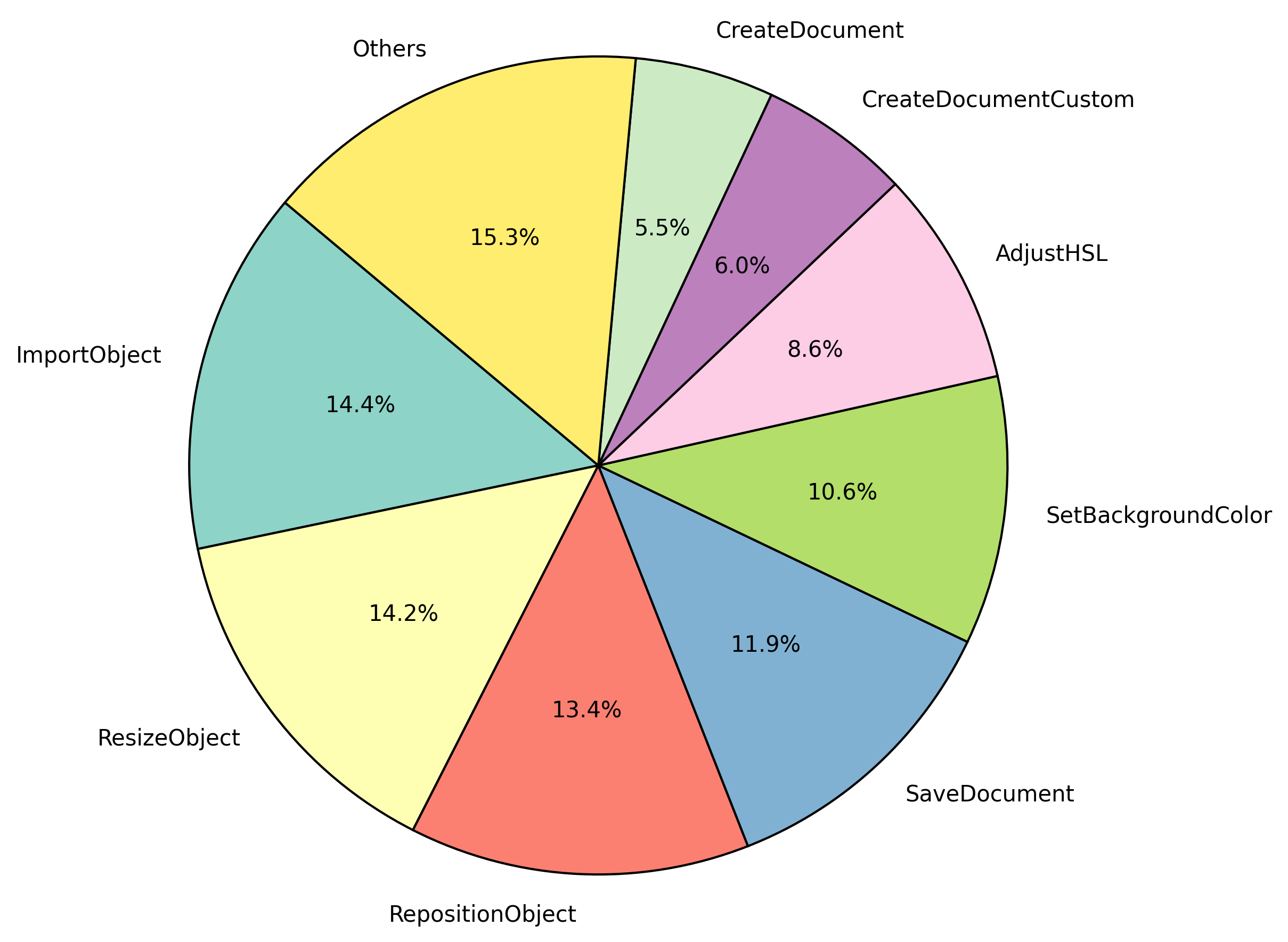}
        \caption{\textsc{Qwen-2.5 14b} \includegraphics[height=1em]{figures/logo/photoshop.png}}
    \end{subfigure}
    \hfill
    \begin{subfigure}{0.3\textwidth}
        \centering
        \includegraphics[height=3cm]{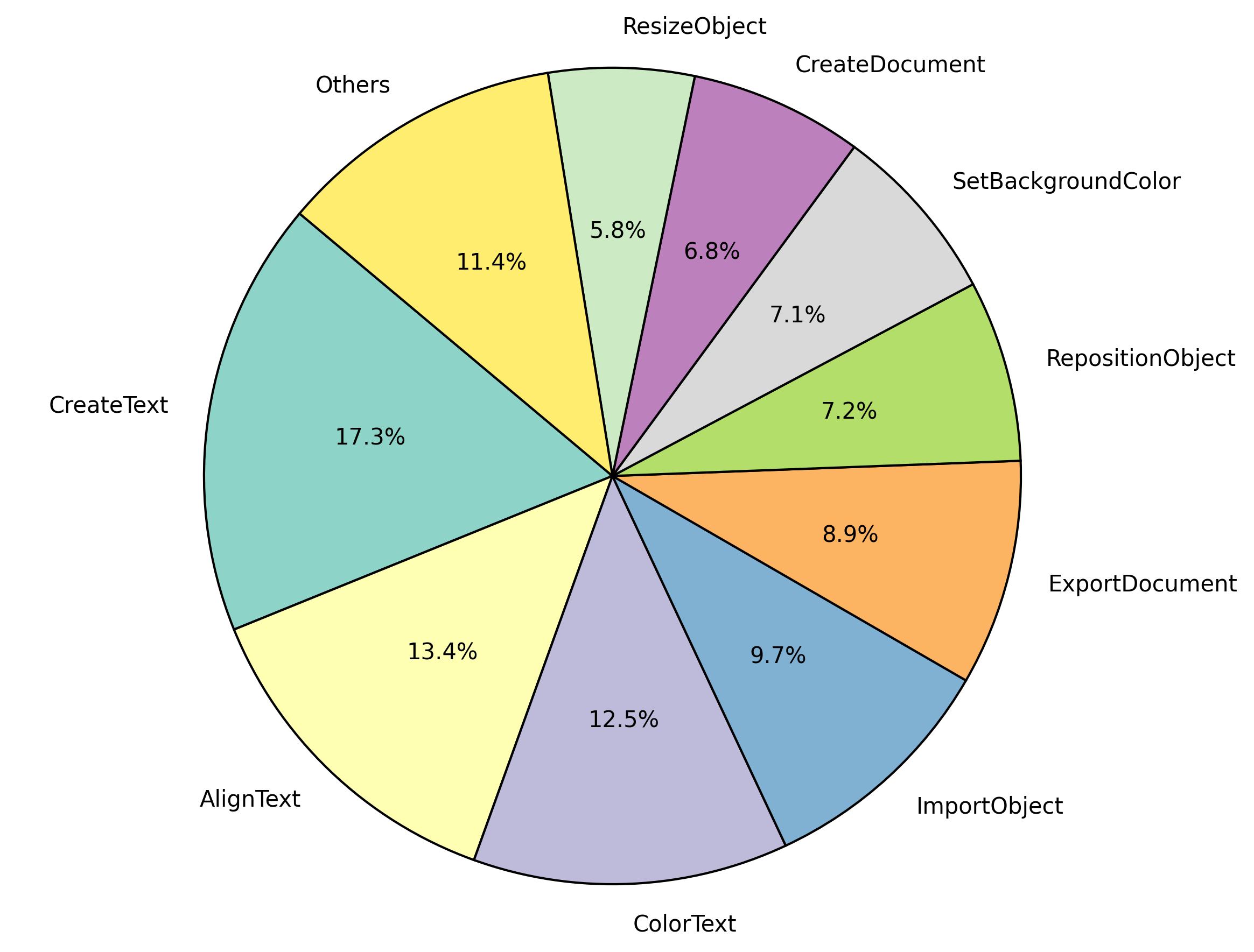}
        \caption{\textsc{Qwen-2.5 14b} \includegraphics[height=1em]{figures/logo/indesign.png}}
    \end{subfigure}

    \begin{subfigure}{0.3\textwidth}
        \centering
        \includegraphics[height=3cm]{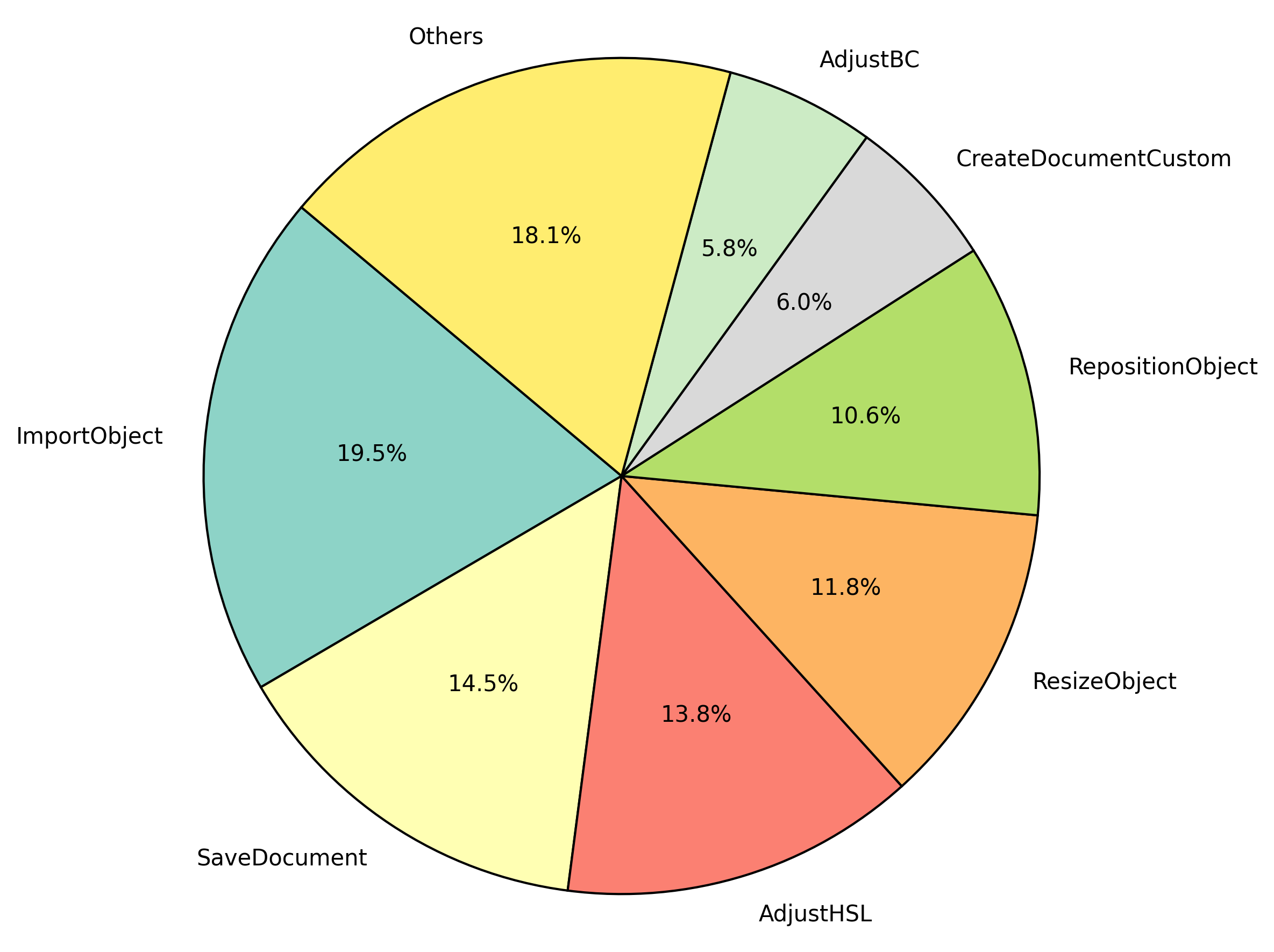}
        \caption{\textsc{GPT-3.5} \includegraphics[height=1em]{figures/logo/photoshop.png}}
    \end{subfigure}
    \hfill
    \begin{subfigure}{0.3\textwidth}
        \centering
        \includegraphics[height=3cm]{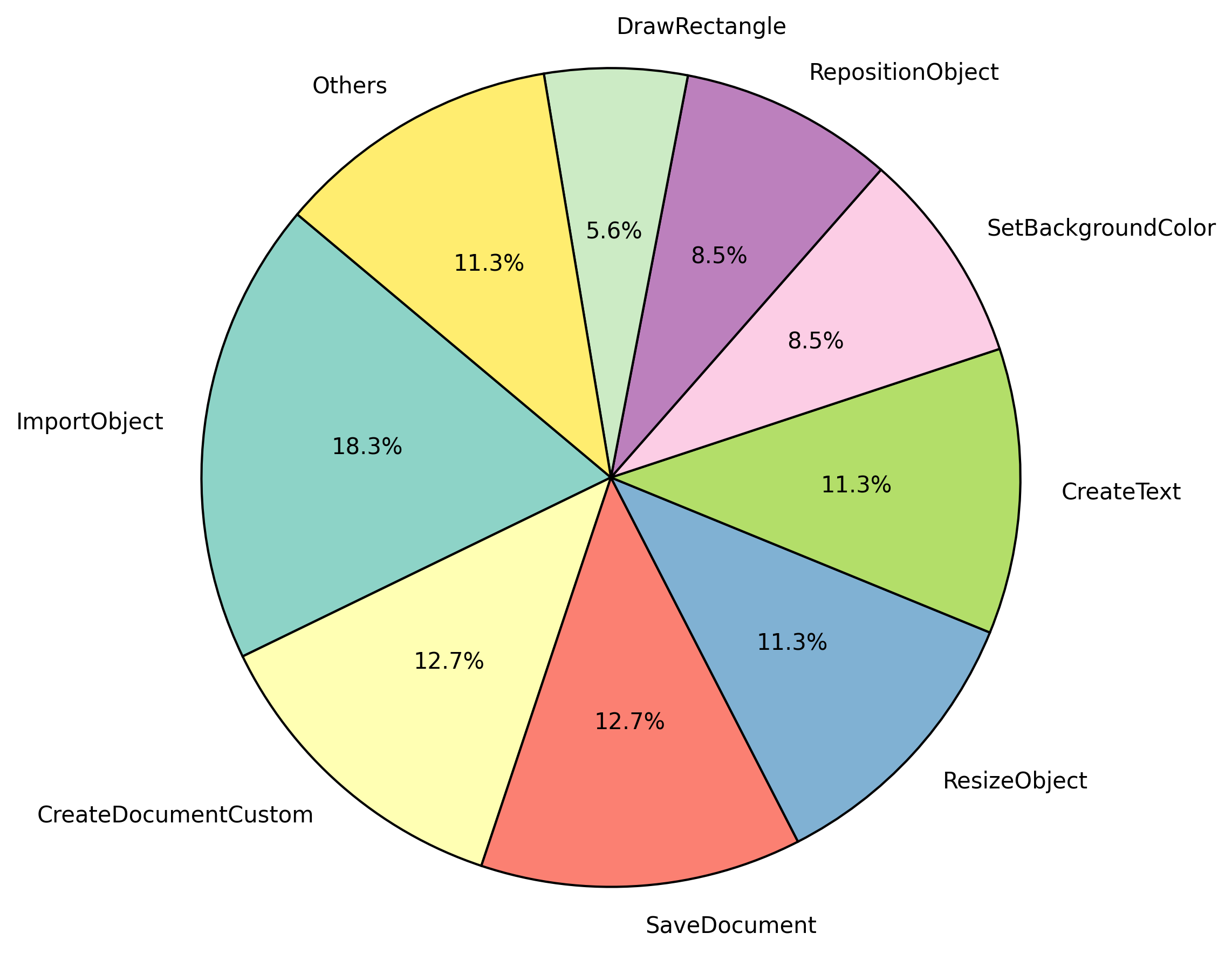}
        \caption{\textsc{GPT-3.5} \includegraphics[height=1em]{figures/logo/illustrator.png}}
    \end{subfigure}
    \hfill
    \begin{subfigure}{0.3\textwidth}
        \centering
        \includegraphics[height=3cm]{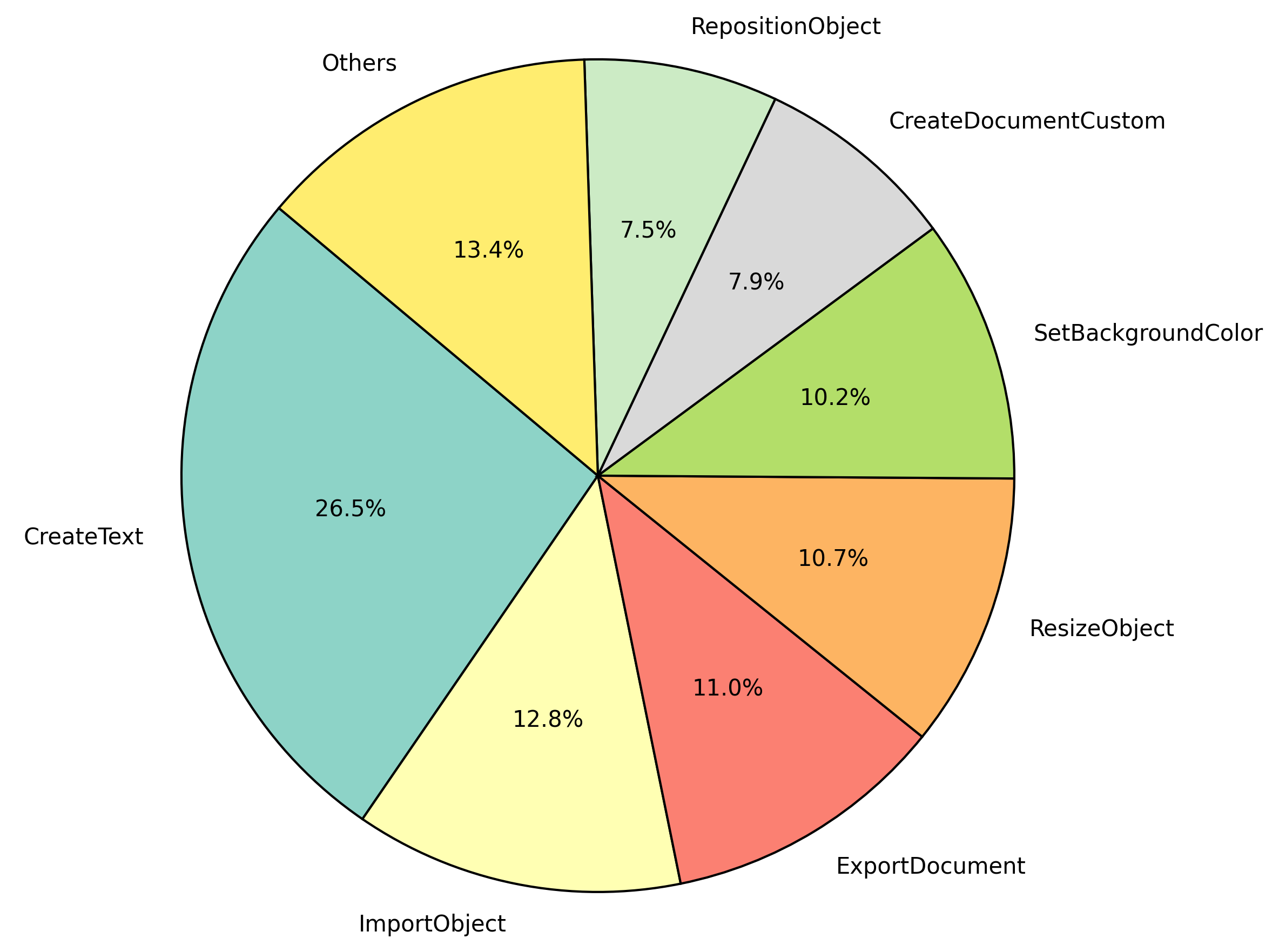}
        \caption{\textsc{GPT-3.5} \includegraphics[height=1em]{figures/logo/indesign.png}}
    \end{subfigure}

    \caption{Action distribution per model and expert agent. Actions under 5.0\% of usage are grouped as ``Others''. \includegraphics[height=1em]{figures/logo/photoshop.png}: Photo Editor; \includegraphics[height=1em]{figures/logo/illustrator.png}: Vector Graphic Editor; \includegraphics[height=1em]{figures/logo/indesign.png}: Layout Designer.}
    \label{fig:detailed_res3}
\end{figure*}

We detail results for action distribution across models and expert agents, aggregated over all design types. As shown in Figure \ref{fig:detailed_res3}, each expert agent exhibits clear preferences for specific actions. The Photo Editor agent primarily utilizes object manipulation functions such as \texttt{ImportObject}, \texttt{ResizeObject}, and \texttt{RepositionObject}. Notably, \textsc{GPT-3.5} frequently applies color correction functions (\texttt{AdjustHSL}, \texttt{AdjustBC}), reflecting the agent's expertise. The common actions used by the Vector Graphic Editor agent vary by model. \textsc{LLaMA-3 8b}, \textsc{Gemma-2 27b} and \textsc{GPT-3.5} predominately use object manipulation functions, whereas \textsc{Gemma-2 9b} and \textsc{Qwen-2.5 7b} uses text-related functions. Meanwhile, the Layout Designer agent primarily uses text-related operations such as \texttt{CreateText}, \texttt{AlignText}, and \texttt{ColorText}. Overall, expert agents struggle to utilize advanced text and object manipulation functions, such as \texttt{OpacityText} and \texttt{PhotoFilter}, regardless of the model.

We present the top-3 most common action sequences per model in Table \ref{tab:common_sequence}. We observe that most sequences closely follow human-annotated workflows, typically starting with document creation (\texttt{CreateDocument} or \texttt{CreateDocumentCustom}), setting the background color (\texttt{SetBackgroundColor}), importing images as objects (\texttt{ImportObject}), manipulating the imported object (\texttt{ResizeObject}, \texttt{RepositionObject}, etc.), saving the document (\texttt{SaveDocument}), and manipulating text elements (\texttt{CreateText}, \texttt{ApplyFont}, \texttt{ColorText}, etc.).

\begin{table*}
\centering
\resizebox{\linewidth}{!}{%
    \begin{tabular}{l l p{1.2\textwidth}}
    \toprule
    \textbf{Model} & \textbf{Occurrence} & \textbf{Action Sequence} \\
    \toprule

    \textbf{\textsc{LLaMA-3 8b}} & 13 & \texttt{CreateDocumentCustom} → \texttt{SetBackgroundColor} → \texttt{ImportObject} → \texttt{ResizeObject} → \texttt{RepositionObject} → \texttt{CreateText} →  \texttt{ApplyFont} → \texttt{ColorText} →  \texttt{AlignText} →  \texttt{ResizeText} →  \texttt{RepositionText} → \texttt{ExportDocument} → \texttt{AdjustHSL} → \texttt{SaveDocument} \\
    \cmidrule{2-3}
    & 12 & \texttt{CreateDocumentCustom} → \texttt{SetBackgroundColor} → \texttt{ImportObject} → \texttt{ResizeObject} → \texttt{RepositionObject} → \texttt{AdjustHSL} → \texttt{SaveDocument} → \texttt{CreateText} →  \texttt{ApplyFont} → \texttt{ColorText} →  \texttt{AlignText} →  \texttt{ResizeText} → \texttt{ExportDocument} \\
    \cmidrule{2-3}
    & 11 & \texttt{CreateDocumentCustom} → \texttt{SetBackgroundColor} → \texttt{ImportObject} → \texttt{ResizeObject} → \texttt{RepositionObject} → \texttt{CreateText} →  \texttt{ApplyFont} → \texttt{ColorText} →  \texttt{AlignText} →  \texttt{ResizeText} → \texttt{ExportDocument} → \texttt{AdjustHSL} → \texttt{SaveDocument} \\
    \midrule

    \textbf{\textsc{Gemma-2 9b}} & 48 & \texttt{CreateDocumentCustom} → \texttt{SetBackgroundColor} → \texttt{ImportObject} → \texttt{ResizeObject} → \texttt{RepositionObject} → \texttt{SaveDocument} → \texttt{CreateText} →  \texttt{ApplyFont} → \texttt{ColorText} →  \texttt{AlignText} → \texttt{ExportDocument} \\
    \cmidrule{2-3}
    & 26 & \texttt{CreateDocumentCustom} → \texttt{SetBackgroundColor} → \texttt{ImportObject} → \texttt{ResizeObject} → \texttt{RepositionObject} → \texttt{SaveDocument} → \texttt{CreateText} →  \texttt{ApplyFont} → \texttt{ColorText} →  \texttt{AlignText} →  \texttt{RepositionText} → \texttt{ExportDocument} \\
    \cmidrule{2-3}
    & 26 & \texttt{CreateDocumentCustom} → \texttt{ImportObject} → \texttt{ResizeObject} → \texttt{RepositionObject} → \texttt{SaveDocument} → \texttt{SetBackgroundColor} → \texttt{CreateText} →  \texttt{ApplyFont} → \texttt{ColorText} →  \texttt{AlignText} → \texttt{ExportDocument} \\
    \midrule

    \textbf{\textsc{Gemma-2 27b}} & 113 & \texttt{CreateDocumentCustom} → \texttt{SetBackgroundColor} → \texttt{ImportObject} → \texttt{ResizeObject} → \texttt{RepositionObject} → \texttt{SaveDocument} → \texttt{CreateText} →  \texttt{ResizeText} → \texttt{ColorText} →  \texttt{AlignText} →  \texttt{RepositionText} → \texttt{ExportDocument} \\
    \cmidrule{2-3}
    & 32 & \texttt{CreateDocumentCustom} → \texttt{SetBackgroundColor} → \texttt{ImportObject} → \texttt{ResizeObject} → \texttt{RepositionObject} → \texttt{SaveDocument} → \texttt{CreateText} →  \texttt{ResizeText} → \texttt{ColorText} →  \texttt{AlignText} → \texttt{ExportDocument} \\
    \cmidrule{2-3}
    & 31 & \texttt{CreateDocumentCustom} → \texttt{SetBackgroundColor} → \texttt{ImportObject} → \texttt{ResizeObject} → \texttt{RepositionObject} → \texttt{AdjustHSL} → \texttt{SaveDocument} → \texttt{CreateText} →  \texttt{ResizeText} → \texttt{ColorText} →  \texttt{AlignText} →  \texttt{RepositionText} → \texttt{ExportDocument} \\
    \midrule

    \textbf{\textsc{Qwen-2.5 7b}} & 30 & \texttt{CreateDocumentCustom} → \texttt{SetBackgroundColor} → \texttt{ImportObject} → \texttt{ResizeObject} → \texttt{RepositionObject} → \texttt{CreateText} → \texttt{ColorText} →  \texttt{AlignText} → \texttt{ExportDocument} \\
    \cmidrule{2-3}
    & 21 & \texttt{CreateDocumentCustom} → \texttt{SetBackgroundColor} → \texttt{ImportObject} → \texttt{ResizeObject} → \texttt{RepositionObject} → \texttt{CreateText} →  \texttt{ResizeText} → \texttt{ColorText} →  \texttt{AlignText} → \texttt{ExportDocument} \\
    \cmidrule{2-3}
    & 13 & \texttt{CreateDocumentCustom} → \texttt{SetBackgroundColor} → \texttt{CreateText} → \texttt{ColorText} →  \texttt{AlignText} → \texttt{ImportObject} → \texttt{ResizeObject} → \texttt{RepositionObject} → \texttt{ExportDocument} \\
    \midrule

    \textbf{\textsc{Qwen-2.5 14b}} & 54 & \texttt{CreateDocument} → \texttt{SetBackgroundColor} → \texttt{ImportObject} → \texttt{ResizeObject} → \texttt{RepositionObject} → \texttt{AdjustHSL} → \texttt{SaveDocument} → \texttt{CreateText} → \texttt{ColorText} →  \texttt{AlignText} → \texttt{ExportDocument} \\
    \cmidrule{2-3}
    & 46 & \texttt{CreateDocumentCustom} → \texttt{SetBackgroundColor} → \texttt{ImportObject} → \texttt{ResizeObject} → \texttt{RepositionObject} → \texttt{AdjustHSL} → \texttt{SaveDocument} → \texttt{CreateDocument} → \texttt{CreateText} → \texttt{ColorText} →  \texttt{AlignText} → \texttt{ExportDocument} \\
    \cmidrule{2-3}
    & 25 & \texttt{CreateDocument} → \texttt{SetBackgroundColor} → \texttt{ImportObject} → \texttt{ResizeObject} → \texttt{RepositionObject} → \texttt{AdjustHSL} → \texttt{SaveDocument} → \texttt{CreateText} →  \texttt{AlignText} → \texttt{ExportDocument} \\
    \midrule

    \textbf{\textsc{GPT-3.5}} & 30 & \texttt{CreateDocumentCustom} → \texttt{SetBackgroundColor} → \texttt{ImportObject} → \texttt{ResizeObject} → \texttt{RepositionObject} → \texttt{CreateText} → \texttt{ColorText} →  \texttt{AlignText} → \texttt{ExportDocument} \\
    \cmidrule{2-3}
    & 21 & \texttt{CreateDocumentCustom} → \texttt{SetBackgroundColor} → \texttt{ImportObject} → \texttt{ResizeObject} → \texttt{RepositionObject} → \texttt{CreateText} →  \texttt{ResizeText} → \texttt{ColorText} →  \texttt{AlignText} → \texttt{ExportDocument} \\
    \cmidrule{2-3}
    & 13 & \texttt{CreateDocumentCustom} → \texttt{SetBackgroundColor} → \texttt{CreateText} → \texttt{ColorText} →  \texttt{AlignText} → \texttt{ImportObject} → \texttt{ResizeObject} → \texttt{RepositionObject} → \texttt{ExportDocument} \\
    
    \bottomrule
    \end{tabular}
}
\caption{Top-3 most common action sequences per model, ordered by frequency. \textbf{Occurrence:} Number of times each specific sequence occurs.}
\label{tab:common_sequence}
\end{table*}

\subsection{Execution Evaluation}
\label{appendix:detailed_res4}
We present the detailed numerical results for execution evaluation by model and design type in Table \ref{tab:detailed_res4}. 

% We further observe a weak positive correlation between content similarity and VQA pass rates ($r$=0.582).

\begin{table*}
\centering
\resizebox{\linewidth}{!}{%
    \begin{tabular}{lllllllllll}
    \toprule
    \textbf{Model} & \textbf{Design Type} & \textbf{Success Rate (\%)} & \textbf{Fidelity} & \textbf{Content Similarity} & \textbf{VQA Pass Rate} & \textbf{Creativity (O)} & \textbf{Creativity (E)} \\
    \toprule

    \multirow{4}{*}{\textbf{\textsc{LLaMA-3 8b}}} & Book Cover & 90.42 & 0.157 & 15.93 & 34.72 & 1.79 & 1.52 \\
    & Business Card & 86.41 & 0.133 & 16.71 & 36.35 & 1.64 & 1.49 \\
    & Postcard & 83.86 & 0.160 & 20.26 & 42.16 & 2.02 & 1.85 \\
    & Poster & 85.12 & 0.161 & 20.20 & 39.27 & 1.79 & 1.33 \\
    \midrule
    
    \multirow{4}{*}{\textbf{\textsc{Gemma-2 9b}}} & Book Cover & 79.66 & 0.166 & 15.16 & 35.94 & 2.03 & 1.68 \\
    & Business Card & 81.32 & 0.182 & 16.12 & 31.42 & 1.77 & 1.59 \\
    & Postcard & 88.59 & 0.193 & 26.99 & 49.42 & 2.10 & 1.70 \\
    & Poster & 80.17 & 0.171 & 19.18 & 32.55 & 1.81 & 1.43 \\
    \midrule
    
    \multirow{4}{*}{\textbf{\textsc{Gemma-2 27b}}} & Book Cover & 15.81 & 0.093 & 14.94 & 39.85 & \textbf{2.27} & 1.93 \\
    & Business Card & 83.38 & 0.107 & 16.91 & 32.34 & 1.85 & 1.59 \\
    & Postcard & 72.76 & 0.167 & 26.90 & \textbf{51.26} & 2.17 & 1.68 \\
    & Poster & 62.85 & 0.168 & 19.58 & 35.04 & 1.92 & 1.33 \\
    \midrule
    
    \multirow{4}{*}{\textbf{\textsc{Qwen-2.5 7b}}} & Book Cover & 64.22 & 0.152 & 15.89 & 35.99 & 1.55 & 1.34 \\
    & Business Card & 38.41 & 0.152 & 18.16 & 47.00 & 1.74 & 1.46 \\
    & Postcard & 19.20 & 0.130 & \textbf{27.39} & 45.59 & 2.04 & 1.90 \\
    & Poster & 58.35 & 0.173 & 20.16 & 36.50 & 1.64 & 1.21 \\
    \midrule
    
    \multirow{4}{*}{\textbf{\textsc{Qwen-2.5 14b}}} & Book Cover & \textbf{97.32} & 0.173 & 16.09 & 33.64 & 1.65 & 1.41 \\
    & Business Card & 83.39 & 0.130 & 19.17 & 27.17 & 1.55 & 1.51 \\
    & Postcard & 80.87 & \textbf{0.194} & 26.13 & 44.49 & 2.10 & 1.72 \\
    & Poster & 97.00 & 0.169 & 20.00 & 28.50 & 1.62 & 1.40 \\
    \midrule
    
    \multirow{4}{*}{\textbf{\textsc{GPT-3.5}}} & Book Cover & 79.61 & 0.165 & 16.91 & 22.56 & 1.87 & 1.83 \\
    & Business Card & 72.70 & 0.174 & 19.06 & 27.65 & 1.74 & 2.02 \\
    & Postcard & 75.03 & 0.174 & 26.05 & 39.70 & 2.09 & 2.05 \\
    & Poster & 51.49 & 0.160 & 20.34 & 25.31 & 2.15 & \textbf{2.08} \\

    \bottomrule
    \end{tabular}
}
\caption{Execution results per model and design type. Best scores for each column is \textbf{bold}. \textbf{Creativity (O):} Originality; \textbf{Creativity (E):} Elaboration}
\label{tab:detailed_res4}
\end{table*}

\subsection{Case Studies}
\label{appendix:case_studies}
In Table \ref{tab:case_study}, we present cases studies of failed executions across design types. We conclude with similar observations from the error analysis (\S \ref{sec:error_analysis}):

\paragraph{LLM agents lack spatial understanding of design elements.}
A qualitative analysis of execution outcomes reveals that most failures stem from agents' lack of spatial understanding and object positioning within a given space. For instance, in the second book cover example by \textsc{LLaMA-3 8b}, the postcard example by \textsc{Gemma-2 9b}, and the first poster example by \textsc{Qwen-2.5 7b}, agents struggle to correctly position text. Additionally, we observe some cases where they fail to correctly position both text and images, such as the second poster example by \textsc{GPT-3.5}.

\paragraph{LLM agents struggle to understand global dependencies.}
As shown in Table \ref{tab:case_study}, many failed executions result from failing to recognize global dependencies between expert agents. Consequently, they generate incomplete outcomes, executing only a subset of actions \---\ either text-related (e.g., the first book cover example by \textsc{LLaMA-3 8b}) or image-related operations (e.g., the first business card example by \textsc{Gemma-2 27b}).

\begin{table*}
\centering
\resizebox{\linewidth}{!}{%
    \begin{tabular}{l p{0.7\textwidth} p{0.25\textwidth} p{0.3\textwidth}} % Adjust column widths
    \toprule
    \textbf{Design Type} & \textbf{User Query} & \textbf{Execution Outcome} & \textbf{Comments (Model)} \\
    \toprule
    \textbf{Book Cover} &
    \begin{minipage}{0.7\textwidth}
        Can you assist in creating a historical novel cover design titled `Secrets of the Pharaohs' with the author's name `Liam Hunter', featuring an illustration of the great pyramids centered in the middle under a golden sunset sky? I want the title in ancient gold at the top center above the author's name, and the author's name in ancient gold at the top center below the title. Also, include the tagline `Unveil the Mysteries of Ancient Egypt.' in ancient gold below the author's name.
    \end{minipage} & 
    \begin{minipage}{0.25\textwidth}
        \centering
        \makebox[\textwidth]{ % Ensures both images fit inside the column
            \includegraphics[width=\textwidth]{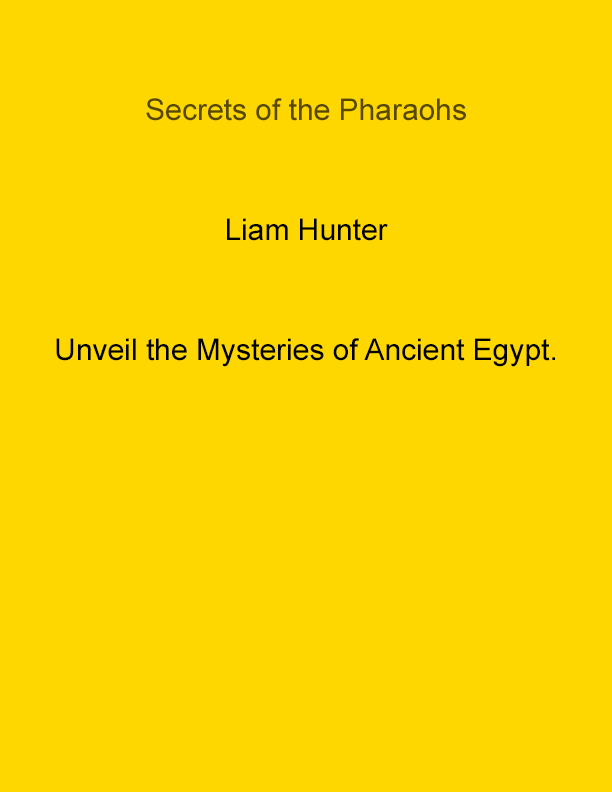}
        }
    \end{minipage} &
    \begin{minipage}{0.3\textwidth}
        Execute text-related manipulations with one expert agent but fail to transfer image-related tasks to the next agent. (\textsc{LLaMA-3 8b})
    \end{minipage} \\
    \midrule

    \textbf{Book Cover} &
    \begin{minipage}{0.7\textwidth}
        Please help me create a travel guide cover design titled `Exploring Tuscany' with a blue sky background, featuring rolling vineyards in the background below a large Tuscan villa centered on the lower half. The title `Exploring Tuscany' should be at the top center in dark green, with 'Travel Guide' below the title and `Discover the Heart of Italy' below the author, both in dark green.
    \end{minipage} & 
    \begin{minipage}{0.25\textwidth}
        \centering
        \makebox[\textwidth]{ % Ensures both images fit inside the column
            \includegraphics[width=\textwidth]{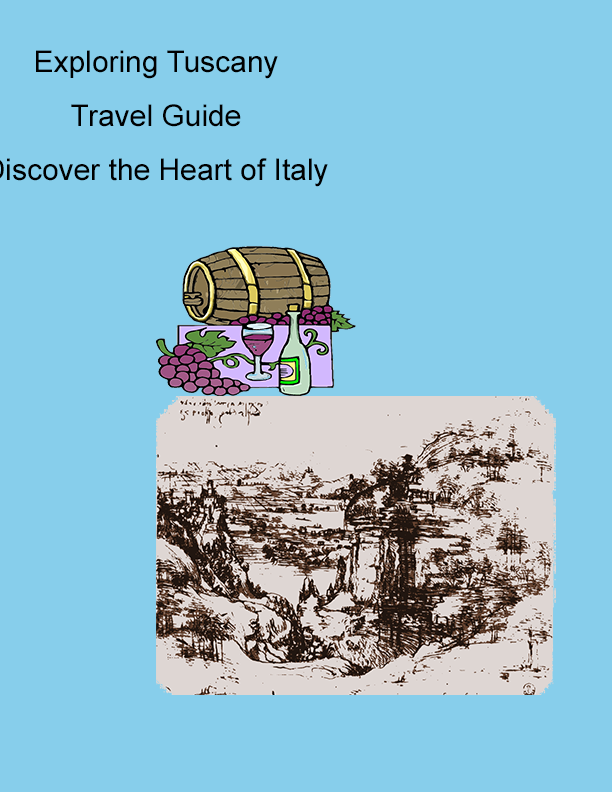}
        }
    \end{minipage} &
    \begin{minipage}{0.3\textwidth}
        Successfully incorporate all design details but fail to position the title and subtitles in the center. (\textsc{LLaMA-3 8b})
    \end{minipage} \\
    \midrule

    \textbf{Book Cover} &
    \begin{minipage}{0.7\textwidth}
        Can you create a book cover design titled `Robot Revolution' with the author's name `Alex Turner', featuring a large futuristic cityscape with medium-sized robots in action, set against a metallic grey background? The title `Robot Revolution' should be in neon blue at the top center, with the author's name `Alex Turner' in neon blue below the title, and the tagline `The Future is Now.' in neon blue below the author's name.
    \end{minipage} & 
    \begin{minipage}{0.25\textwidth}
        \centering
        \makebox[\textwidth]{ % Ensures both images fit inside the column
            \includegraphics[width=\textwidth]{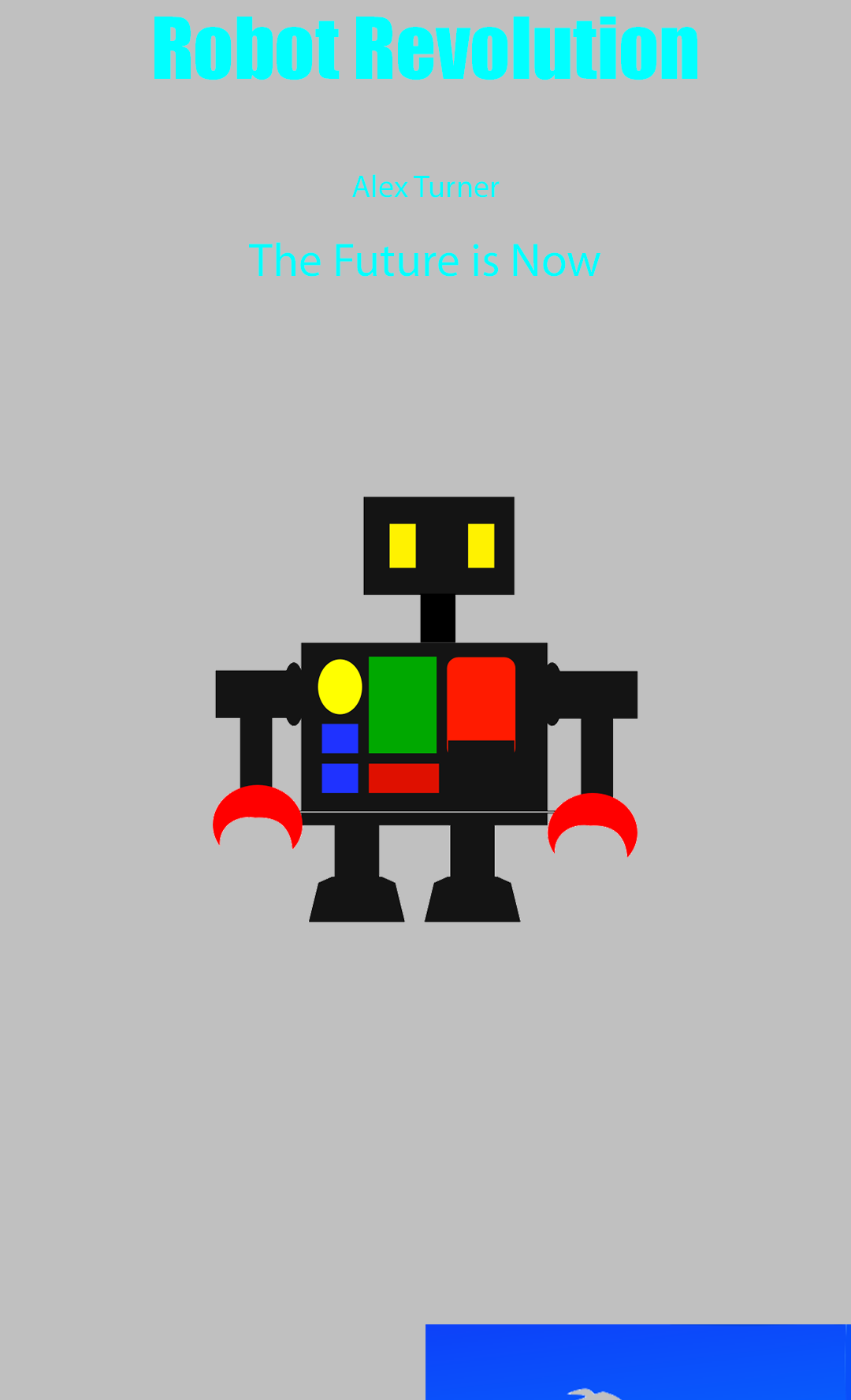}
        }
    \end{minipage} &
    \begin{minipage}{0.3\textwidth}
        Successfully incorporate all design details but fail to position the image of a futuristic cityscape in the center. (\textsc{Gemma-2 9b})
    \end{minipage} \\
    \midrule

    \textbf{Business Card} &
    \begin{minipage}{0.7\textwidth}
        Create an business card design for `Sweet Tooth' Candy Shop with a pastel pink background. Please include the company name in huge white font at the top center. I want to include a candy illustration centered below the company name and contact details of `Phone: +1 555-1234\textbackslash n Email: info@sweettooth.com\textbackslash n Address: 123 Candy Lane, Sugar Town' in medium white font, placed bottom center.
    \end{minipage} & 
    \begin{minipage}{0.25\textwidth}
        \centering
        \makebox[\textwidth]{ % Ensures both images fit inside the column
            \includegraphics[width=\textwidth]{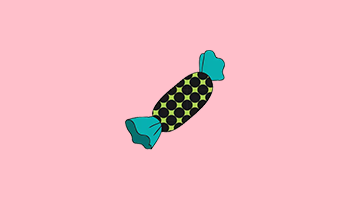}
        }
    \end{minipage} &
    \begin{minipage}{0.3\textwidth}
        Execute image-related manipulations with one expert agent but fail to transfer text-related tasks to the next agent. (\textsc{Gemma-2 27b})
    \end{minipage} \\
    \midrule

    \textbf{Business Card} &
    \begin{minipage}{0.7\textwidth}
        I need help creating a business card design for a multimedia artist and animator named `Pixel Dreams' with a dark blue background. Please include the company name in huge white font centered. Add a tagline `Animation \& Multimedia Art' in medium light blue font below the company name. I want a pixelated star illustration of medium size placed above the company name. Also, include the contact details `Email: contact@pixeldreams.com\textbackslash n Phone: +1 234-567-890\textbackslash n Website: www.pixeldreams.com' in small white font centered.
    \end{minipage} & 
    \begin{minipage}{0.25\textwidth}
        \centering
        \makebox[\textwidth]{ % Ensures both images fit inside the column
            \includegraphics[width=\textwidth]{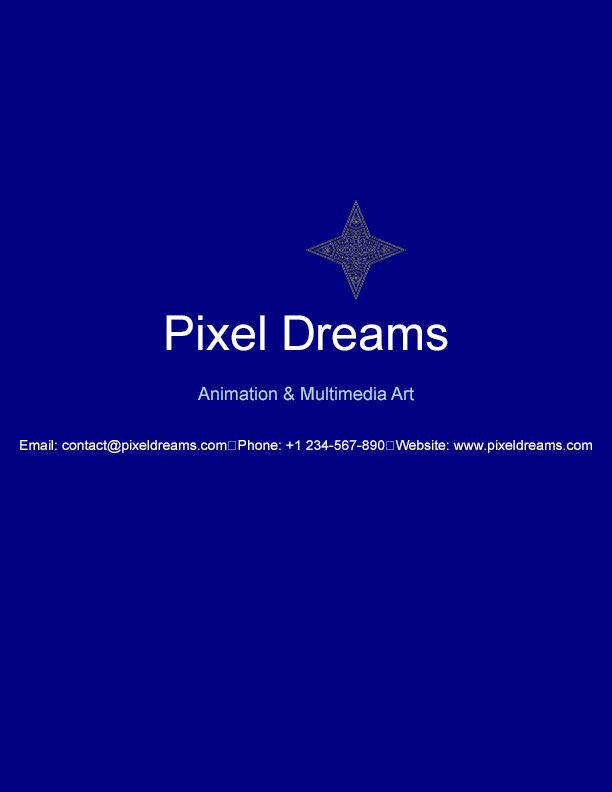}
        }
    \end{minipage} &
    \begin{minipage}{0.3\textwidth}
        Successfully incorporate all design details but fail to add ``\textbackslash n'' in contact details and match the business card dimensions. (\textsc{Gemma-2 27b})
    \end{minipage} \\
    \midrule

    \textbf{Postcard} &
    \begin{minipage}{0.7\textwidth}
        Create an earth day postcard design with the message `Love Your Mother Earth!' centered at the top in green on a light blue background, featuring a large globe with hands wrapped around it centered below the message and medium-sized trees around the globe.
    \end{minipage} & 
    \begin{minipage}{0.25\textwidth}
        \centering
        \makebox[\textwidth]{ % Ensures both images fit inside the column
            \includegraphics[width=\textwidth]{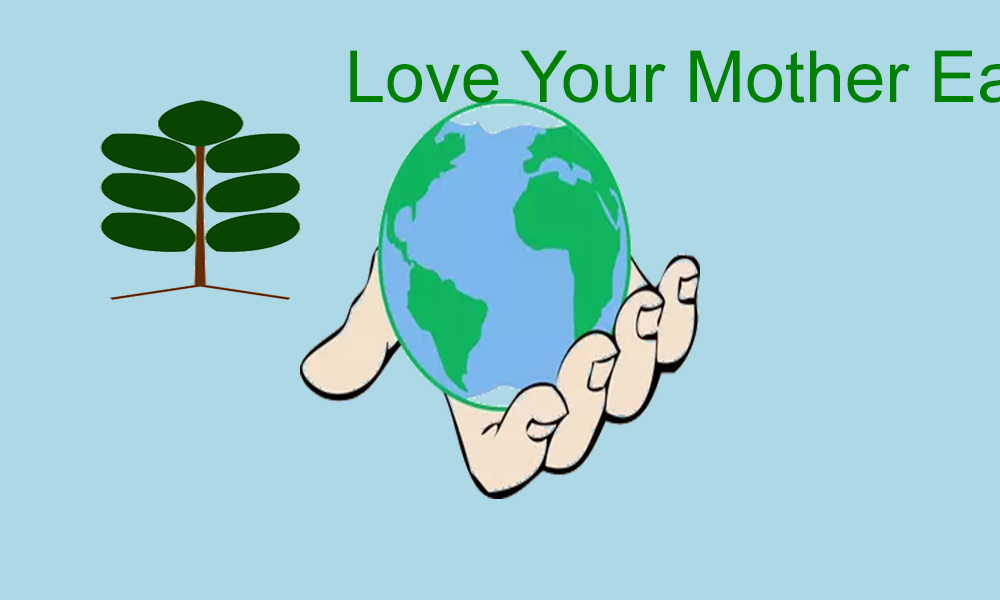}
        }
    \end{minipage} &
    \begin{minipage}{0.3\textwidth}
        Successfully incorporate all design details but fail to position the message centered at the top. (\textsc{Gemma-2 9b})
    \end{minipage} \\
    \midrule

    \textbf{Poster} &
    \begin{minipage}{0.7\textwidth}
        Can you assist in creating a promotional poster design for Spring Blossom Festival on a pink background, featuring a large illustration of cherry blossoms at the top, a huge bold title `SPRING BLOSSOM FESTIVAL' in white font at the center, and event details `April 5-7, 2023\textbackslash n Botanical Gardens, Kyoto' in medium white font at the bottom center.
    \end{minipage} & 
    \begin{minipage}{0.25\textwidth}
        \centering
        \makebox[\textwidth]{ % Ensures both images fit inside the column
            \includegraphics[width=\textwidth]{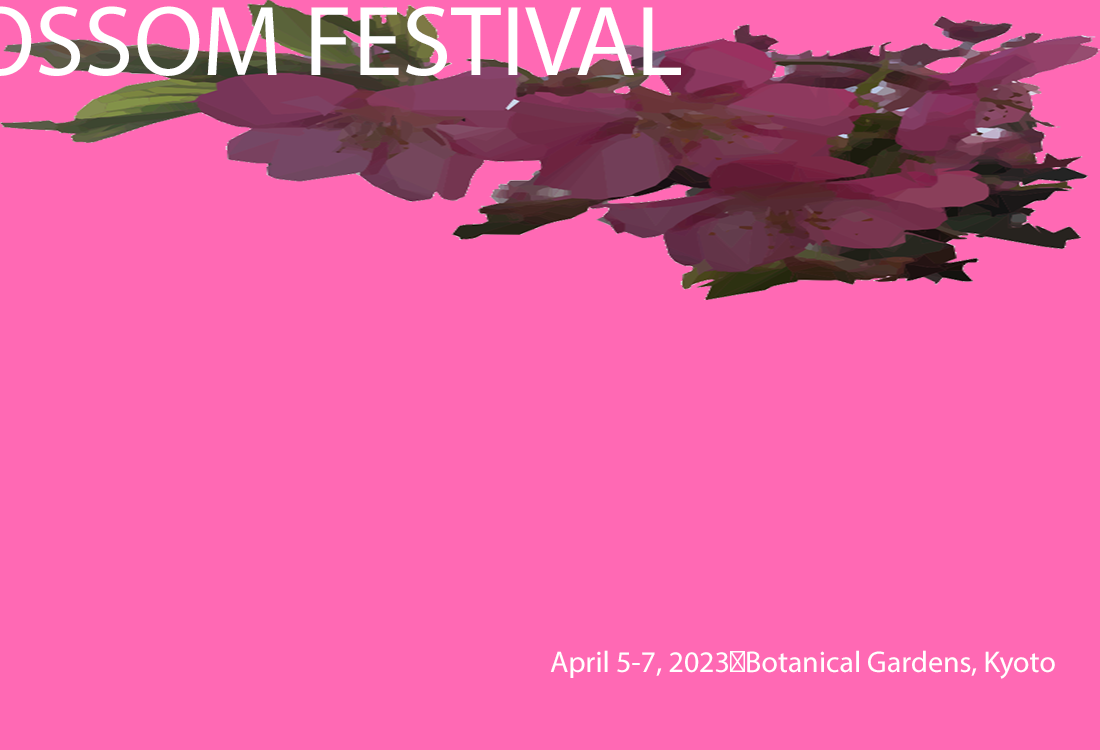}
        }
    \end{minipage} &
    \begin{minipage}{0.3\textwidth}
        Fail to position the title at the center and the event details text at bottom center. (\textsc{Qwen-2.5 7b})
    \end{minipage} \\
    \midrule

    \textbf{Poster} &
    \begin{minipage}{0.7\textwidth}
        Can you assist in creating a poster design for a photography workshop on a beige background, featuring a large image of a vintage camera in the center, the workshop title `CAPTURE THE MOMENT' in large dark brown font at the top center, and details about the workshop `Photography Workshop\textbackslash n July 10-12, 2023\textbackslash n Downtown Studio\textbackslash n Sign up at www.photoworkshop.com' in medium dark brown font at the bottom center.    
    \end{minipage} & 
    \begin{minipage}{0.25\textwidth}
        \centering
        \makebox[\textwidth]{ % Ensures both images fit inside the column
            \includegraphics[width=\textwidth]{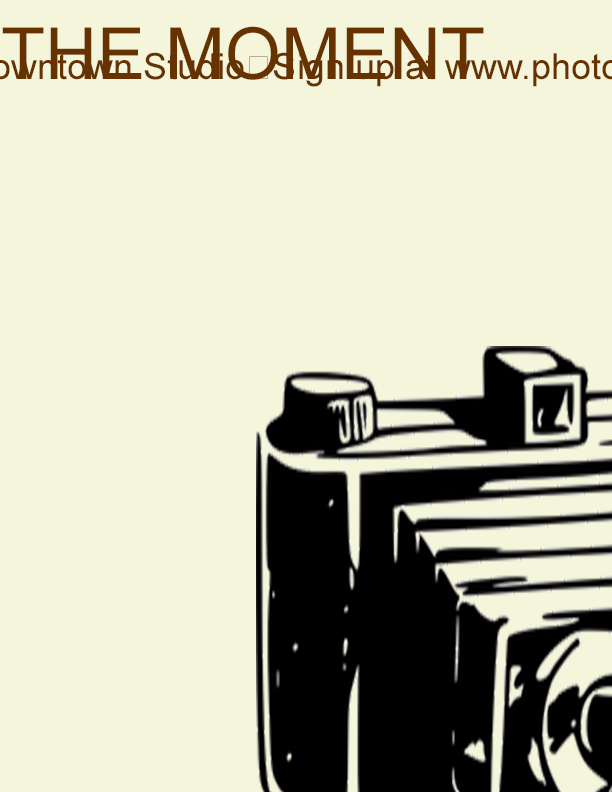}
        }
    \end{minipage} &
    \begin{minipage}{0.3\textwidth}
       Fail to position the image of a vintage camera in the center, details at bottom center, and resize the title to match the dimensions. (\textsc{GPT-3.5})
    \end{minipage} \\

    \bottomrule
    \end{tabular}
}
\caption{Case studies of failed execution outcomes per design type across models. Failures primarily stem from lack of spatial understanding, resulting in incorrect parameter values and misunderstanding of global dependencies between expert agents.}
\label{tab:case_study}
\end{table*}

\begin{figure*}
\begin{prompt}[title={JavaScript for \texttt{AdjustBC} (Photo Editor agent)}]
\begin{lstlisting}[basicstyle=\ttfamily\tiny]
function promptForLayerName() {
    var layerName = arguments[0];
    
    if (layerName == null || layerName == "") {
        throw new Error("Layer with the name '" + layerName + "' does not exist.");
    }
    return layerName;
}

function promptForAdjustmentValues() {
    var brightness = parseInt(arguments[1], 10);
    var contrast = parseInt(arguments[2], 10);

    if (isNaN(brightness) || isNaN(contrast)) {
        throw new Error("Invalid input provided. Please run the script again and provide valid numbers.");
    }
    return { brightness: brightness, contrast: contrast };
}

function layerExists(layerName) {
    var ref = new ActionReference();
    ref.putName(charIDToTypeID("Lyr "), layerName);
    try {
        var desc = executeActionGet(ref);
        return true;
    } catch (e) {
        return false;
    }
}

function selectLayerByName(layerName) {
    var idselect = charIDToTypeID("slct");
    var desc = new ActionDescriptor();
    var idnull = charIDToTypeID("null");
    var ref = new ActionReference();
    var idLyr = charIDToTypeID("Lyr ");
    ref.putName(idLyr, layerName);
    desc.putReference(idnull, ref);
    var idMkVs = charIDToTypeID("MkVs");
    desc.putBoolean(idMkVs, false);
    executeAction(idselect, desc, DialogModes.NO);
}

function applyBrightnessContrastAdjustment(brightness, contrast) {
    var idBrtC = charIDToTypeID("BrgC");
    var desc = new ActionDescriptor();
    desc.putUnitDouble(charIDToTypeID("Brgh"), charIDToTypeID("#Prc"), brightness);
    desc.putUnitDouble(charIDToTypeID("Cntr"), charIDToTypeID("#Prc"), contrast);
    executeAction(idBrtC, desc, DialogModes.NO);
}

function adjustBrightnessContrast() {
    if (!layerExists(layerName)) {
        throw new Error("Layer with the name '" + layerName + "' does not exist.");
    }

    var layerName = promptForLayerName();
    if (layerName == null) {
        throw new Error("Layer name does not exist.");
    }

    var adjustments = promptForAdjustmentValues();
    if (adjustments == null) {
        throw new Error("Parameter values are not provided.");
    }

    selectLayerByName(layerName);
    applyBrightnessContrastAdjustment(adjustments.brightness, adjustments.contrast);
}

adjustBrightnessContrast();
\end{lstlisting}
\end{prompt}
\caption{JavaScript code snippet for \texttt{AdjustBC} for the Photo Editor agent.}
\label{fig:javascript}
\end{figure*}

\end{document}